\documentclass[dvipsnames,table]{article}

\PassOptionsToPackage{numbers, compress}{natbib}

\usepackage{amsmath,amsfonts,bm}

\def\eqref#1{equation~\ref{#1}}

\def\1{\bm{1}}

\def\rk{{\textnormal{k}}}

\DeclareMathAlphabet{\mathsfit}{\encodingdefault}{\sfdefault}{m}{sl}
\SetMathAlphabet{\mathsfit}{bold}{\encodingdefault}{\sfdefault}{bx}{n}

\def\gA{{\mathcal{A}}}

\def\gC{{\mathcal{C}}}

\def\gE{{\mathcal{E}}}
\def\gF{{\mathcal{F}}}
\def\gG{{\mathcal{G}}}

\def\gI{{\mathcal{I}}}
\def\gJ{{\mathcal{J}}}

\def\gN{{\mathcal{N}}}
\def\gO{{\mathcal{O}}}

\def\gT{{\mathcal{T}}}

\newcommand{\R}{\mathbb{R}}

\usepackage[main, final]{neurips_2026}

\usepackage{wrapfig}
\usepackage[utf8]{inputenc} 
\usepackage[T1]{fontenc}    
\usepackage{hyperref}       
\usepackage{url}            
\usepackage{booktabs}       
\usepackage{amsfonts}       
\usepackage{amsmath}
\usepackage{amssymb}
\usepackage{mathtools}
\usepackage{multirow,adjustbox}
\usepackage{physics}
\usepackage{algorithm}
\usepackage{algpseudocode}
\usepackage{multicol}
\usepackage{ifthen}

\usepackage{subcaption, graphicx}

\usepackage{amsthm}
\usepackage{nicefrac}       
\usepackage{microtype}      
\usepackage{xcolor}
\usepackage{tcolorbox}

\usepackage{paralist}
\usepackage{comment}
\usepackage{enumitem}
\usepackage{soul}

\usepackage[capitalize,noabbrev]{cleveref}
\usepackage[textsize=tiny]{todonotes}
\usepackage{arydshln}

\theoremstyle{plain}
\newtheorem{theorem}{Theorem}[section]

\newtheorem{lemma}[theorem]{Lemma}

\newtheorem*{reptheorem}{Theorem}
\theoremstyle{definition}
\newtheorem{definition}[theorem]{Definition}

\theoremstyle{remark}

\newtheorem{example}{Example}
\DeclareMathOperator{\HASH}{HASH}

\definecolor{stdblue}{HTML}{D9E0F3}
\definecolor{bestgray}{HTML}{EFEFEF}

\crefformat{enumi}{\textbf{#2Q#1#3}}

\newcommand{\TODO}[2]{%
    \ifthenelse{\equal{#1}{Marco}}{\textcolor{blue}{\textbf{TODO (#1):} #2}}{%
    \ifthenelse{\equal{#1}{Martin}}{\textcolor{red}{\textbf{TODO (#1):} #2}}{%
    \ifthenelse{\equal{#1}{G}}{\textcolor{violet}{\textbf{TODO (#1):} #2}}{%
    \ifthenelse{\equal{#1}{L}}{\textcolor{orange}{\textbf{TODO (#1):} #2}}{%
    \textcolor{red}{\textbf{TODO (#1):} #2}}}}}}%

\newcommand{\cC}{\mathcal{C}}
\newcommand{\cT}{\mathcal{T}}
\newcommand{\cN}{\mathcal{N}}

\newcommand{\cJ}{\mathcal{J}}
\newcommand{\cV}{\mathcal{V}}

\title{HOPSE: Scalable Higher-Order Positional and Structural Encoder for Combinatorial Representations}

\author{%
  Guillermo Bernárdez\thanks{Equal contribution. Correspondence: \texttt{topological.intelligence@gmail.com}.} \\
  University of California, Santa Barbara\\
  \texttt{guillermo\_bernardez@ucsb.edu} \\
  \And
  Marco Montagna\footnotemark[1] \\
  Sapienza University of Rome \\
  \texttt{marco.montagna@uniroma1.it} \\
  \And
  Louis Van Langendonck\footnotemark[1] \\
  Universitat Politècnica de Catalunya \\
  \texttt{louis.van.langendonck@upc.edu}
  \And
  Martin Carrasco \\
  University of Fribourg \\
  \texttt{martin.carrascocastaneda@unifr.ch} \\
  \And
  Amirreza Akbari \\
  Aalto University \\
  \texttt{amirreza.akbari@aalto.fi} \\
  \And
  Louisa Cornelis \\
  University of California, Santa Barbara \\
  \texttt{louisacornelis@ucsb.edu} \\
  \And
  Mathilde Papillon \\
  University of California, Santa Barbara \\
  \texttt{papillon@ucsb.edu}
  \And
  Pere Barlet-Ros \\
  Universitat Politècnica de Catalunya \\
  \texttt{pere.barlet@upc.edu} \\
  \And
  Nina Miolane\thanks{Joint last authors.} \\
  University of California, Santa Barbara \\
  \texttt{ninamiolane@ucsb.edu} \\
  \And
  Lev Telyatnikov\footnotemark[2] \thanks{Research completed prior to joining Capital One.} \\
  IMOS Lab, EPFL \\
  Capital One \\
  \texttt{lev.telyatnikov@capitalone.com} \\
}

\begin{document}

\maketitle

\begin{abstract}

While Graph Neural Networks (GNNs) have proven highly effective at modeling relational data, pairwise connections cannot fully capture multi-way relationships naturally present in complex real-world systems. In response to this, Topological Deep Learning (TDL) leverages more general combinatorial representations ---such as simplicial or cellular complexes--- to accommodate higher-order interactions. Existing TDL methods often extend GNNs through Higher-Order Message Passing (HOMP), but face critical scalability challenges due to the steep complexity overhead of propagating messages through combinatorial structures. 
To overcome this limitation, we propose HOPSE (Higher-Order Positional and Structural Encoder), a framework \emph{free of learnable message passing layers} that uses Hasse graph decompositions to derive efficient and expressive encodings over \emph{arbitrary higher-order domains}. 
Notably, HOPSE training and inference scale linearly with the size of combinatorial representations while preserving the expressive power and permutation equivariance of the HOMP approaches. 
Experiments on molecular and topological benchmarks show that it matches or surpasses state-of-the-art performance while consistently achieving speedups over HOMP-based models, opening a new path for scalable TDL. The code is available at \url{https://github.com/geometric-intelligence/topobench}.
\end{abstract}



\section{Introduction}
\label{sec:introduction}

Graph Neural Networks (GNNs)~\citep{scarselli2008graph,gilmer2017neural} have demonstrated impressive capabilities in modeling data structured as graphs~\citep{zhou2020graph,bronstein2021geometric}. However, their reliance on edges to encode interactions inherently limits them to pairwise relationships, which may not fully capture the complexity of many real-world systems. For instance, social dynamics often involve group interactions, while physical phenomena like electrostatic interactions in proteins can span multiple atoms simultaneously.

To address such limitations, Topological Deep Learning (TDL)~\citep{bodnar2023topological,hajij2023tdl} introduces a framework for modeling multi-way interactions using \emph{higher-order topological domains} such as hypergraphs, simplicial, cell, and combinatorial complexes. Within TDL, standard Topological Neural Networks (TNNs)~\citep{feng2019hypergraphconv,bunch2020simplicial,hajij2020cell,hajij2022, papillon2023architectures} extend the graph-based message passing paradigm through Higher-Order Message Passing (HOMP), enabling information flow across all the elements (i.e., nodes, edges, faces, etc.) of a combinatorial representation.
This generalization has shown strong performance across diverse fields--e.g., social networks~\citep{knoke2019social}, protein biology~\citep{jha2022prediction}, physics-informed learning~\citep{wei2024physicsmeetstopologyphysicsinformed}, computer networks~\citep{bernardez2025ordered,bernardez2023topological}--where interactions naturally extend beyond pairwise connections.



However, despite their theoretical advantages, TNNs face substantial challenges stemming from the HOMP paradigm. Unlike GNNs, TNNs' ability to propagate information through the full hierarchy of a higher-order domain--where elements of different \emph{ranks} (e.g., nodes, edges, faces, etc.) are interconnected--leads to an explosion in the number of possible sequences of \emph{message passing routes} across these different levels~\citep{telyatnikov2025topobench,montagna2024topological,papamarkou2024position} (please see \cref{app:expanded_problem_formulation_neighbourhoods_and_paths} for a detailed analysis).
On top of this vast architectural search space, TNNs operating under the HOMP paradigm are also known to be computationally demanding. Message propagation often requires significant distributed computing resources even for moderately sized inputs~\citep{papamarkou2024position,gurugubelli2024sann, leko2026batch}. This combination of combinatorial design complexity and high computational cost during training and inference  \emph{severely limits existing TNNs' scalability}, posing a major barrier to 
large-scale or time-sensitive real-world applications.

\textbf{Contributions.} 
Driven by the need for more scalable TDL, this work introduces \emph{\textbf{Higher-Order Positional and Structural Encoder (\textsc{HOPSE})}}, a novel framework designed to overcome HOMP's associated computational hurdles. Specifically, \textsc{HOPSE} main features and contributions include:
\begin{itemize}[leftmargin=10pt,itemsep=1pt,parsep=0pt,topsep=0pt]

    \item \textbf{Higher-order topological encodings:} By decomposing a combinatorial representation into simpler graph ensembles, called Hasse graphs \citep{hajij2023tdl, papillon2024topotune}, \textsc{HOPSE} leverages efficient and established graph Positional and Structural Encodings (PSEs) to obtain analogous representations for higher-order domains. In particular, our framework accommodates both \textit{connectivity-only} PSEs--relying strictly on topology--and \textit{feature-aware} ones--leveraging both connectivity and original features. 

    \item \textbf{Scalability and computational efficiency:} In contrast to HOMP-based networks, \textsc{HOPSE}'s forward pass complexity scales linearly with respect to the number of elements in the combinatorial representation (\cref{sec:subsec_methodology}). This is achieved by computing the encodings as a one-time preprocessing step, significantly accelerating both training and inference in practice (\cref{sec:experiments}). 
    
    \item \textbf{Theoretical guarantees:} We show that \textsc{HOPSE} is as expressive in capturing relational information as standard HOMP-based models (e.g. \citet{papillon2023architectures}) in distinguishing non-isomorphic combinatorial complexes (\cref{thm:theorem_ccwl}), while preserving cell permutation equivariance (\cref{thm:permutation_eq}) and \emph{avoiding any learnable} message passing during training and inference.

    \item \textbf{Flexible instantiations:} Two realizations are presented: \textbf{HOPSE-M}, relying on handcrafted graph-theoretical encodings, and \textbf{HOPSE-GPSE}, using a pretrained \textsc{GPSE} encoder~\citep{canturk2023graph}--offering trade-offs between preprocessing cost and representational richness (\cref{sec:realization_of_framework}).
        
    \item \textbf{Extensive empirical validation:} Comprehensive experiments on molecular benchmarks~\citep{morris2020,huang2022artificial}, and \texttt{MANTRA}~\citep{ballester2025mantra} validate the practical effectiveness of \textsc{HOPSE}, showcasing its ability to efficiently capture complex relational structures and outperform existing baselines while being consistently faster than HOMP-based models (\cref{sec:experiments}).

\end{itemize}

To ensure full reproducibility, we integrate \textsc{HOPSE} into the open source \texttt{TopoBench}~\citep{telyatnikov2025topobench} (\url{https://github.com/geometric-intelligence/topobench.git}), including the exact hyperparameter search configurations and ablation scripts used to generate our results.

\section{Background}%
\label{sec:background}

This section introduces key concepts underpinning TDL. For brevity and generality, the following discussion centers on combinatorial complexes, which subsume all of the discrete topological domains leveraged by TDL~\citep{hajij2023tdl}. \cref{app:extended_background} provides mathematical definitions of these other domains, such as simplicial and cellular complexes, along with an extended theoretical background.

\paragraph{Combinatorial complex.} A \emph{combinatorial complex} $\gT$ is a triple $(\mathcal{V}, \mathcal{C}, \textrm{rk})$ consisting of a set $\mathcal{V}$, a subset $\mathcal{C}$ of the powerset $\mathcal{P}(\mathcal{V}) \backslash{\emptyset}$, and a rank function $\textrm{rk}: \mathcal{C} \rightarrow \mathbb{Z}_{\geq 0}$ with the following properties:
\begin{enumerate}[leftmargin=20pt,itemsep=1pt,parsep=0pt,topsep=0pt]
    \item For all $v \in \mathcal{V},\{v\} \in \mathcal{C}$ and $\textrm{rk}(\{v\})=0$;
    \item The function $\textrm{rk}$ is order-preserving, i.e., if $\sigma, \tau \in \mathcal{C}$ satisfy $\sigma \subseteq \tau$, then $\textrm{rk}(\sigma) \leq$ $\textrm{rk}(\tau)$;
\end{enumerate}
The elements of $\mathcal{V}$ represent the nodes, while the elements of $\mathcal{C}$ are called cells (i.e., groups of nodes). A cell $\sigma \in \mathcal{C}$ of rank $k:=\textrm{rk}(\sigma)$ is referred to as a $k$-cell. Lastly, the dimension of $\gT$ is defined as the maximal rank among its cells: $\mathrm{dim}(\mathcal{C}):= \max_{\sigma \in \mathcal{C}} \textrm{rk}(\sigma)$. 

\paragraph{Signals over combinatorial complexes.}
A combinatorial complex $\mathcal{T} = (\mathcal{V}, \mathcal{C}, \mathrm{rk})$ can be equipped with signals, generalizing the concept of featured graph\footnote{Attributed graph and featured graph are used interchangeably to comply with the literature.} (see \cref{def:featured_graph}) to higher-order structures. In this context, a signal is a set of functions $\mathcal{F} = \{F_r\}_{r \geq 0}$, where $F_r: \{\sigma \in \mathcal{C} \mid \mathrm{rk}(\sigma) = r\} \to \mathbb{R}^{D_r}$, where $D_r$ is the dimension of the feature space for rank-$r$ cells. For this work, it is important to emphasize that a complex can have multiple attribute functions over each $r$-cell. More formally, $\mathcal{F} = \{F_{r,k}\}_{k=1}^{K}$ for all $r \geq 0$, where each $F_{r,k}$ shares the same domain: $\text{Dom}(F_{r,k}) = \text{Dom}(F_r)$.

\begin{figure}[t]
    \vspace{-20pt}
    \centering
    \includegraphics[width=\textwidth]{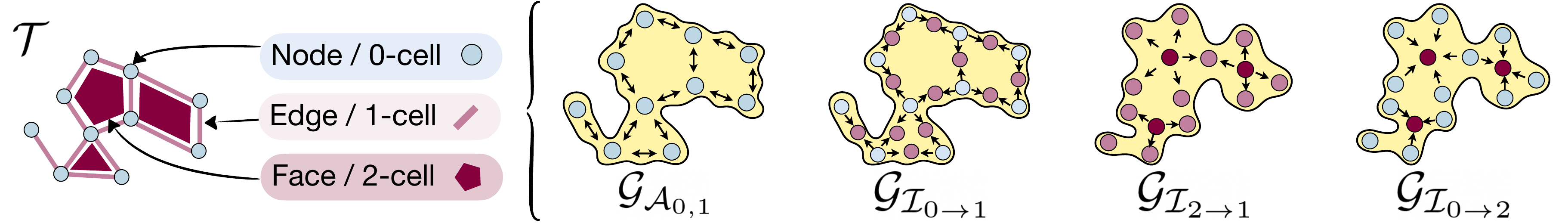}
\caption{\textbf{Neighborhood expansions of a complex.} Given a complex $\gT$ (left), four examples of strictly augmented Hasse graphs $\mathcal{G}_\mathcal{N}$ are illustrated corresponding to 4 neighborhood functions $\mathcal{N}$: adjacency of nodes w.r.t. edges ($\gA_{0,1}$), incidence from nodes to edges ($\gI_{0\to 1}$), incidence from faces to nodes ($\gI_{2\to 1}$), and incidence from nodes to faces ($\gI_{0\to 2}$).} 

\label{fig:background}
\vspace{-10pt}
\end{figure}

\paragraph{Neighborhood Relations: Adjacencies and Incidences.} Combinatorial complexes can be equipped with a notion of neighborhood among cells that induces  topological structures. 
Formally, denoting by $\mathcal{P}(\cdot)$ the power set, a neighborhood function $\mathcal{N}: \mathcal{C} \rightarrow \mathcal{P}(\mathcal{C})$ on a complex $\gT$ maps each cell $\sigma \in \mathcal{C}$ to a collection of \emph{neighbor cells} $\mathcal{N}(\sigma) \subset \mathcal{C}$. Traditional neighborhood functions include \emph{adjacencies} ($\mathcal{A}_{t,s}$), connecting cells of the \textit{same rank} $t$ through relations w.r.t. cells of another rank $s$, and \emph{incidences} ($\mathcal{I}_{s\to t}$), linking cells of different ranks from a source rank $s$ to a \textit{different} target rank $t$. \cref{fig:background} presents some instances, and \cref{app:background_neighbourhoods} provides formal definitions.

\paragraph{Strictly augmented Hasse graphs.} Given a complex $\gT$, each particular neighborhood function $\mathcal{N}$ induces a strictly augmented Hasse graph $\mathcal{G}_{\mathcal{N}} = (\mathcal{C}_\mathcal{N}, \mathcal{E}_\mathcal{N})$ \citep{papillon2024topotune}, defined as the directed graph whose nodes and edges are given, respectively, by 
$\mathcal{C}_{\gN} = \{ \sigma \in \gC \mid \gN(\sigma) \neq \emptyset \}$ and $\mathcal{E}_{\gN} = \{ (\tau, \sigma) \mid \sigma, \tau \in \gC_{\gN},\ \tau \in \gN(\sigma) \}$.
\cref{fig:background} shows some examples of the graph expansions induced by these neighborhood functions. Notably, \emph{strictly} comes from the fact that the set of nodes of $\mathcal{G}_{\gN}$ is restricted to those cells that have at least one neighbor according to $\gN$ ($\mathcal{C}_{\gN}$). In contrast, the original definition of \emph{augmented Hasse graph} considers the full set $\mathcal{C}$~\citep{hajij2023tdl}.

\section{Related Works}

\paragraph{Encodings in GNNs.} 
A range of positional and structural encodings (PSEs) have been proposed to enhance node representations. Positional encodings typically capture a node’s location within the graph (e.g., via distances or coordinates), while structural encodings describe the node’s role or function in the topology (e.g., node degrees). 
Classical approaches of positional encodings include spectral methods based on Laplacian eigenvectors ~\citep{JMLR:v24:22-0567}, while a common structural encoding is the random-walk embedding ~\citep{dwivedi2022long, wang2022equivariant}, which reflects multi-hop connectivity; these handcrafted features help differentiate nodes with otherwise similar local neighborhoods. More recently, transformer-based GNNs have incorporated PSEs into attention mechanisms \citep{dwivedi2022graph, rampavsek2022recipe,ma2023graph}. 
Other works focus on decoupling the positional information from the structure and learning the embeddings instead of picking them manually~\citep{dwivedi2022graph}. 
Building upon this, Graph Positional and Structural Encoder (\textsc{GPSE})~\citep{canturk2023graph} offers a unified and transferable solution by learning a shared latent space of PSEs through self-supervised pretraining. \textsc{GPSE}’s embeddings can be integrated into downstream GNNs, achieving competitive performance and advancing the development of general-purpose graph models. 

\paragraph{Combinatorial Complex Neural Networks.}
To overcome the pairwise limitations of graphs, recent work explicitly represents higher-order interactions through topological domains. 
Combinatorial complexes provide a unified framework for modeling higher-order interactions with hierarchical organization, and their associated Combinatorial Complex Neural Networks (\textsc{CCNNs}) enable message passing across cells of varying rank (HOMP paradigm), naturally capturing multi-way relationships \citep{hajij2023tdl}.
Generalizing and subsuming \textsc{CCNNs}, \textsc{TopoTune}~\citep{papillon2024topotune} offers a more flexible and powerful framework: it systematically decompose a complex into a set of strictly augmented Hasse graphs--each one capturing a particular neighborhood--in a modular way that preserves the higher-order structure. In doing so, \textsc{TopoTune} not only facilitates the design of new TNN architectures but also help bridge the gap between graph deep learning and TDL. However, existing TNNs 
suffer from the high computational complexity of HOMP, limiting their practical use in large real-world applications.

\paragraph{Encodings in Higher-Order Domains.}
Encoding positional information for higher-order elements is challenging: unlike graphs where features reside on nodes, each cell rank in a combinatorial complex potentially requires its own descriptor. While largely unexplored, some works address this. For instance, Cellular Transformer (\textsc{CT})~\citep{ballester2024attending} extends transformers using manually crafted positional encodings, but its attention mechanism incurs quadratic complexity, limiting scalability. Alternatively, Simplicial-Aware Neural Networks (\textsc{SANN})~\citep{gurugubelli2024sann} bypass message passing by precomputing multi-hop feature aggregations across simplicial neighborhoods into fixed-length vectors. Despite achieving constant overhead, \textsc{SANN} is fundamentally limited because: \textit{(i)} it is restricted to simplicial complexes, and \textit{(ii)} it relies on hard-coded aggregations rather than arbitrary structural encodings, missing potentially rich higher-order patterns.

\paragraph{In Pursuit of Efficiency.}
While precomputing structural information accelerates GNNs---e.g., via adjacency powers~\citep{frasca2020sign}, subgraph aggregation~\citep{zeng2021decoupling}, or feature partitioning~\citep{li2021training}---this paradigm remains underexplored in higher-order domains. Recent efforts like Simplicial Scattering Networks~\citep{madhu2024simplicialunsupervised} use simplicial relationships to design parameter-free architectures. To our knowledge, SANN~\citep{gurugubelli2024sann} is the only approach efficiently precomputing higher-order encodings within a learning framework, but remains limited to simplicial feature aggregation. In contrast, \textsc{HOPSE} is fundamentally more general: where \textsc{SANN} is restricted to simplicial complexes and fixed aggregations, \textsc{HOPSE} handles \emph{any} combinatorial complex by decomposing it into Hasse graph ensembles~\citep{papillon2024topotune}, enabling the use of \emph{any} established graph PSE to generate higher-order encodings. 

\section{HOPSE: Higher-Order Positional and Structural Encoder}
\label{sec:HOPSE}

\cref{sec:subsec_methodology} outlines the proposed \textsc{HOPSE} pipeline, \cref{sec:subsec_theorerical} establishes its theoretical guarantees, and \cref{sec:realization_of_framework} presents two practical realizations of the framework.

\begin{tcolorbox}[
    colback=white,
    colframe=black,
    boxrule=0.5pt,
    left=5pt,
    right=5pt,
    top=5pt,
    bottom=5pt,
    boxsep=0pt,
    arc=10pt,
    outer arc=10pt
]

Let $\gT$ be an arbitrary combinatorial complex, and $\gN_C \subset \gN_\gT$ any subset of neighborhoods functions on $\gT$. The following \textbf{notation} is introduced to support the presentation of the framework:

\paragraph{Strictly augmented multi-attributed Hasse graphs.} Let $\gN \in \gN_\gC$ be a neighborhood function. A \emph{strictly augmented multi-attributed Hasse graph} denoted by $\gJ_{\gN} = (\gC_\gN, \gE_\gN,\mathcal{F}_\gN)$  is a strictly augmented Hasse graph $\mathcal{G}_{\gN}$ \cite{papillon2024topotune} with a set of signals $\mathcal{F_N}$, both induced by $\mathcal{N}$. The shorthand $\gF_{\gN, k} = \{F_{r, k}\}_{r\geq 0}$ is the set of all $k$-attribute functions for each rank $r$ that act on the subset of cells in a neighborhood $\gC_\gN$ as $F_{r,k}: \{ \sigma \in \gC_\gN\} \to \R^{D_r}$.
To streamline the management of multiple associated signals, the \textit{restriction} of $\gJ_{\gN}$ to the $k$-th feature is defined as $\eval{\gJ_{\gN}}_k = (\gC_\gN, \gE_\gN, \gF_{\gN, k})$. Lastly, $\mathbf{J}_C = \left\{ \gJ_{\gN} \right\}_{\gN \in \gN_C}$ denotes the corresponding \emph{Hasse graph decomposition} of $\gT$ w.r.t. a given neighborhood function set $\gN_C$.

\paragraph{Rank-targeted neighborhoods functions.}
The \emph{rank-targeted neighborhood} is defined as the collection of neighborhood functions whose target cells have the same fixed rank $r$: $\mathcal{N}^{\to r} := \{ \mathcal{N} \in \mathcal{N}_{\mathcal{C}} \mid \forall \sigma \in \mathcal{C}, \forall \tau \in \mathcal{N}(\sigma): \mathrm{rk}(\tau) = r \}$. 
\end{tcolorbox}

\subsection{Methodology}
\label{sec:subsec_methodology}
\begin{figure}[t]
    \vspace{-20pt}
    \centering
    \includegraphics[width=\textwidth]{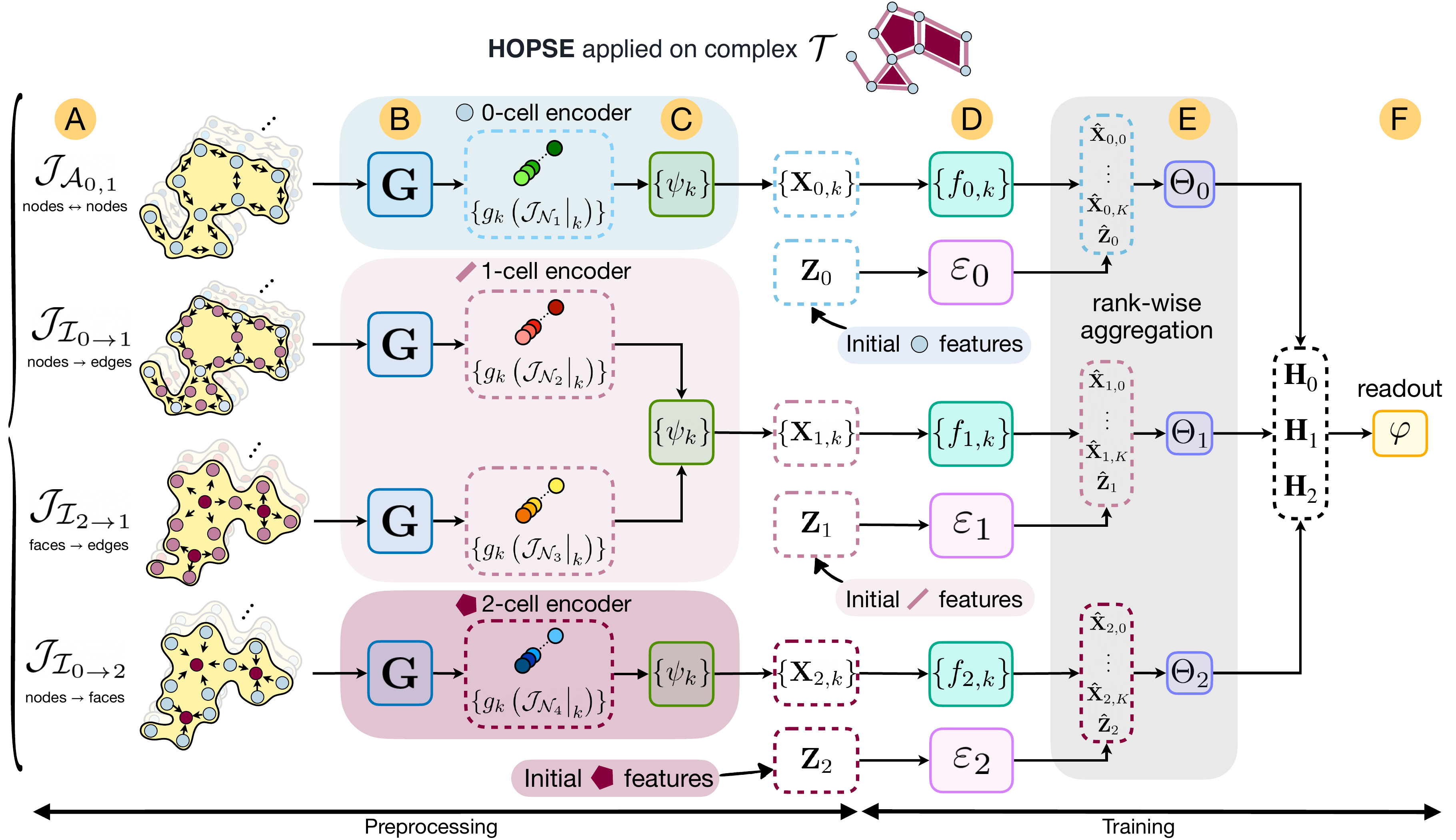}
\caption{\textbf{HOPSE pipeline.} (\textbf{A}) Considering the collection of neighborhood functions $\gN_C =\{\gA_{0,1}, \gI_{0\to 1}, \gI_{2\to 1}, \gI_{0\to 2}\}$, the input combinatorial complex \(\mathcal{T}\) is decomposed into the corresponding multi-attributed Hasse graphs expansions $\{\mathcal{J}_{\gN}\}_{\gN_C}$, each one with $k$ associated signals. 
(\textbf{B}) PSEs are computed for each Hasse graph, \(\mathbf{G}(\mathcal{J}_{\mathcal{N}_i})\), as defined in \cref{eq:family_func_g}. 
(\textbf{C}) Higher-order PSEs are derived via the target rank-aware aggregation $\psi_k$ (see \cref{eq:learn_f}). 
(\textbf{D}) The PSEs $\mathbf{X}_{r,k}$ are transformed using $f_{r,k}(\cdot)$, and initial cell features $\mathbf{Z}_r$ are embedded via $\varepsilon_r(\cdot)$, following \cref{eq:x_cal_hat,eq:initial_features}. 
(\textbf{E}) Rank-specific, structure-aware representations $\mathbf{H}_r$ are learned using $\Theta_r(\cdot)$, as described in \cref{eq:final_embedding}. 
(\textbf{F}) The final task-specific readout $\varphi(\cdot)$ integrates all $\mathbf{H}_r$ representations (\cref{eq:task_specific}).}


\label{fig:pipeline}
\vspace{-10pt}
\end{figure}

This section describes \textsc{HOPSE}, a method for enriching cell features in combinatorial complexes with positional and structural encodings derived from Hasse graph decompositions. \textsc{HOPSE} enables expressive higher-order modeling without relying on the computational overhead of HOMP. 
The full pipeline is illustrated in \cref{fig:pipeline}, with references using the corresponding labels (A–F). 

\paragraph{Extract positional and structural information.} 
Let $\gN_C$ be a collection of neighborhood functions. For each neighborhood function $\gN \in \gN_C$, the Hasse graph decomposition of $\gT$ enables the treatment of every $\gJ_{\gN}$ as an independent component (see \cref{fig:pipeline}\textbf{A}), each representing a unique set of connectivity relationships.
$\mathbf{G}$ is a set of Positional and Structural Encoders (PSEs) such that
\begin{equation}
  \mathbf{G} \colon \{1,\dots,K\} \times \mathbf{J}_\gC \longrightarrow \mathbb{R}^{N \times D_k}, \quad
  \left(k, \gJ_{\gN}\right) \longmapsto g_k \left( {\gJ_{\gN}\bigr|}_k \right).
\label{eq:family_func_g}
\end{equation}
Here $g_k(\cdot): \mathbf{J}_\gC \to \mathbb{R}^{|\gC_\gN|_{\text{target}} \times D_k}$ denotes a particular PSE, with $|\gC_\gN|_{\text{target}}$ being the number of nodes in the Hasse graph constructed from the neighborhood function $\gN$ corresponding to the target cells (shown in \cref{fig:pipeline}\textbf{B}). 
Each $g_k(\cdot)$ maps to a different $D_k$, allowing PSEs to have different dimensions.

\paragraph{Aggregate encodings across neighborhoods.}
The proposed neighborhood aggregation mechanism $\psi_k(\cdot)$ operates by grouping the encodings of cells with the same target rank $r$ to produce a unique embedding for each cell and each $k \in \{1, \dots, K\}$ (see \cref{fig:pipeline}\textbf{C}). The embeddings for cells of rank $r$ are computed as:
\begin{equation}\label{eq:learn_f}
    \mathbf{X}_{r, k} = \psi_k(\{ g_k \left({\gJ_{\gN}\bigr|}_k \right) \mid \gN \in \gN^{\to r} \}),
\end{equation}

where $\mathbf{X}_{r, k} \in \R^{N_{r} \times D_{k'}}$, with $N_{r}$ denoting the number of $r$-cells and $D_{k'}$ representing the output feature dimension for attribute $k$. The functions $\psi_k(\cdot)$ can also be interpreted as rank-level aggregators, where the aggregation operation can be any standard reduction method. Common choices include concatenation, sum, average, or even learnable transformations such as MLPs. The main advantage of using non-learnable aggregation functions for $\psi_k$ is that the embeddings $\mathbf{X}_{r, k}$ can be precomputed during dataset preprocessing, since they do not require learning. 


\paragraph{Learning embeddings of cells.}
In the \textsc{HOPSE} framework, each $r$-cell is associated with two complementary sources of information: \textit{(i)} higher-order positional and structural encodings $\mathbf{X}_{r,k}$ aggregated from multiple neighborhoods, and \textit{(ii)} initial features $\mathbf{Z}_r$ provided by the dataset or derived during the construction of the higher-order structure. Both types of features are independently projected into latent spaces through sets of learnable functions (see \cref{fig:pipeline}\textbf{D}):
\vspace{-1pt}
\begin{align}
\hat{\mathbf{X}}_{r,k} &= f_{r,k}(\mathbf{X}_{r,k}), \label{eq:x_cal_hat} \\
\hat{\mathbf{Z}}_r &= \varepsilon_r(\mathbf{Z}_r), \label{eq:initial_features}
\end{align}
where $f_{r,k}(\cdot)$ is applied to each feature $k$ of the $r$-cells, and $\varepsilon_r(\cdot)$ to their initial features.

\paragraph{Learning neighborhood-aware representations.}  

Next, to integrate the structural information with the cells' initial attributes, the extracted and transformed higher-order PS encodings and the cell features are passed through the aggregator $\Theta_r(\cdot)$ (see \cref{fig:pipeline}\textbf{E}) to obtain the final neighborhood-aware representations \( \mathbf{H}_r \in \mathbb{R}^{N_r \times D_r} \):
\begin{equation} \label{eq:final_embedding}
    \mathbf{H}_{r} = \Theta_{r}(\hat{\mathbf{Z}}_r, \hat{\mathbf{X}}_{r,1}, \dots, \hat{\mathbf{X}}_{r,K}).
\end{equation}

\paragraph{Learning the downstream task.} Since the structural information is now contained in the cell features, they can be treated independently. 
For example, for \emph{complex-level} tasks where the network needs an output for the whole complex, the global transformation $\varphi(\cdot)$ combines 
the aggregated embeddings from all ranks (see \cref{fig:pipeline}\textbf{F}). The overall output is

\begin{equation}\label{eq:task_specific}
\text{HOPSE}(\gT) = \varphi\Bigl(\mathbf{H}_0, \mathbf{H}_1, \ldots, \mathbf{H}_{R}\Bigr).
\end{equation}

\begin{tcolorbox}[
    colback=white,
    colframe=black,
    boxrule=0.5pt,
    left=5pt,
    right=5pt,
    top=5pt,
    bottom=5pt,
    boxsep=0pt,
    arc=10pt,
    outer arc=10pt
]
\paragraph{Computational Complexity.}
Notably, no training is required up to and including \cref{eq:learn_f}, allowing matrices $\mathbf{X}_{r,k}$ to be fully precomputed in a preprocessing phase. By offloading all structure-dependent computations to preprocessing, \textsc{HOPSE}'s complexity scales linearly with the number of cells as
\begin{equation}
    C_{\mathrm{HOPSE}} = \sum_{r=0}^R \gO(N_r D_r^2 (TK+1) L)
\end{equation}
where $N_r$ is the number of $r$-cells, $K$ is the number of characterization functions $g_k(\cdot)$, $T$ is the number of neighborhood functions considered, $D_r$ is the feature dimension of $r$-cells, and $L$ is the number of layers of each MLP that serves to model $f_{r,k}$. The efficiency of this design is further validated by experimental runtime results in \cref{tbl:best_rerun_runtime}. See \cref{app:complexity} for a comprehensive computational and memory complexity analysis, including a comparison with other models.
\end{tcolorbox}

\subsection{Theoretical Guarantees}
\label{sec:subsec_theorerical}

The expressivity of GNNs is commonly analyzed through their correspondence with the Weisfeiler–Leman (WL) isomorphism test \cite{huang2021short}. In the context of TDL, \citet{papillon2024topotune} proposes a generalization of WL over arbitrary neighborhoods of a combinatorial complex (CC), called the CCWL test. The following theorem uses the CCWL to establish the expressivity of \textsc{HOPSE}.
\begin{theorem}
\label{thm:theorem_ccwl}
\textsc{HOPSE} is more powerful than the CCWL test in distinguishing non-isomorphic combinatorial complexes if the functions $\varepsilon_r(\cdot)$, $\psi_k (\cdot)$, $f_{r,k}(\cdot)$, and $\Theta_{r}(\cdot)$ are injective for all $k \in \{1,..., K\}$ and for all $r \in \{0,..., R\}$ and $g_k(\cdot)$ is equally expressive as the WL test on the corresponding Hasse graph for at least one $k \in \{1,..., K\}$.
\end{theorem}

The proof shows that if $g_k(\cdot)$ can distinguish two non-isomorphic Hasse graphs, \textsc{HOPSE}, under our injectivity assumptions, can propagate this distinct structural information to distinguish the corresponding combinatorial complexes. The argument concludes by identifying specific pairs of combinatorial complexes that are distinguished by \textsc{HOPSE} but remain indistinguishable under the CCWL test, thereby establishing that \textsc{HOPSE} is strictly more powerful. Full details and the construction of these cases are provided in \cref{app:proofs}.

Another key property for models working on combinatorial data is cell permutation equivariance---the requirement that if the input cells are reordered, the model's outputs are reordered in the same way. The following theorem establishes that \textsc{HOPSE} preserves this property under reasonable conditions:

\begin{theorem} \label{thm:permutation_eq}
\textsc{HOPSE} is cell permutation equivariant if the functions $\varepsilon_r(\cdot)$, $g_k(\cdot)$, $\psi_k(\cdot)$, $f_{r,k}(\cdot)$, and $\Theta_{r}(\cdot)$ are cell permutation equivariant for all $k \in \{1, \dots, K\}$ and all $r \in \{0, \dots, R\}$.
\end{theorem}


This follows from the fact that compositions of equivariant functions preserve equivariance. The proof is detailed in \cref{app:proofs}. While equivariance is a formal requirement, it is satisfied in \textsc{HOPSE} by applying standard MLPs independently to each cell. While an MLP is not equivariant to permutations of its internal feature dimensions, it behaves equivariantly with respect to the cell set when applied in a shared, row-wise fashion across the rank-$r$ collection.

\subsection{Realization of the Framework}
\label{sec:realization_of_framework}

The practical implementation of the proposed framework requires defining a set of $T$ neighborhood functions of interest $\gN_{C} = \{\gN_1, \dots, \gN_T\}$ (discussed in \cref{sec:experiments}), a set of characterization functions $\mathbf{G}$, neighborhood aggregation functions $\psi_k(\cdot)$, rank-specific processing functions $f_{r,k}(\cdot)$, and a structure mixing function $\Theta_{r}(\cdot)$. In particular, this work adopts $\psi_k(\cdot) = \mathrm{CONCAT}(\cdot)$ for aggregation, fully encoding the output of each $g_k(\cdot)$ . This work adopts trainable multilayer perceptrons (MLPs) for both $f_{r,k}(\cdot)$ and $\Theta_{r}(\cdot)$. The choice of functions $\mathbf{G}$ defines the two realizations of the framework, \textbf{HOPSE-M} and \textbf{HOPSE-GPSE}, offering a trade-off between \emph{representational power} and \emph{computation cost}.\footnote{Note that, in practice, some of our particular \textsc{HOPSE} instantiations might not fulfill the injectivity assumption. However, they help illustrate the flexibility of our proposed framework to accommodate customized $g_k$ functions.} Additional implementation details are provided in \cref{app:framework_realization}. 


\textbf{HOPSE-M\textnormal{(anual)}.} This realization incorporates manually selected positional and structural encodings commonly used in graph learning. Specifically, $\mathbf{G} = \{g_{1}, \ldots, g_{7}\}$ is partitioned into two groups based on whether the encoding depends solely on graph connectivity, or additionally on node features. 
\begin{itemize}[leftmargin=10pt,itemsep=1pt,parsep=0pt,topsep=0pt]
    \item \textbf{Connectivity-only encodings} ($g_1$--$g_4$) capture topological properties independently 
of node attributes:
\begin{inparaenum}[(i)]
    \item Laplacian eigenvector-based encodings (\textsc{LapPE})~\citep{dwivedi2021generalization},
    \item electrostatic potential encodings (\textsc{ElectrostaticPE})~\citep{kreuzer2021rethinking},
    \item random-walk encodings (\textsc{RWSE})~\citep{dwivedi2022graph}, and
    \item heat kernel signatures (\textsc{HKdiagSE})~\citep{sun2009concise}.
\end{inparaenum}

    \item \textbf{Feature-based encodings} ($g_5$--$g_7$) additionally condition on node features to assess whether encodings that couple topology with node semantics yield additional representational benefits:
\begin{inparaenum}[(i)]
\setcounter{enumi}{4}
    \item heat kernel feature encodings (\textsc{HKFE})~\citep{xu2019graphheat},
    \item $K$-hop feature encodings (\textsc{KHopFE})~\citep{feng2022khop}, and
    \item personalized PageRank feature encodings (\textsc{PPRFE})~\citep{Gasteiger2019appnp}. 
\end{inparaenum}
\end{itemize}

\textbf{HOPSE-GPSE.} This variant leverages the pretrained \textsc{GPSE} network~\citep{canturk2023graph} to encode structural and positional information. In this realization, $\mathbf{G} = \{g_{1}\}$ with $g_1 = \textsc{GPSE}(\cdot)$.\footnote{Following its original pre-training protocol, \textsc{GPSE} is fed random noise as input node features rather than the cells' original attributes, which are instead preserved and injected downstream (\cref{fig:pipeline}\textbf{D}).} Unlike manual encodings that rely on computationally intensive operations such as eigendecomposition, \textsc{GPSE} produces latent descriptors using standard message passing. This makes \textbf{HOPSE-GPSE} particularly suitable for large combinatorial complexes with many cells, offering a scalable alternative without significant loss in expressiveness. It is worth mentioning that \textsc{GPSE} comes with its own limitations in terms of expressivity and generalization capability (see \citet{franks2024towards}), and its use is meant to showcase the possibility of using pretrained models as $g_k(\cdot)$. 

\section{Experiments}
\label{sec:experiments}

This section empirically evaluates the proposed HOPSE framework, benchmarked against state-of-the-art TNNs. 
Both instantiations of the framework, \textbf{HOPSE-M} and \textbf{HOPSE-GPSE} (see \cref{sec:realization_of_framework}), are assessed on graph molecular datasets, as well as on expressivity and topological tasks. This work adopts the standardized experimental protocol introduced in \texttt{TopoBench}~\citep{telyatnikov2025topobench} to ensure fair and reproducible comparisons through a comprehensive optimization and evaluation suite.\footnote{\texttt{TopoBench}'s evaluation framework differs from some original implementations to ensure fairness (e.g.. by imposing common encoders and readout across models), so care should be taken when comparing results outside this environment. See also \cref{app:cwn_protocol} for a detailed discussion of the CWN baseline's protocol discrepancy with its original evaluation.}
The experimental setup is designed to systematically address the following research questions:


\vspace{-4pt}
\begin{enumerate}[label=\bfseries Q\arabic*., leftmargin=2em]
    \item \label{q:performance_tnn} Does HOPSE match or exceed the predictive performance of state-of-the-art HOMP networks on standard and fine-grained molecular benchmarks? 
    \item \label{q:performance_topology} Can a message-passing-free architecture effectively capture complex, pure topological properties—such as Betti numbers and orientability—on synthetic benchmarking tasks? 
    
    \item \label{q:accelerate} To what extent does the reduced complexity of HOPSE, relative to HOMP-based methods, yield practical improvements in execution cost and scalability on standard TDL benchmarks?

    \item \label{q:ablations} How do the choices of neighborhood configurations ($\gN_{C}$) and specific positional/structural encodings influence the framework’s performance across different topological domains?
    
\end{enumerate}

\subsection{Evaluation setup}

\paragraph{Datasets.}\label{sec:subsec_datasets}
We evaluate the framework on standard TDL benchmarks~\citep{telyatnikov2025topobench} and molecular property prediction tasks. For inductive graph benchmarks, we use the \textsc{TUDataset} collection---\textsc{MUTAG}, \textsc{PROTEINS}, \textsc{NCI1}, and \textsc{NCI109}~\citep{morris2020}---to ensure consistent comparisons with prior work, despite recent critiques of their benchmarking utility~\citep{coupette2025no}. 
Additionally, motivated by the suggestions of \citet{bechler-speicher2025position}, 
we include four clinically relevant molecular prediction datasets from \cite{huang2022artificial}: \textsc{BBB}, \textsc{CYP3A4}, \textsc{Cl.Hep.}, and \textsc{Caco2}. All graph datasets are transformed into higher-order domains via standard cycle-based (cell) and clique complex (simplicial) liftings.
Finally, we use the \textsc{MANTRA} family~\citep{ballester2025mantra} for evaluating purely topological tasks. Using its 43,000+ manifold triangulations, we focus on three inductive simplicial tasks: predicting homeomorphism type (\textsc{NAME}), orientability (\textsc{ORIENT}), and Betti numbers ($\beta_1,\beta_2$). See \cref{app:extended_results} for further details.

\paragraph{Baselines and Models.}
We evaluate against standard GNNs (\textsc{GCN}~\citep{kipf2017semi}, \textsc{GAT}~\citep{velickovic2017graph}, \textsc{GIN}~\citep{xu2018powerful}) and state-of-the-art HOMP networks: \textsc{SCCNN}~\citep{yang2025hodge} and \textsc{SANN}~\citep{gurugubelli2024sann} for simplicial complexes; \textsc{CWN}~\citep{bodnar2021weisfeiler} and \textsc{CCCN}~\citep{mi2002learning} for cell complexes; and \textsc{TopoTune}~\citep{papillon2024topotune} for both. We compare these to our framework's realizations introduced in \cref{sec:realization_of_framework}: \textbf{HOPSE-M} (manual PSEs) and \textbf{HOPSE-GPSE} (pre-trained neural encodings). The manual variants are further subdivided into connectivity-only (\textbf{HOPSE-M-C}, $g_1$--$g_4$) and feature-aware (\textbf{HOPSE-M-F}, $g_5$--$g_7$) PSEs. Full details of the performed hyperparameter search can be found in \cref{app:hyperparameter_search}.


\paragraph{Training and Evaluation.}
To ensure fair comparisons, five stratified splits with a 50/25/25 train/validation/test ratio are considered for each dataset and task. The optimal hyperparameter configuration is selected based on the average validation performance across five random seeds, and we report the corresponding mean and standard deviation over the test set. Evaluation metrics correspond to the respective task optimization directions ($\uparrow$ for accuracy/F1, $\downarrow$ for error). 

\subsection{Experimental Results}

\definecolor{stdblue}{HTML}{C9DAF8}
\definecolor{bestgray}{HTML}{D9D9D9}
\begin{table}[t]
\caption{\textbf{Overall Results Table.} Evaluation of predictive performance on molecular and topological benchmarks. We report the test mean $\pm$ standard deviation over 5 seeds for best validation configurations. \protect\colorbox{bestgray}{\textbf{Bold}} highlights the top-performing model per column, while \protect\colorbox{stdblue}{blue} denotes models with no statistically significant difference from the best score (at a 95\,\% confidence level).}
\label{tbl:hopse_wandb_graph_sim_compact}
\centering
\begin{adjustbox}{width=1.\textwidth}
\renewcommand{\arraystretch}{1.4}
\begin{tabular}{@{}llcccccccccccc@{}}
\toprule
  &  & \multicolumn{8}{c}{\mbox{Graph (Molecular)}} & \multicolumn{4}{c}{\mbox{Simplicial (\texttt{MANTRA})}} \\
\cmidrule(lr){3-10} \cmidrule(lr){11-14}
 & \textbf{Model} & \scriptsize MUTAG ($\uparrow$) & \scriptsize PROTEINS ($\uparrow$) & \scriptsize NCI1 ($\uparrow$) & \scriptsize NCI109 ($\uparrow$) & \scriptsize BBB ($\uparrow$) & \scriptsize CYP3A4 ($\uparrow$) & \scriptsize Cl.Hep. ($\downarrow$) & \scriptsize Caco2 ($\downarrow$) & \scriptsize NAME ($\uparrow$) & \scriptsize ORIENT ($\uparrow$) & \scriptsize $\beta_1$ ($\uparrow$) & \scriptsize $\beta_2$ ($\uparrow$) \\
\midrule
\multirow{3}{*}{\rotatebox[origin=c]{90}{\textbf{Graph}}} & GCN & \cellcolor{stdblue}{\scriptsize $85.11 \pm 5.55$} & \cellcolor{stdblue}{\scriptsize $75.13 \pm 2.47$} & {\scriptsize $74.69 \pm 1.90$} & {\scriptsize $73.34 \pm 1.11$} & {\scriptsize $79.61 \pm 1.87$} & {\scriptsize $78.68 \pm 0.10$} & {\scriptsize $34.76 \pm 0.03$} & {\scriptsize $0.81 \pm 0.14$} & {\scriptsize $22.80 \pm 3.18$} & {\scriptsize $47.94 \pm 0.01$} & {\scriptsize $36.67 \pm 10.87$} & {\scriptsize $48.09 \pm 0.31$} \\
 & GAT & \cellcolor{stdblue}{\scriptsize $85.11 \pm 4.66$} & \cellcolor{stdblue}{\scriptsize $73.48 \pm 3.02$} & {\scriptsize $74.94 \pm 2.05$} & {\scriptsize $74.62 \pm 1.17$} & {\scriptsize $83.40 \pm 0.12$} & {\scriptsize $77.19 \pm 0.15$} & {\scriptsize $36.54 \pm 1.22$} & {\scriptsize $0.65 \pm 0.01$} & {\scriptsize $18.38 \pm 1.17$} & {\scriptsize $47.94 \pm 0.01$} & {\scriptsize $7.36 \pm 0.16$} & {\scriptsize $47.94 \pm 0.01$} \\
 & GIN & \cellcolor{stdblue}{\scriptsize $83.83 \pm 3.95$} & \cellcolor{stdblue}{\scriptsize $75.84 \pm 1.82$} & {\scriptsize $73.83 \pm 1.13$} & {\scriptsize $73.07 \pm 1.25$} & {\scriptsize $82.91 \pm 0.30$} & \cellcolor{stdblue}{\scriptsize $78.82 \pm 0.73$} & \cellcolor{stdblue}{\scriptsize $33.93 \pm 0.60$} & {\scriptsize $0.88 \pm 0.01$} & {\scriptsize $76.87 \pm 1.01$} & {\scriptsize $56.29 \pm 0.41$} & {\scriptsize $88.13 \pm 0.01$} & {\scriptsize $56.11 \pm 0.53$} \\
\midrule
\multirow{4}{*}{\rotatebox[origin=c]{90}{\textbf{Simplicial}}} & TopoTune & \cellcolor{stdblue}{\scriptsize $84.26 \pm 8.14$} & \cellcolor{stdblue}{\scriptsize $75.56 \pm 3.09$} & \cellcolor{stdblue}{\scriptsize $77.45 \pm 1.07$} & \cellcolor{stdblue}{\scriptsize $76.44 \pm 0.98$} & \cellcolor{stdblue}{\scriptsize $85.07 \pm 1.06$} & {\scriptsize $78.82 \pm 0.16$} & {\scriptsize $35.70 \pm 0.88$} & {\scriptsize $0.76 \pm 0.03$} & {\scriptsize $87.95 \pm 0.57$} & {\scriptsize $75.43 \pm 1.10$} & {\scriptsize $89.77 \pm 0.27$} & {\scriptsize $80.37 \pm 1.62$} \\
 & SCCNN & {\scriptsize $74.47 \pm 3.01$} & {\cellcolor{bestgray}\scriptsize\boldmath $77.06 \pm 2.04$} & {\scriptsize $71.40 \pm 1.45$} & {\scriptsize $71.87 \pm 0.93$} & {\scriptsize $82.96 \pm 0.55$} & {\scriptsize $77.29 \pm 0.08$} & {\scriptsize $36.40 \pm 0.01$} & {\scriptsize $1.04 \pm 0.01$} & {\cellcolor{bestgray}\scriptsize\boldmath $95.43 \pm 0.44$} & {\cellcolor{bestgray}\scriptsize\boldmath $84.93 \pm 0.60$} & {\scriptsize $89.66 \pm 0.10$} & {\scriptsize $79.62 \pm 1.23$} \\
 & SANN & \cellcolor{stdblue}{\scriptsize $78.72 \pm 8.82$} & \cellcolor{stdblue}{\scriptsize $75.84 \pm 3.01$} & {\scriptsize $69.03 \pm 1.32$} & {\scriptsize $69.10 \pm 1.52$} & {\scriptsize $84.33 \pm 0.43$} & {\scriptsize $77.79 \pm 0.24$} & {\scriptsize $34.31 \pm 0.35$} & {\cellcolor{bestgray}\scriptsize\boldmath $0.33 \pm 0.01$} & {\scriptsize $78.30 \pm 1.40$} & {\scriptsize $58.49 \pm 1.72$} & {\scriptsize $88.13 \pm 0.01$} & {\scriptsize $56.38 \pm 0.01$} \\
 & \textbf{HOPSE-M-F} & {\scriptsize $77.45 \pm 3.46$} & \cellcolor{stdblue}{\scriptsize $75.20 \pm 2.08$} & {\scriptsize $71.85 \pm 0.68$} & {\scriptsize $70.98 \pm 1.51$} & \cellcolor{stdblue}{\scriptsize $84.24 \pm 1.10$} & {\scriptsize $76.09 \pm 0.50$} & {\scriptsize $34.94 \pm 0.64$} & \cellcolor{stdblue}{\scriptsize $0.34 \pm 0.01$} & {\scriptsize $76.11 \pm 0.06$} & {\scriptsize $66.58 \pm 0.50$} & {\scriptsize $88.13 \pm 0.01$} & {\scriptsize $56.38 \pm 0.01$} \\
 & \textbf{HOPSE-M-C} & \cellcolor{stdblue}{\scriptsize $86.38 \pm 3.95$} & \cellcolor{stdblue}{\scriptsize $74.84 \pm 2.52$} & {\scriptsize $72.90 \pm 1.03$} & {\scriptsize $72.02 \pm 0.86$} & {\scriptsize $82.96 \pm 1.25$} & {\scriptsize $75.00 \pm 0.65$} & {\scriptsize $34.47 \pm 0.25$} & {\scriptsize $0.37 \pm 0.01$} & {\scriptsize $93.96 \pm 0.29$} & \cellcolor{stdblue}{\scriptsize $83.94 \pm 2.78$} & {\cellcolor{bestgray}\scriptsize\boldmath $90.33 \pm 0.14$} & {\cellcolor{bestgray}\scriptsize\boldmath $83.70 \pm 0.99$} \\
 & \textbf{HOPSE-GPSE} & \cellcolor{stdblue}{\scriptsize $85.96 \pm 5.65$} & \cellcolor{stdblue}{\scriptsize $75.63 \pm 2.07$} & {\scriptsize $76.38 \pm 0.38$} & \cellcolor{stdblue}{\scriptsize $75.41 \pm 1.10$} & \cellcolor{stdblue}{\scriptsize $85.62 \pm 0.82$} & \cellcolor{stdblue}{\scriptsize $79.03 \pm 0.29$} & {\scriptsize $35.45 \pm 0.59$} & \cellcolor{stdblue}{\scriptsize $0.34 \pm 0.01$} & {\scriptsize $81.38 \pm 3.67$} & {\scriptsize $65.32 \pm 5.50$} & {\scriptsize $88.13 \pm 0.01$} & {\scriptsize $57.02 \pm 0.79$} \\
\midrule
\multirow{4}{*}{\rotatebox[origin=c]{90}{\textbf{Cell}}} & TopoTune & \cellcolor{stdblue}{\scriptsize $77.02 \pm 9.36$} & \cellcolor{stdblue}{\scriptsize $76.06 \pm 2.01$} & \cellcolor{stdblue}{\scriptsize $77.32 \pm 1.54$} & \cellcolor{stdblue}{\scriptsize $76.57 \pm 0.68$} & \cellcolor{stdblue}{\scriptsize $84.93 \pm 1.07$} & {\cellcolor{bestgray}\scriptsize\boldmath $79.71 \pm 0.57$} & {\scriptsize $35.40 \pm 0.06$} & {\scriptsize $0.85 \pm 0.01$} & - & - & - & - \\
 & CWN & \cellcolor{stdblue}{\scriptsize $81.70 \pm 7.32$} & \cellcolor{stdblue}{\scriptsize $76.20 \pm 2.05$} & {\scriptsize $73.77 \pm 1.37$} & {\scriptsize $73.86 \pm 1.20$} & {\scriptsize $82.96 \pm 1.24$} & {\scriptsize $76.84 \pm 0.28$} & {\scriptsize $35.49 \pm 0.27$} & {\scriptsize $1.00 \pm 0.01$} & - & - & - & - \\
 & CCCN & {\scriptsize $77.87 \pm 3.95$} & \cellcolor{stdblue}{\scriptsize $73.69 \pm 2.18$} & \cellcolor{stdblue}{\scriptsize $76.73 \pm 1.83$} & \cellcolor{stdblue}{\scriptsize $74.97 \pm 1.03$} & {\scriptsize $83.25 \pm 0.01$} & {\scriptsize $76.24 \pm 0.15$} & {\scriptsize $35.58 \pm 0.09$} & {\scriptsize $0.93 \pm 0.01$} & - & - & - & - \\
 & \textbf{HOPSE-M-F} & \cellcolor{stdblue}{\scriptsize $80.85 \pm 5.55$} & \cellcolor{stdblue}{\scriptsize $76.27 \pm 1.25$} & {\scriptsize $75.35 \pm 1.40$} & \cellcolor{stdblue}{\scriptsize $75.04 \pm 1.76$} & \cellcolor{stdblue}{\scriptsize $85.02 \pm 1.01$} & {\scriptsize $75.43 \pm 0.33$} & \cellcolor{stdblue}{\scriptsize $33.24 \pm 0.24$} & {\scriptsize $0.36 \pm 0.01$} & - & - & - & - \\
 & \textbf{HOPSE-M-C} & {\cellcolor{bestgray}\scriptsize\boldmath $88.09 \pm 2.89$} & \cellcolor{stdblue}{\scriptsize $75.56 \pm 2.06$} & {\scriptsize $73.93 \pm 1.87$} & {\scriptsize $72.84 \pm 0.87$} & {\scriptsize $83.79 \pm 0.97$} & {\scriptsize $76.37 \pm 0.26$} & {\scriptsize $35.73 \pm 1.01$} & {\scriptsize $0.38 \pm 0.01$} & - & - & - & - \\
 & \textbf{HOPSE-GPSE} & \cellcolor{stdblue}{\scriptsize $85.53 \pm 4.13$} & \cellcolor{stdblue}{\scriptsize $76.06 \pm 2.02$} & {\cellcolor{bestgray}\scriptsize\boldmath $78.64 \pm 1.31$} & {\cellcolor{bestgray}\scriptsize\boldmath $76.94 \pm 1.04$} & {\cellcolor{bestgray}\scriptsize\boldmath $86.11 \pm 0.86$} & \cellcolor{stdblue}{\scriptsize $78.91 \pm 0.64$} & {\cellcolor{bestgray}\scriptsize\boldmath $33.07 \pm 0.43$} & {\scriptsize $0.36 \pm 0.01$} & - & - & - & - \\
\bottomrule
\end{tabular}
\end{adjustbox}
\end{table}


\paragraph{Overall Performance Comparison.} \cref{tbl:hopse_wandb_graph_sim_compact} shows all three proposed \textsc{HOPSE} models consistently rival or exceed robust baselines, including complex HOMP methods, answering \ref{q:performance_tnn} and \ref{q:performance_topology} affirmatively. Importantly, for a fair comparison, GNN baselines are evaluated both with and without the same encodings used by HOPSE, with the best-performing configuration reported (see \cref{tbl:gnn_ablation}). On pure topological tasks, \textbf{HOPSE-M-C} achieves state-of-the-art results on $\beta_1$ and $\beta_2$. For \textsc{ORIENT}, it performs within the 95\,\% confidence interval of the top-performing \textsc{SCCNN} while being a close second on \textsc{NAME}, effectively matching it without the overhead of simplicial message passing. 
On inductive graph and molecular benchmarks, \textsc{HOPSE} remains highly competitive. \textbf{HOPSE-GPSE} (Cell) achieves the best performance on \textsc{Cl.Hep.}, \textsc{BBB}, \textsc{NCI1}, and \textsc{NCI109}. Furthermore, \textbf{HOPSE-M-F} and \textbf{HOPSE-GPSE} (Cell) match the top-performing \textsc{SANN} on \textsc{Caco2} (Simplicial). 
Overall, our proposed model delivers top-tier topological expressivity and versatile molecular performance. 

\paragraph{Time Comparison.} 
While HOPSE introduces a one-time initial overhead to extract structural and positional encodings (\cref{tbl:preprocess_ablation_preproc_time}), this cost is heavily offset during the training phase. By bypassing the computationally expensive topological neighborhood aggregations required by complex HOMP methods (e.g. SCCNN, CWN), the proposed models achieve remarkable per-epoch efficiency. As shown in \cref{tbl:best_rerun_epoch_time}, both \textbf{HOPSE-M} and \textbf{HOPSE-GPSE} consistently execute significantly faster per epoch compared to these state-of-the-art HOMP baselines across both simplicial (SCCNN) and cell domains (CCCN, CWN). Consequently, when evaluating the total end-to-end training time (\cref{tbl:best_rerun_runtime}), HOPSE models consistently deliver the fastest overall computation. In nearly all evaluated datasets across the cell and simplicial domains, HOPSE variants achieve the lowest mean runtime or perform within the 95\,\% confidence interval of the fastest model. This confirms that HOPSE provides training speedups over competitive topological baselines while maintaining top-tier predictive accuracy (\ref{q:accelerate}). A more granular analysis of how HOPSE scales as the number of higher-order components within the dataset grows is provided in Appendix~\ref{app:scalability}, confirming that our efficiency advantage becomes even more pronounced as structural complexity increases.

\paragraph{Parameter Count and Architectural Tradeoffs.} 
\cref{tbl:model_parameters} details the number of parameters across all model configurations, highlighting a core architectural tradeoff inherent to the framework. By substituting computationally expensive topological neighborhood aggregations with rich structural and positional encodings, \textsc{HOPSE} relies on more expressive, parameter-dense MLPs to process these representations. Consequently, in the simplicial domain, variants like \textbf{HOPSE-M-C} exhibit higher parameter counts compared to certain HOMP baselines such as \textsc{SCCNN}. Conversely, in the cell domain, HOPSE maintains highly competitive or even lower parameter footprints compared to models like \textsc{CWN}. Ultimately, this demonstrates that \textsc{HOPSE} effectively trades a modest increase in parameter capacity for its substantial gains in execution speed and scalability (\ref{q:accelerate}).



\begin{wraptable}{r}{0.46\textwidth}
    \centering
    \vspace{-13pt} 
    \caption{\textbf{Taxonomy of neighborhoods considered in our evaluation.}}
    \label{tbl:neighborhood_configurations}
    \renewcommand{\arraystretch}{1.2}
    \footnotesize
    \begin{tabular}{@{}ll@{}}
        \toprule
        \textbf{Group} & \textbf{Included Neighborhoods} \\
        \midrule
        \textit{Adj-1} & $\{ \mathcal{A}_{0,1} \}$ \\
        \textit{Adj-2} & $\{ \mathcal{A}_{0,1}, \mathcal{A}_{0,2} \}$ \\
        \textit{Adj-3} & $\{ \mathcal{A}_{0,1}, \mathcal{A}_{1,2}, \mathcal{A}_{0,2}, \mathcal{A}_{1,0}, \mathcal{A}_{2,1}, \mathcal{A}_{2,0} \}$ \\
        \midrule
        \textit{Inc-1} & $\{ \mathcal{I}_{0 \to 1}, \mathcal{I}_{0 \to 2} \}$ \\
        \textit{Inc-2} & $\{ \mathcal{I}_{0 \to 1}, \mathcal{I}_{1 \to 2}, \mathcal{I}_{0 \to 2}, \mathcal{I}_{1 \to 0}, \mathcal{I}_{2 \to 1}, \mathcal{I}_{2 \to 0} \}$ \\
        \bottomrule
    \end{tabular}
    \vspace{-10pt} 
\end{wraptable}
\paragraph{Choice of $\gN_{C}$.} 
The selection of the neighborhood function $\gN_{C}$ is a critical design choice controlling the expressivity of the HOPSE framework. As detailed in \cref{tbl:neighborhood_configurations}, we evaluate adjacency-based (\textit{Adj-1} to \textit{Adj-3}) and incidence-based (\textit{Inc-1}, \textit{Inc-2}) configurations of increasing complexity. Our results, reported in \cref{tbl:neighborhood_ablation}, reveal a clear architectural trend: the most expansive configurations—particularly \textit{Adj-3} and \textit{Inc-2}, which aggregate up and down relations across dimensions including 2nd-order rank jumps—are essential for maximizing performance on pure topological tasks such as \textsc{NAME}, \textsc{ORIENT}, and Betti number predictions. Conversely, for standard inductive graph and molecular regression tasks, simpler configurations like \textit{Adj-1} (utilizing only graph-based adjacency) and \textit{Adj-2} often perform competitively, acting as an effective regularizer while maintaining high computational efficiency (\ref{q:ablations}).

\paragraph{Ablation of structural and positional encodings.} 
To isolate the contribution of different encodings, \cref{tbl:ablation_encodings} evaluates each encoding individually across varying neighborhood configurations ($\gN_C$). The results reveal a dataset-dependent behavior rather than a single universal encoding. For standard graph and molecular regression benchmarks, \textsc{RWSE} and \textsc{ElectrostaticPE} prove highly effective, frequently achieving the best performance and maintaining stability across datasets such as \textsc{MUTAG} and \textsc{CYP3A4}. Conversely, pure topological tasks in the simplicial domain exhibit a strict reliance on spectral properties. In these scenarios, \textsc{LapPE} decisively outperform all other alternatives, unlocking maximum expressivity when paired with the expansive $\emph{Adj-2}$ neighborhood. Ultimately, this ablation demonstrates that the modularity of the \textsc{HOPSE} framework is a key strength, allowing the encoding strategy to be precisely tailored to the structural priors of the target task (\ref{q:ablations}). Last, but not least, we further test whether HOMP baselines benefit from augmenting their initial features with \textsc{CombinedPE} encodings (\cref{app:homp_hasse_ablation}). Gains are dataset-dependent rather than systematic: PSEs improve the mean in only 13 of 40 model--dataset pairs, and the best method per column in \cref{tbl:hopse_wandb_graph_sim_compact} is unchanged. These results suggest that HOPSE's advantage stems from its architecture, not merely the encodings. 

\section{Concluding remarks}
\label{sec:conclusions}

    

This work introduces \textsc{HOPSE}, a \emph{scalable} framework designed to extract and leverage higher-order structural and positional information for \emph{arbitrary combinatorial representations} inspired by previous works regarding expressivity \citep{gurugubelli2024sann}. 
Theoretically, \textsc{HOPSE} matches the expressivity of state-of-the-art TNNs while avoiding the HOMP paradigm, thereby substantially reducing the computational cost of training and inference. Moreover, training and inference scale linearly with dataset size. 
These claims are supported by experiments on molecular graph benchmarks as well as on pure higher-order topological tasks, with both instantiations--\textbf{HOPSE-M} and \textbf{HOPSE-GPSE}--matching or surpassing state-of-the-art performance while achieving consistent speed-ups over HOMP-based models. By offering a compelling trade-off between scalability and expressivity, this work not only enables broader applicability of TDL but also facilitates the integration of higher-order interactions into other areas of deep learning--mirroring the widespread adoption of graph-based methods~\citep{zhou2023improving,wang2025dynamic,han2022vision,han2023vision}.

\paragraph{Limitations.} 
While \textsc{HOPSE} offers significant scalability advantages during training and inference, a few practical considerations remain. First, computing exact manual encodings (e.g., \textsc{LapPE}) can be computationally expensive for massive datasets, making sparse approximations or scalable neural encoders like \textsc{GPSE} preferable in such regimes. Second, replacing topological message passing with independent MLPs yields a larger parameter footprint than baseline GNNs. Third, while our expressivity guarantees assume strict injectivity, practical instantiations utilizing standard pooling or bounded MLPs inherently approximate these conditions. Fourth, spectral PSEs such as \textsc{LapPE} require symmetric operators, so Hasse graphs are made undirected before encoding; reciprocal incidences (e.g., $\mathcal{I}_{0\to1}$ and $\mathcal{I}_{1\to0}$) thus collapse to the same graph and lose their directionality, an effect we do not quantify here, but which direction-aware PSEs could readily address within our framework. Lastly, our strict separation over CCWL is established analytically, as no controlled expressivity benchmark for combinatorial complexes exists yet---an open challenge for the TDL community. 

\paragraph{Future Work.} Looking forward, future work will explore deeper integrations of these encodings into standard TNNs or transformer-based architectures to see if they provide further benefits. Finally, while \textbf{HOPSE-GPSE} currently utilizes a frozen pretrained encoder for efficiency, exploring joint end-to-end training remains a promising avenue to unlock further task-specific performance gains.

\section*{Author contributions}

G.B., M.M., and L.V.L. led the final code implementation, writing of the manuscript, and design, execution, and analysis of the final experiments. L.T. and G.B. conceived the research idea. In the earlier stages of the project, L.T. and M.C. led the code implementation and the execution of experiments, as well as drafted the initial manuscript text. In the final stages, L.T. contributed to the final experimental design and setup. A.A., M.C., and M.M. contributed to the theoretical results, with A.A. leading the main theoretical contributions. L.C. and M.P. contributed to the code implementation, final experimental design, and manuscript editing. P.B.-R. and N.M. provided project support. Additionally, N.M. provided infrastructure and contributed to writing. L.T. and G.B. supervised and coordinated the project together.


\newpage










\begin{ack}
G. B., L. C., M. P., and N. M. acknowledge partial support from NSF grant 2602079 and the Noyce foundation. 
G.B. also acknowledges partial support from Arlequin AI. 
L. C. also acknowledges partial support from an NSF GRFP and the Chan Zuckerberg Foundation. 
M. P. also acknowledges support from NSF Career grant 2240158 and the National Science and Engineering Research Council of Canada. 
A. A. acknowledges the financial support of the Finnish Ministry of Education and Culture through the Quantum Doctoral Education Pilot Program (QDOC VN/3137/2024-OKM-4) and the Research Council of Finland through the Finnish Quantum Flagship project (358877, Aalto University). 
L.V.L and P.B. acknowledge support from Grant PCI2023-145974-2, funded by MICIU/AEI/10.13039/501100011033 and cofunded by the European Union (GRAPHS4SEC project), as well as from the I+D+i project titled BLOSSOMS, grant PID2024-158530OB-I00, funded by MICIU/AEI/10.13039/501100011033/ and by ERDF/EU. 
P.B. is also supported by the Catalan Institution for Research and Advanced Studies (ICREA Academia). 
L.V.L. is also supported by the predoctoral program AGAUR-FI ajuts (2025 FI-3 00065) Joan Oro, of the Department of Research and Universities of the Generalitat of Catalonia, as well as the European Social Plus Fund.
M. M. acknowledges support from Sapienza grant RG123188B3EF6A80 (CENTS).
\end{ack}

\bibliographystyle{plainnat}
\bibliography{biblio}  

\appendix


\newpage

\appendix
\newpage
\section{Extended background}
\label{app:extended_background}
\subsection{Topological domains} 
\label{appendix:topological_domains}

This section draws primarily from Appendix A of \citet{telyatnikov2025topobench}. It begins with the fundamental concept of featured graphs, which forms the basis for understanding more intricate structures. Then it proceeds to explore higher-order domains --- including simplicial complexes, cell complexes, and combinatorial complexes --- each providing distinct capabilities for representing various types of relationships and hierarchies within data.

\begin{definition}
\label{def:featured_graph}
Let $\mathcal{G} = (V, E)$ be a graph, with node set $V$ and edge set $E$. A featured graph is a tuple $\mathcal{G}_{F} = (V, E, F_V, F_E)$, where $F_V: V \to \mathbb{R}^{d_{v}}$ is a function that maps each node to a feature vector in $\mathbb{R}^{d_{v}}$ and $F_E: E \to \mathbb{R}^{d_{e}}$ is a function that maps each edge to a feature vector in $\mathbb{R}^{d_{e}}$.
\end{definition}

A topological domain is a graph generalization that captures pairwise and higher-order relationships between entities~\citep{bick2023higher,battiston2021physics}. When working with topological domains, two key properties come into play: set-type relations and hierarchical structures represented by rank functions~\citep{hajij2023tdl, papillon2023architectures}.

\begin{definition}[Set-type relation]
A relation in a topological domain is called a \textit{set-type relation} if another relation in the domain does not imply its existence.
\end{definition}

\begin{definition}[Rank function]
A \textit{rank function} on a higher-order domain \(\mathcal{X}\) is an order-preserving function \( rk \colon \mathcal{X} \to \mathbb{Z}_{\geq 0} \) such that \( x \subseteq y \) implies \( rk(x) \leq rk(y) \) for all \( x, y \in \mathcal{X} \).
\end{definition}

Set-type relations emphasize the independence of connections within a domain, allowing for flexible representation of complex interactions. In contrast, rank functions introduce a hierarchical (also called part-whole) organization that facilitates the representation and analysis of nested relationships.



\paragraph{Simplicial complexes.} Simplicial complexes extend graphs by incorporating hierarchical part-whole relationships through the multi-scale construction of cells. In this structure, nodes correspond to rank $0$-cells, which can be combined to form edges (rank $1$-cells). Edges can then be grouped to form faces (rank 2 cells), and faces can be combined to create volumes (rank $3$-cells), continuing in this manner. Consequently, the faces of a simplicial complex are triangles, volumes are tetrahedra, and higher-dimensional cells follow the same pattern. A key feature of simplicial complexes is their strict hierarchical structure, where each $k$-dimensional simplex is composed of $(k-1)$-dimensional simplices, reinforcing a strong sense of hierarchy across all levels.

\begin{definition}[Simplicial Complex]
A \textit{simplicial complex (SC)} in a non-empty set \(S\) is a pair \(SC = (S, \mathcal{X})\), where \(\mathcal{X} \subset \mathcal{P}(S) \setminus \{\emptyset\}\) satisfies: if \(x \in SC\) and \(y \subseteq x\), then \(y \in SC\). The elements of \(\mathcal{X}\) are called \textit{simplices}.
\end{definition}

\begin{example}[3D Surface Meshes]

3D models of objects, such as those used in computer graphics or for representing anatomical structures, are often constructed using triangular meshes. These meshes naturally form simplicial complexes, where the vertices of the triangles are 0-simplices, the edges are 1-simplices, and the triangular faces themselves are 2-simplices.

\end{example}

\paragraph{Cell complexes.}
Cell complexes provide a hierarchical interior-to-boundary structure, offering clear topological and geometric interpretations, but they are not based on set-type relations. Unlike simplicial complexes, cell complexes are not limited to simplexes; faces can involve more than three nodes, allowing for a more flexible representation. This increased flexibility grants cell complexes greater expressivity compared to simplicial complexes \cite{bodnar2021weisfeiler, bodnar2023topological}.

\begin{definition}[Cell complex]
A \textit{regular cell complex} is a topological space \(S\) partitioned into subspaces (cells) \(\{x_\alpha\}_{\alpha \in P_S}\), where \(P_S\) is an index set, satisfying:
\begin{enumerate}
    \item \(S = \cup_{\alpha \in P_S} \text{int}(x_\alpha)\), where \(\text{int}(x)\) denotes the interior of cell \(x\).
    \item For each \(\alpha \in P_S\), there exists a homeomorphism \(\psi_\alpha\) (\textit{attaching map}) from \(x_\alpha\) to \(\mathbb{R}^{n_\alpha}\) for some \(n_\alpha \in \mathbb{N}\). The integer \(n_\alpha\) is the \textit{dimension} of cell \(x_\alpha\).
    \item For each cell \(x_\alpha\), the boundary \(\partial x_\alpha\) is a union of finitely many cells of strictly lower dimension.
\end{enumerate}
\end{definition}

\begin{example}[Molecular structures.]
    Molecules admit natural representations as cell complexes by considering atoms as nodes (i.e., cells of rank zero), bonds as edges (i.e., cells of rank one), and rings as faces (i.e., cells of rank two).
\end{example}

\paragraph{Combinatorial complexes.} Combinatorial complexes combine hierarchical structure with set-type relations, enabling a flexible yet comprehensive representation of higher-order networks.

\begin{definition}[Combinatorial complex]
A \textit{combinatorial complex (CC)} is a triple \((\mathcal{V}, \mathcal{C}, rk)\) consisting of a set \(\mathcal{V}\), a subset \(\mathcal{C} \subset \mathcal{P}(\mathcal{V}) \setminus \{\emptyset\}\), and a function \(rk \colon \mathcal{C} \to \mathbb{Z}_{\geq 0}\) satisfying:
\begin{enumerate}
    \item For all \(v \in \mathcal{V}\), \(\{v\} \in \mathcal{C}\) and \(rk(v) = 0\).
    \item The function \(rk\) is order-preserving: if \(x, y \in \mathcal{C}\) with \(x \subseteq y\), then \(rk(x) \le rk(y)\).
\end{enumerate}

Moreover, a \emph{featured combinatorial complex} is a 4-tuple $(\mathcal{V}, \mathcal{C}, rk, F)$, where $F: \mathcal{C} \to \mathbb{R}^{d_c}$ is a function that maps each cell to a feature vector in $\mathbb{R}^{d_c}$.

\end{definition}

\begin{example}[Geospatial structures.]
Geospatial data, comprised of grid points (0-cells), road polylines (1-cells), and census tract polygons (2-cells), can be effectively represented using combinatorial complexes. A visual example is provided in Figure 2 (Right) of~\citet{battiloro2024n}.
\end{example}

\smallskip

\paragraph{Featured topological domains.}  
A featured graph is a graph whose nodes or edges are equipped with feature functions~\citep{sanchez2021gentle}. Similarly, each topological element (e.g., simplex or cell) can carry feature vectors to represent some information.

\begin{definition}[Featured topological domain]
A \emph{featured topological domain} is a pair \((\mathcal{X}, F)\), where \(\mathcal{X}\) is a topological domain and \(F = \{F_i\}_{i \geq 0}\) is a collection of feature functions. Each function \(F_i\) maps the \(i\)-dimensional elements of \(\mathcal{X}\), denoted \(\mathcal{X}_i\), to a feature space \(\mathbb{R}^{k_i}\):
\[
F_i \colon \mathcal{X}_i \to \mathbb{R}^{k_i}.
\]
\end{definition}

\subsection{Neighborhood functions and augmented Hasse graphs}
\label{app:background_neighbourhoods}

Discrete topological domains are equipped with a notion of neighborhood among cells that confers on them a topological structure; this section revisits the main neighborhood-related concepts in the general combinatorial complex case.

\begin{definition}[Neighborhood function]
    Given a combinatorial complex $\gT=(\mathcal{V}, \mathcal{C}, \rk)$, a neighborhood function $\mathcal{N}: \mathcal{C} \rightarrow \mathcal{P}(\mathcal{C})$ on a combinatorial complex is a function that assigns to each cell $\sigma \in \mathcal{C}$ a collection of \emph{neighbor cells} $\mathcal{N}(\sigma) \subset \mathcal{C}\cup \emptyset$.
\end{definition}

Each neighborhood function can be used to create a Strictly Augmented Hasse graph.
\begin{definition}[Strictly augmented Hasse graph]
    Each particular neighborhood function $\mathcal{N}$ induces a strictly augmented Hasse graph $\mathcal{G}_{\mathcal{N}} = (\mathcal{C}_\mathcal{N}, \mathcal{E}_\mathcal{N})$ on $\mathcal{T}$ \citep{papillon2024topotune,hajij2023tdl}, defined as the directed graph whose nodes and edges are given, respectively, by $\mathcal{C}_\mathcal{N} =\{\sigma \in \mathcal{C} \, | \, \mathcal{N}(\sigma) \neq \emptyset \}$ and $\mathcal{E}_\mathcal{N} =\{(\tau, \sigma) \, | \, \tau \in \mathcal{N}(\sigma)\}$.
\end{definition}

Examples of neighborhood functions are incidences (connecting cells with different ranks) and adjacencies (connecting cells with the same rank), although other neighborhood functions can be defined for specific tasks \citep{battiloro2024n}. More in detail, given a combinatorial complex $\gT$:

\begin{definition}(Incidences)
     The incidence neighborhood function of a $t$-cell $\sigma\in\mathcal{T}$ with respect to a rank $s$ is defined as 
    \begin{equation}\label{eq:incidences}
    \begin{split}
        \mathcal{I}_{s\to t}(\sigma) =
        \left\{
        \begin{array}{ll}
             \{\tau \in \mathcal{C} \, | \, \textrm{rk}(\tau) = s, \sigma \subset \tau \} & \text{ if $t < s$;} \\
             \{\tau \in \mathcal{C} \, | \, \textrm{rk}(\tau) = s, \tau \subset \sigma\} & \text{ if $t > s$.}
        \end{array}
        \right.
    \end{split}
    \end{equation}
    A $s$-cell $\tau\in\mathcal{T}$ is said to be co-incident of $\sigma$ if it contains $\sigma$ and $s>t$; analogously, a $s$-cell $\tau\in\mathcal{T}$ is incident of $\sigma$ if it is is contained by $\sigma$ and $s<t$.
\end{definition}

Incidences in turn induce adjacency neighborhoods of cells of the same rank as follows:

\begin{definition}[Adjacencies]
    The adjacency neighborhood function of a $t$-cell $\sigma\in\mathcal{T}$ with respect to a rank $s$ is defined as
\begin{equation}\label{eq:adjacencies}
\begin{split}
    \mathcal{A}_{s,t}(\sigma) = \{\tau \in \mathcal{C} \, | \, \textrm{rk}(\tau) = t, \mathcal{I}_{s\to t}(\sigma) \cap \mathcal{I}_{s\to t}(\tau) \neq \emptyset \}
\end{split}
\end{equation}
In this case, two $t$-cells $\sigma$ and $\tau$ are said to be adjacent if they are both contained in a $s$-cell $\delta\in\mathcal{T}$.
\end{definition}

\section{Counting neighborhoods and message passing routes}
\label{app:expanded_problem_formulation_neighbourhoods_and_paths}
\begin{figure}[htbp]

    \centering

    \begin{subfigure}[b]{0.23\textwidth}
        \includegraphics[width=\linewidth]{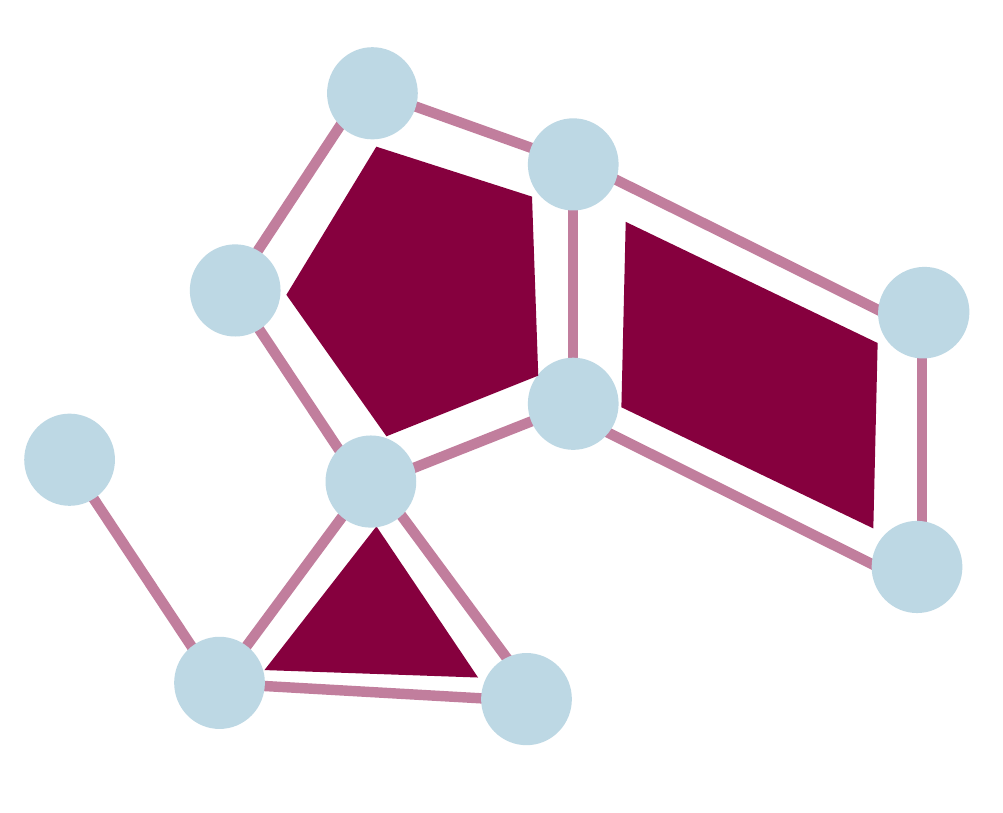}
        \caption{}
    \end{subfigure}

    \vspace{0.8em}

    

    \begin{subfigure}[b]{0.23\textwidth}
        \includegraphics[width=\linewidth]{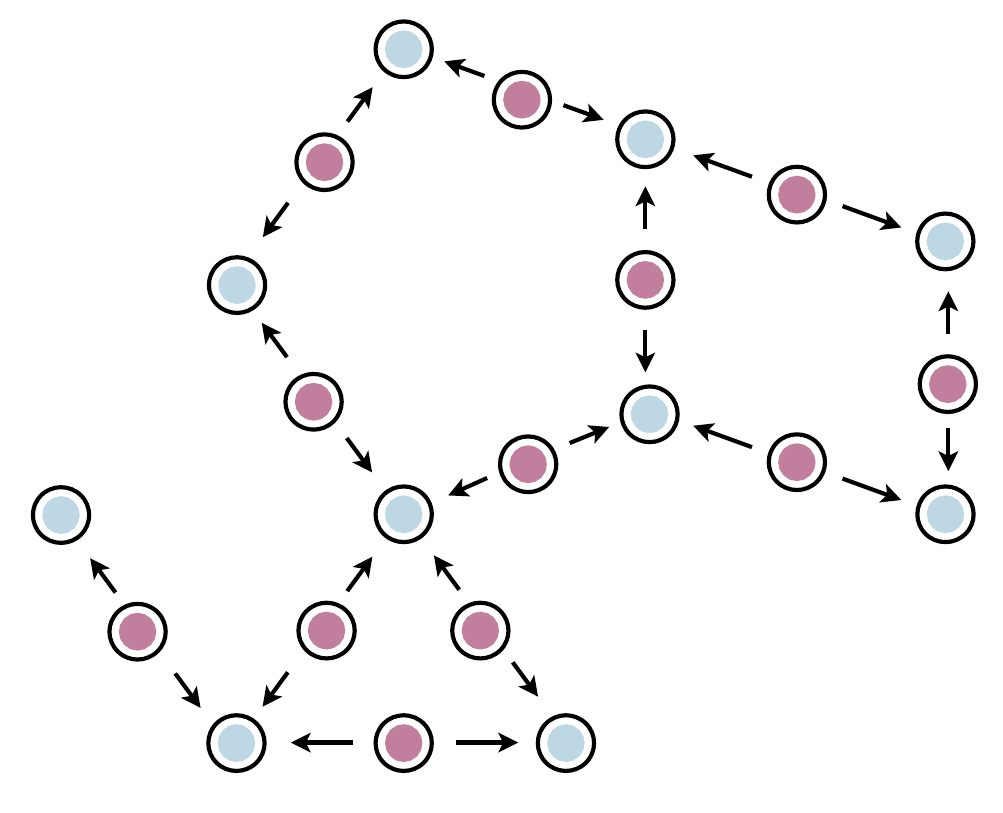}
        \caption{$\gI_{1\to0}$}
    \end{subfigure}
    \begin{subfigure}[b]{0.23\textwidth}
        \includegraphics[width=\linewidth]{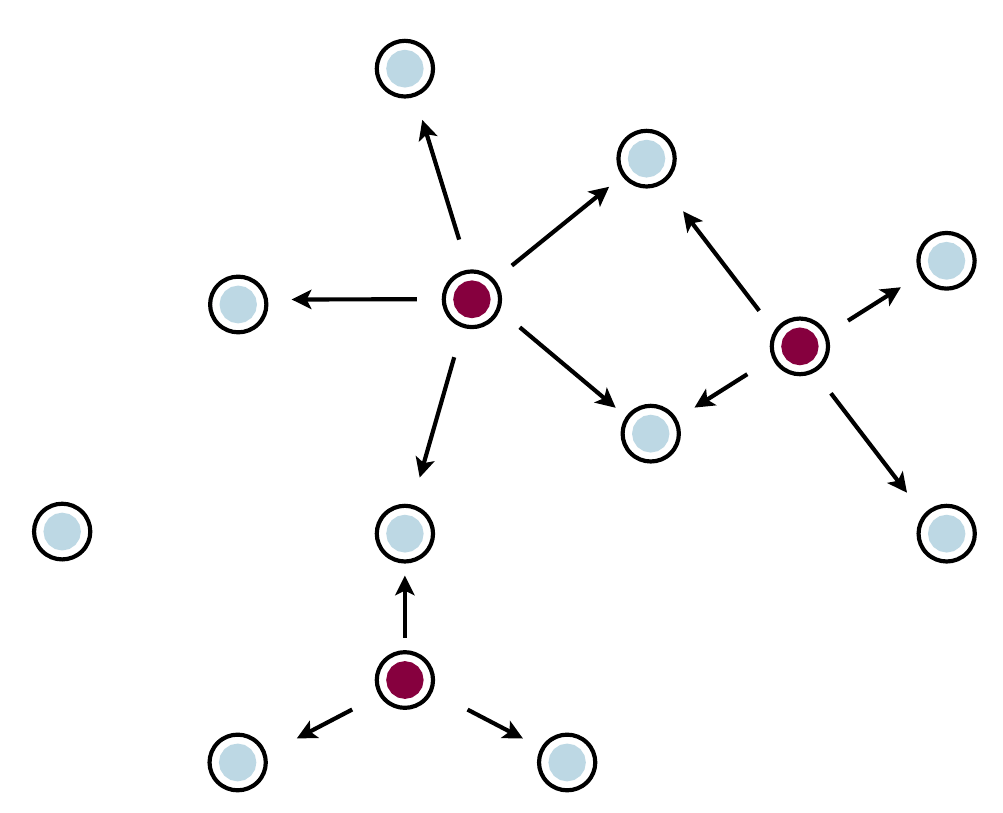}
        \caption{$\gI_{2\to0}$}
    \end{subfigure}
    \begin{subfigure}[b]{0.23\textwidth}
        \includegraphics[width=\linewidth]{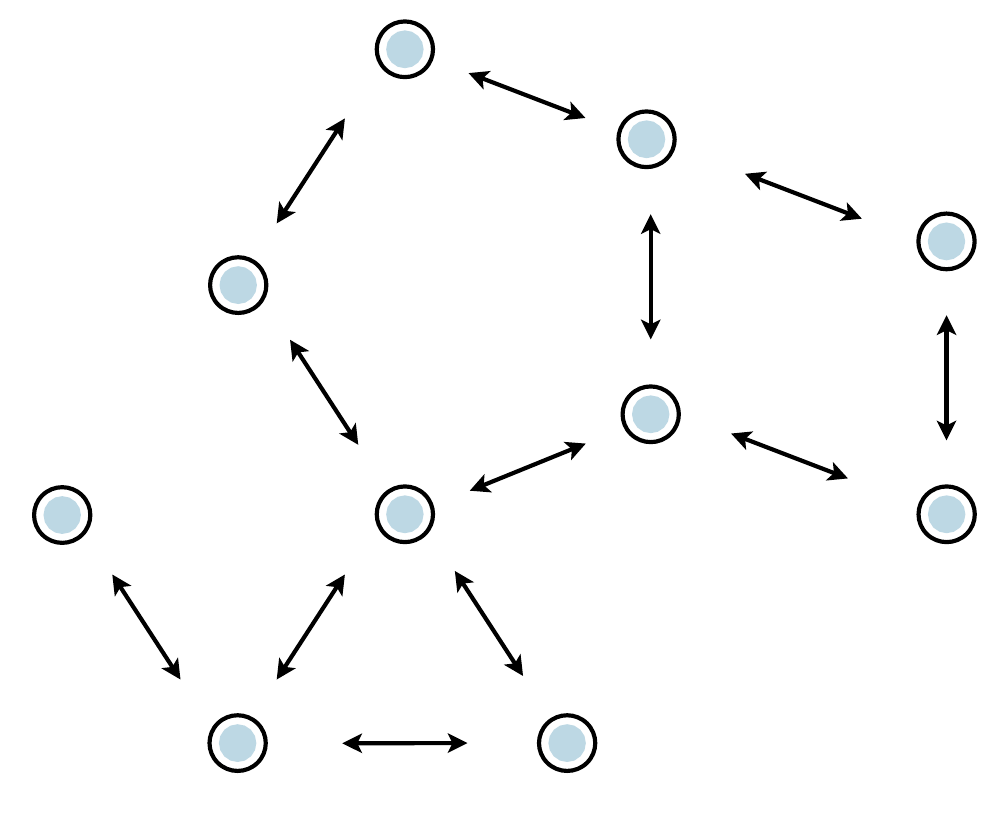}
        \caption{$\gA_{0,1}$}
    \end{subfigure}
    \begin{subfigure}[b]{0.23\textwidth}
        \includegraphics[width=\linewidth]{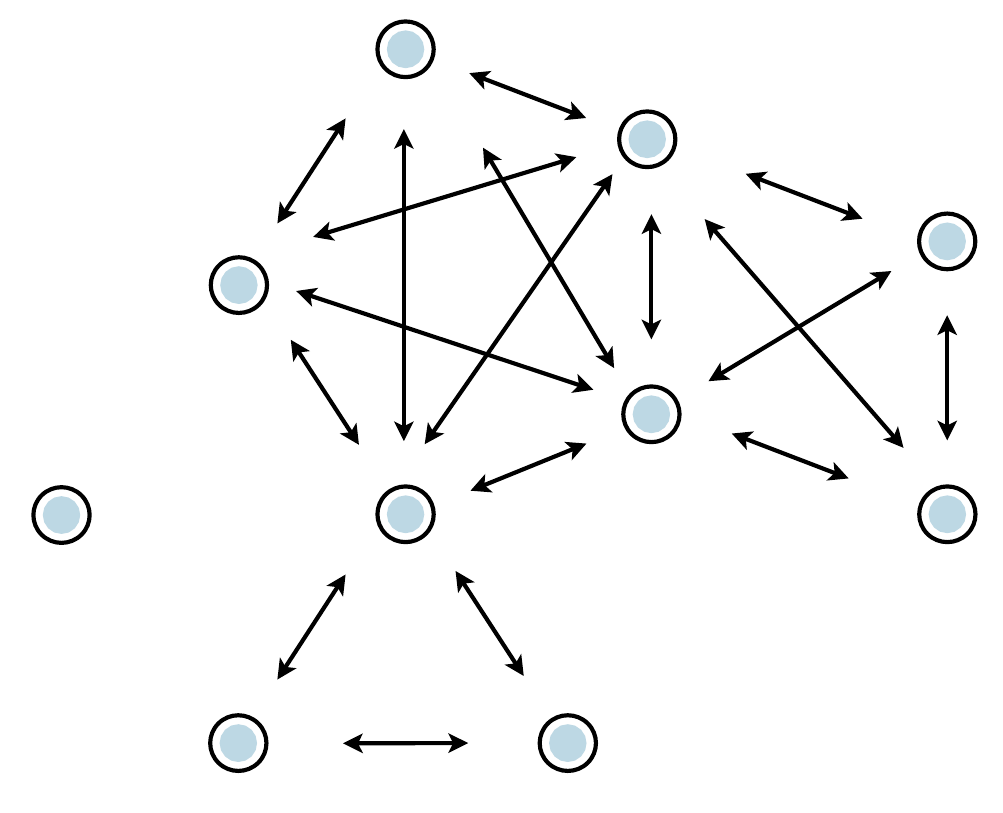}
        \caption{$\gA_{0,2}$}
    \end{subfigure}

    \vspace{0.8em}

    \begin{subfigure}[b]{0.23\textwidth}
        \includegraphics[width=\linewidth]{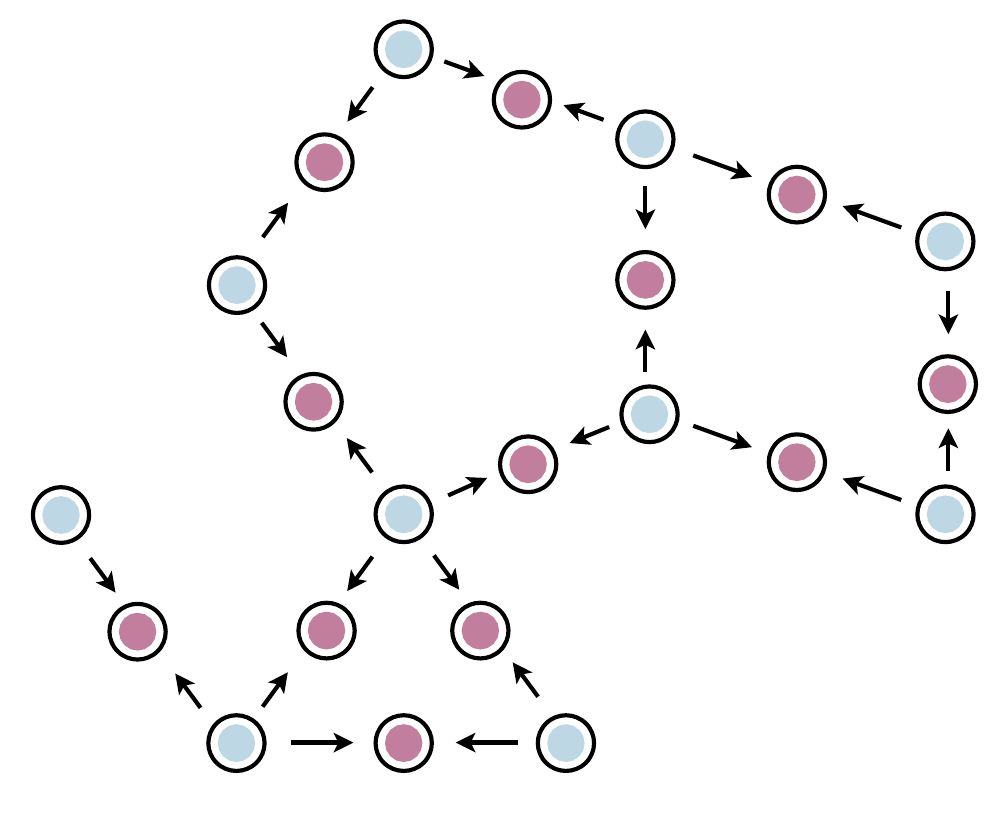}
        \caption{$\gI_{0\to1}$}
    \end{subfigure}
    \begin{subfigure}[b]{0.23\textwidth}
        \includegraphics[width=\linewidth]{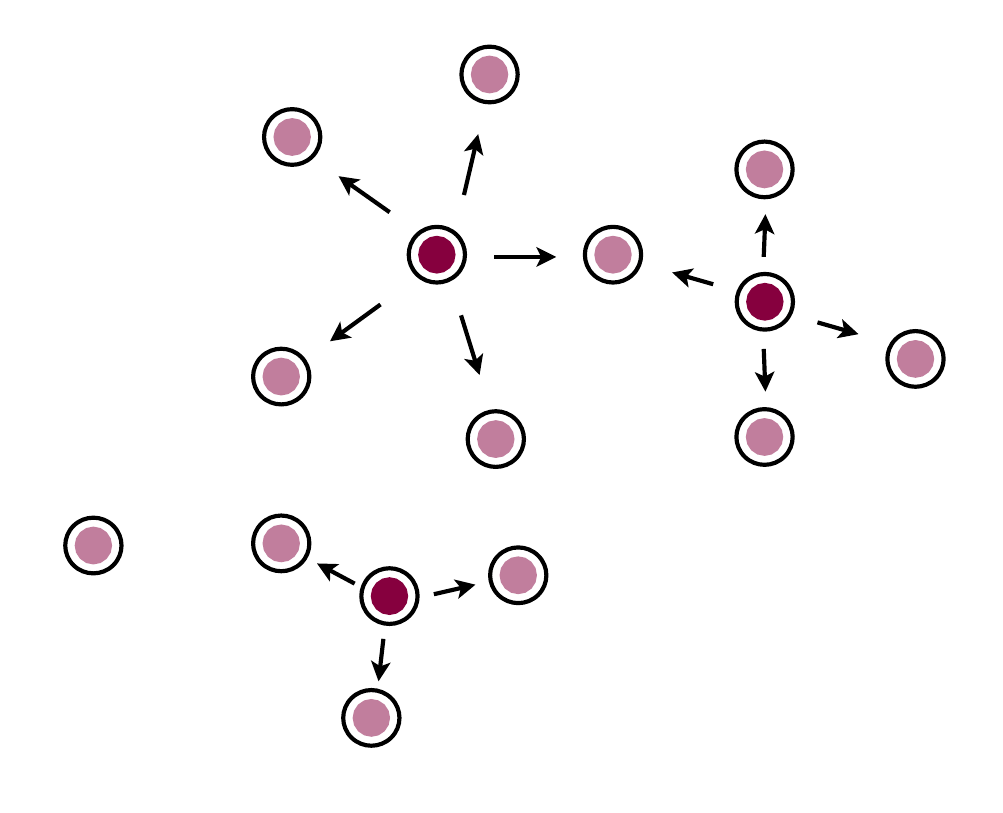}
        \caption{$\gI_{2\to1}$}
    \end{subfigure}
    \begin{subfigure}[b]{0.23\textwidth}
        \includegraphics[width=\linewidth]{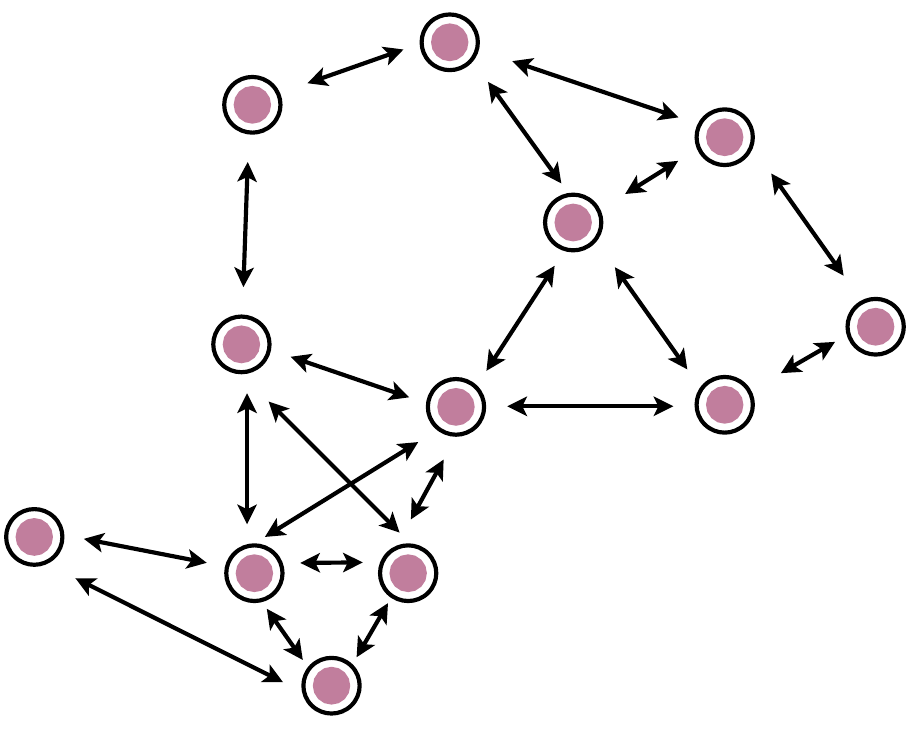}
        \caption{$\gA_{1,0}$}
    \end{subfigure}
    \begin{subfigure}[b]{0.23\textwidth}
        \includegraphics[width=\linewidth]{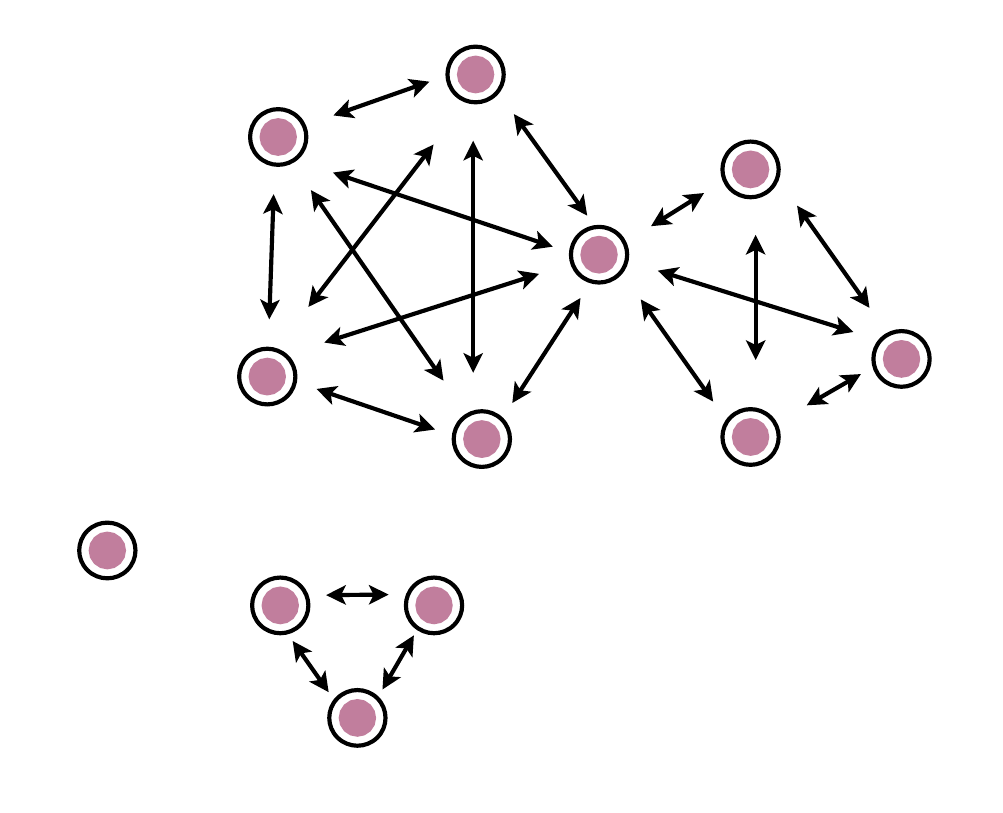}
        \caption{$\gA_{1,2}$}
    \end{subfigure}

    \vspace{0.8em}

    \begin{subfigure}[b]{0.23\textwidth}
        \includegraphics[width=\linewidth]{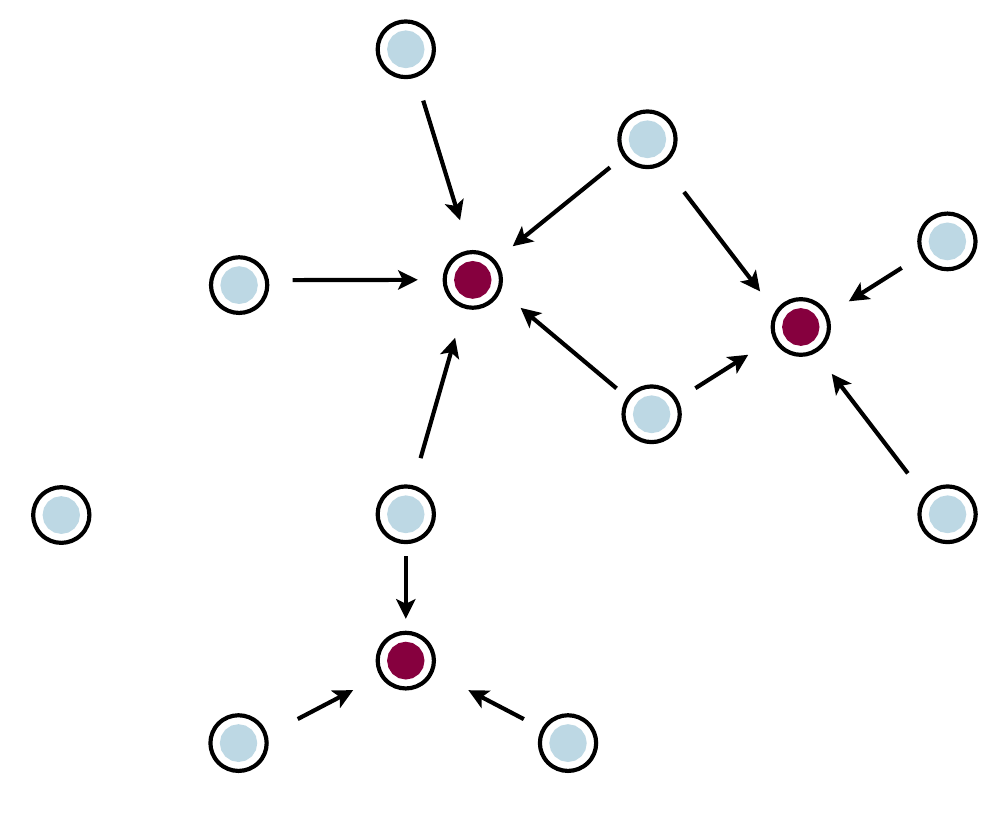}
        \caption{$\gI_{0\to2}$}
    \end{subfigure}
    \begin{subfigure}[b]{0.23\textwidth}
        \includegraphics[width=\linewidth]{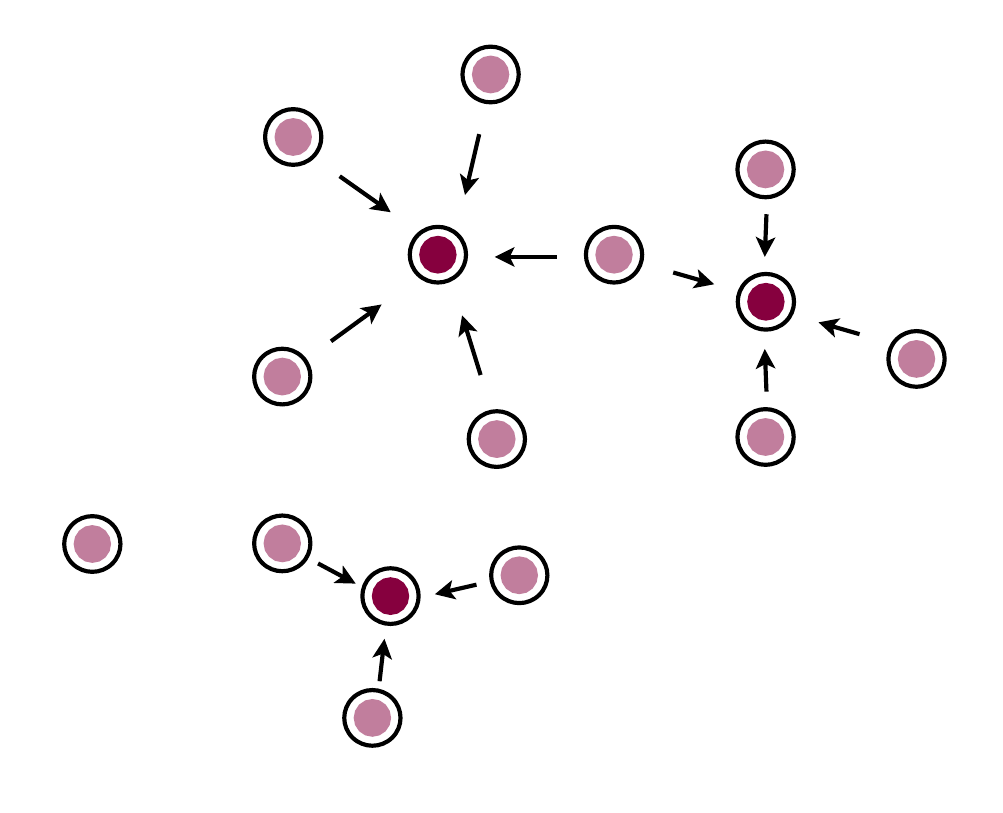}
        \caption{$\gI_{1\to2}$}
    \end{subfigure}
    \begin{subfigure}[b]{0.23\textwidth}
        \includegraphics[width=\linewidth]{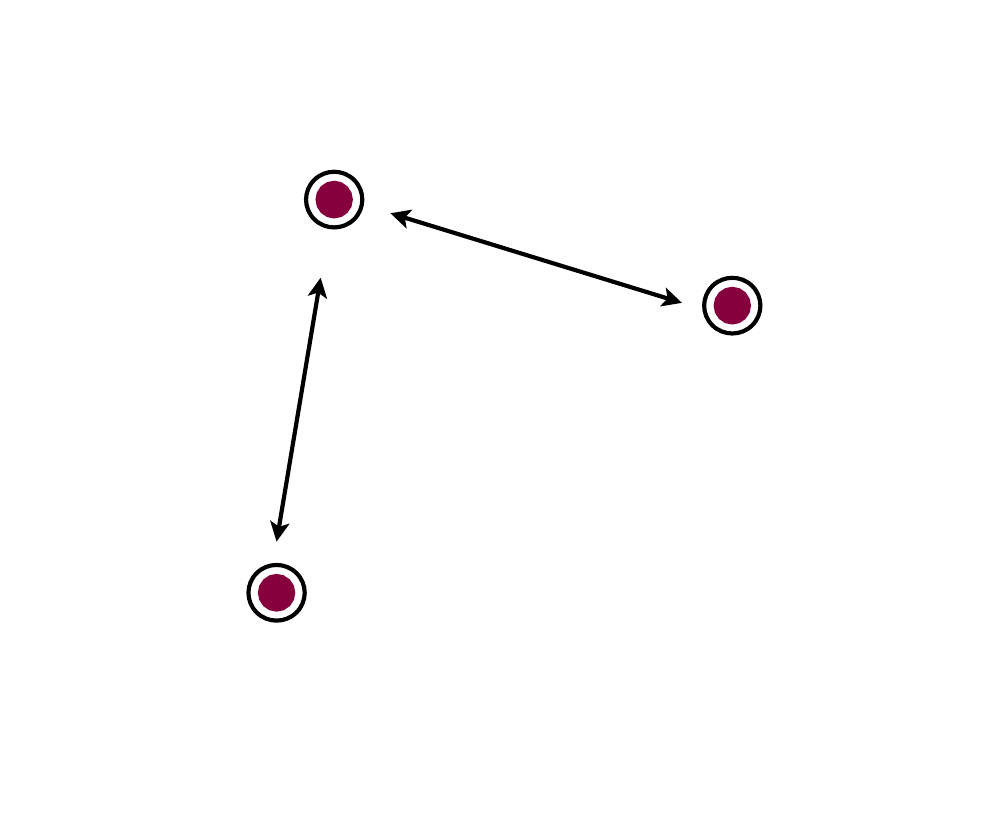}
        \caption{$\gA_{2,0}$}
    \end{subfigure}
    \begin{subfigure}[b]{0.23\textwidth}
        \includegraphics[width=\linewidth]{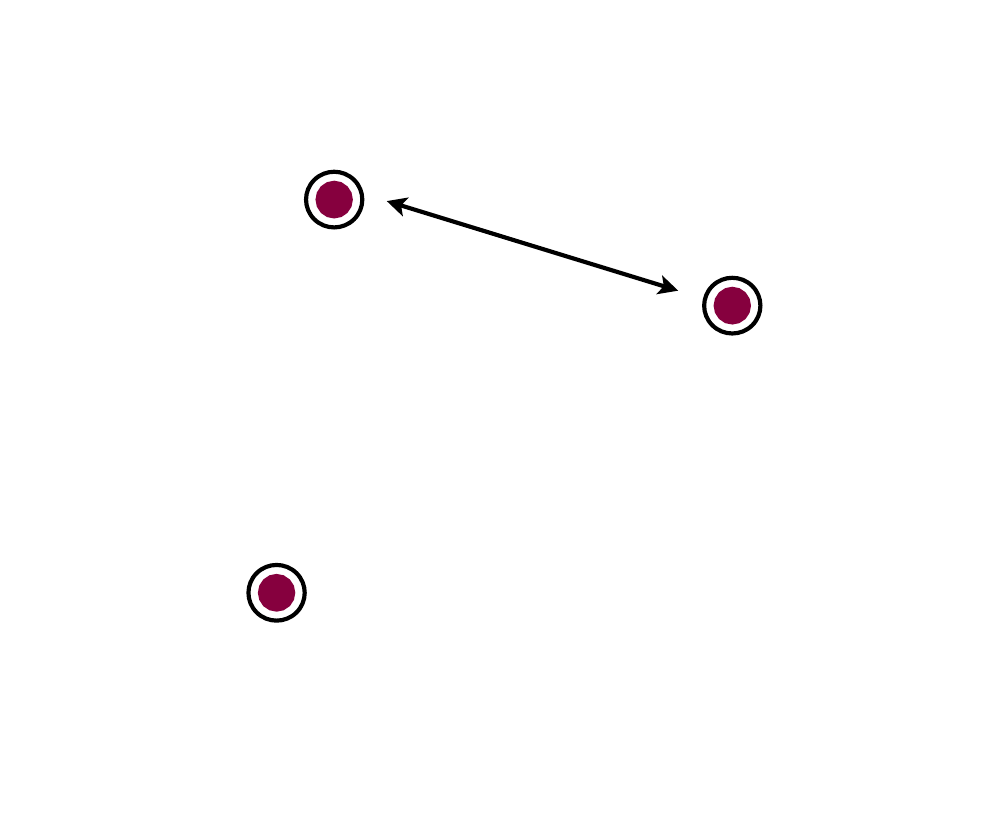}
        \caption{$\gA_{2,1}$}
    \end{subfigure}

    \caption{\textbf{Neighborhoods in Combinatorial Complexes.} Illustration of a combinatorial complex (top) and different neighborhood types, grouped by the target entity: nodes (second row), edges (third row), and faces (bottom row).}
    \label{app:fig_neighbourhoods}
\end{figure}

This section provides a detailed discussion of two key concepts in Topological Deep Learning: \emph{expanded neighborhoods} and \emph{the combinatorial explosion of message passing routes}. This discussion, while emphasizing the expressive power of combinatorial representations, highlights the challenges they pose when scaling to real-world problems.

\subsection{Neighborhood types in combinatorial complexes}
Combinatorial representations allow for a variety of neighborhood functions, depending on how cells of different ranks interact. For a 2-dimensional combinatorial complex (with $0$-, $1$-, and $2$-cells), \cref{app:fig_neighbourhoods} illustrates all possible neighborhood functions.
These relations can be grouped into two main groups:

\begin{itemize}
    \item \textbf{Incidence Relations:}
    \begin{itemize}
        \item (b) $\gI_{1\to0}$: edges $\rightarrow$ nodes
        \item (c) $\gI_{2\to0}$: faces $\rightarrow$ nodes
        \item (f) $\gI_{0\to1}$: nodes $\rightarrow$ edges
        \item (g) $\gI_{2\to1}$: faces $\rightarrow$ edges
        \item (j) $\gI_{0\to2}$: nodes $\rightarrow$ faces
        \item (k) $\gI_{1\to2}$: edges $\rightarrow$ faces
    \end{itemize}
    
    \item \textbf{Adjacency Relations:}
    \begin{itemize}
        \item (d) $\gA_{0,1}$: nodes $\rightarrow$ nodes (via edges)
        \item (e) $\gA_{0,2}$: nodes $\rightarrow$ nodes (via faces)
        \item (h) $\gA_{1,0}$: edges $\rightarrow$ edges (via nodes)
        \item (i) $\gA_{1,2}$: edges $\rightarrow$ edges (via faces)
        \item (l) $\gA_{2,0}$: faces $\rightarrow$ faces (via nodes)
        \item (m) $\gA_{2,1}$: faces $\rightarrow$ faces (via edges)
    \end{itemize}
\end{itemize}

\subsection{Counting neighborhoods}

The combinatorial complex depicted in \cref{app:fig_neighbourhoods} has rank $ R = \dim(\gT) = 2$, which results in \emph{12 distinct neighborhood functions}. These neighborhood functions can be divided into the two categories from above: (i) incidence relations and (ii) adjacency relations.

Under these assumptions, the number of incidence relations is $|I| = (R+1)R$, as each of the $R+1$ ranks can be connected to any of the remaining $R$ ranks. Similarly, each rank gives rise to $R$ possible adjacency relations (each defined through a different intermediary rank), also resulting in $|A| = (R+1)R$.
Combining these yields a total of
\[
|\gN_{\gT}| = 2(R+1)R
\]
distinct neighborhood functions. For the example in \cref{app:fig_neighbourhoods}, where $R = 2$, this formula recovers the 12 neighborhoods observed.

\subsection{Message passing routes}

The number of neighborhood functions increases quadratically with the maximum rank, which may initially seem manageable, especially since most practical datasets involve low-rank structures (e.g., $R=2$ or $R=3$). However, the true combinatorial complexity arises when considering how the results neighborhoods can be composed. To formalize this notion, the following definition is introduced.

\begin{definition}[Message passing route]
    A message passing route is an ordered set of neighborhood functions that a model uses to propagate information across a data structure.
\end{definition}

Information may flow \emph{intra-rank}, through adjacency relations, or \emph{inter-rank}, through incidence relations. For example, a network might use the message passing route $[\gI_{2 \to 1}, \gI_{1 \to 0}, \gI_{0 \to 2}, \gA_{0,1}]$ to transmit information from faces to edges, then from edges to nodes, from nodes back to faces, and finally between nodes connected by an edge.

Given a fixed combinatorial complex, the number of possible message passing routes is infinite, since any sequence can reuse the same neighborhood relations and grow to arbitrary length. In what follows, the analysis is restricted to a subset of message passing routes that are particularly relevant for the design of higher-order neural architectures.

\begin{definition}[Minimal message passing route]
    A minimal message passing route is a message passing route of minimal length that updates all ranks in a higher-order representation at least once. Furthermore, it ensures that, after a sufficient number of message passing steps, the representation of each rank depends on all other ranks.
\end{definition}

Minimal routes are of particular interest, as they guarantee full interdependence between all cells---a desirable property for expressive models. Moreover, minimal length implies simplicity in design, yielding the most compact routes that preserve this full dependency. For instance, the previously introduced route $[\gI_{2 \to 1}, \gI_{1 \to 0}, \gI_{0 \to 2}, \gA_{0,1}]$ can be transformed into a minimal message passing route by removing the adjacency $\gA_{0,1}$. In this reduced route, information from all cells becomes accessible to all others within two message passing steps. In fact, consider the case of $0$-cells: after the first step, they depend on $1$-cells; after the second step, since $1$-cells now incorporate information from $2$-cells, the $0$-cells become indirectly connected to them as well.

By definition, minimal message passing routes rely exclusively on incidence relations, as adjacencies do not propagate information inter-rank. Since each incidence targets a specific rank, and every rank must be updated once, minimal routes necessarily have length $R+1$.
To enumerate all possible minimal routes, consider the set of ranks $T = \{0, 1, \dots, R\}$, of size $R+1$. There are $(R+1)!$ distinct permutations of these ranks, each defining a unique order in which the ranks are updated. Given such a permutation $\{r_0, r_1, \dots, r_{R}\}$, a corresponding minimal message passing route can be constructed by selecting the sequence of incidence relations $\gI_{r_{j-1} \to r_j}$ for $j = 1, \dots, R$, and closing the cycle with $\gI_{r_{R} \to r_0}$. This construction yields exactly one valid minimal route per permutation, resulting in a total of $N = (R+1)!$ such routes.
In the case of graphs, where $R = 1$, this results in exactly two minimal message passing routes.

\subsection{Beyond minimal routes: including adjacencies}
Minimal message passing routes represent only a small subset of the possible routes that can be constructed using the full set of neighborhoods in a combinatorial complex. A natural extension involves augmenting these minimal routes with adjacency operations, a common practice in many TDL architectures. For simplicity, the analysis here is restricted to the case where adjacencies are applied strictly after all incidence relations.

As previously discussed, for a given rank $r$, there are $R$ possible adjacency relations of the form $\gA_{r,x}$. Since adjacencies operate intra-rank, they do not affect inter-rank dependencies and merely update the representation of the corresponding rank. As a consequence, the order in which these adjacencies are applied does not impact the outcome of the message passing route. For example, applying $\gA_{0,0}$ followed by $\gA_{1,1}$ produces the same result as applying them in reverse order.

Given that each of the $R+1$ ranks can be updated with any of their $R$ possible adjacencies, there are $(R+1)^{R}$ combinations of adjacency updates. When combined with the $(R+1)!$ minimal incidence-based routes, the total number of extended message passing routes becomes $\hat{N}=(R+1)! \cdot (R+1)^{R}$.

\section{Complexity Analysis}\label{app:complexity}
To analyze the time complexity (in terms of FLOPs) and the memory footprint of the \textsc{HOPSE} framework, we distinguish between two primary components: (i) the one-time preprocessing of topological descriptors, and (ii) the operations performed during each training iteration.

\paragraph{Preprocessing Overhead and Real-World Implementation.} 
The structural and positional encodings $g_k(\cdot)$ characterize the graph topology independently of the training loop. This allows the tensors $\mathbf{X}_{r, k}$ to be pre-computed once. The complexity of this step depends heavily on the chosen encoding:

\begin{itemize}
    \item \textbf{Exact Spectral Methods:} Encodings such as \texttt{LapPE} and exact \texttt{HKdiagSE} require eigendecomposition of the Laplacian matrix, which scales as $\mathcal{O}(V^3)$ for $V$ nodes. Similarly, exact \texttt{PPRFE} requires matrix inversion, also $\mathcal{O}(V^3)$.
    \item \textbf{Sparse Approximations:} In real-world large-scale applications, we utilize optimized sparse methods. The Chebyshev polynomial approximation for \texttt{HKFE} and the \textsc{APPNP} power-iteration for \texttt{PPRFE} both scale as $\mathcal{O}(K|E|)$, where $K$ is the number of iterations and $|E|$ is the number of edges.
\end{itemize}

By offloading these structure-dependent computations to a preprocessing phase, the training complexity becomes decoupled from the graph's edge density.


\paragraph{Train time computation.} The trainable components of the HOPSE framework are the functions $\varepsilon_r(\cdot)$, $f_{r,k}(\cdot)$, and $\Theta_r(\cdot)$, for all $r = 1, \dots, R$ and $k = 1, \dots, K$. For these functions, HOPSE uses MLPs with $L$ layers.
An MLP with $L$ layers applied to $N$ elements with input and output dimensions $D_{\text{in}}$ and $D_{\text{out}}$ has a computational complexity of $\mathcal{O}(N D_{\text{in}} D_{\text{out}})$. For simplicity, let us consider fixed input and output dimensions for each rank. 
The complexity of generating the initial features $\hat{\mathbf{Z}}_r$ in \cref{eq:initial_features} is
$\mathcal{O}(N_rD_r^2L)$,
where $N_r$ is the number of cells of rank $r$, and $D_r$ is the feature dimension.
Next, we consider the cost of computing the transformed encodings $\mathbf{\hat{X}}_{r, k}$. This cost is scaled by a factor $T$, representing the number of independent neighborhood components aggregated for a given rank. Specifically, if $\psi_k(\cdot)$ concatenates the results of different encodings across neighborhoods, $T$ corresponds to the number of neighborhoods considered ($|\mathcal{N}_C|$). If $\psi_k(\cdot)$ is instead a pooling operation (e.g., sum, mean, or max), $T$ reflects the number of distinct neighborhood types. Assuming each neighborhood utilizes $K$ encodings, the complexity for all transformed structural features is:
\[
\mathcal{O}(N_r D_r^2 T K L),
\]
assuming that each function $f_{r,k}$ preserves the feature dimensionality.
Assuming concatenation is used as the aggregation operation in \cref{eq:final_embedding}, and that the output features have dimension $D_r$, the complexity of computing the final embeddings $\mathbf{H}_r$ is:
\[
\mathcal{O}(N_r D_r^2 (TK+1) L).
\]
Combining all components, the total complexity for rank-$r$ cells is:
\begin{equation*}
    C_r = \mathcal{O}\left(N_r D_r^2 L + N_r D_r^2 TK L + N_r D_r^2 (TK+1) L\right)=\mathcal{O}(N_rD_r^2(TK+1)L).
\end{equation*}
The final complexity, assuming that all cells have the same input and hidden dimensions, is:
\[
C_{\text{HOPSE}} = \sum_{r=0}^R C_r = \sum_{r=0}^R \mathcal{O}(N_r D_r^2 (TK+1) L).
\]

Crucially, this complexity is \textbf{linear} with respect to the number of cells $N_r$ and independent of the number of edges $|E|$ or the specific connectivity of the complex.

\paragraph{Comparison to Message-Passing Neural Networks (MPNNs).} 
A standard MPNN layer has a complexity of approximately $\mathcal{O}(|E|D + |V|D^2)$. Summing over $L_{\text{GNN}}$ layers (we consider, for simplicity, a GNN with the same number of layers as the MLP), the cost is:
\begin{equation}
    C_{\text{MPNN}} = \mathcal{O}(L_{\text{GNN}}(|E|D + |V|D^2)).
\end{equation}
In contrast, \textsc{HOPSE} training complexity (working on a graph, we have $R=0$, and $T=1$ necessarily) is $\mathcal{O}(L |V| D^2 K)$. \textsc{HOPSE} becomes more efficient than an MPNN when the following condition is met:
\begin{equation}
    |E| + |V|D > |V|DK \cdot \frac{L}{L_{\text{GNN}}}.
\end{equation}
In practice, since $K$ is a small constant (e.g., $K=4$ in our \textsc{HOPSE-M-C} configuration) and $D$ is the feature dimension, \textsc{HOPSE} is significantly faster for dense graphs where $|E| \gg |V|$. Furthermore, \textsc{HOPSE} offers a distinct advantage in capturing long-range dependencies. While the depth of the \textsc{HOPSE} MLP is independent of the graph diameter, standard GNNs require the number of message-passing layers to grow linearly with the distance of the required interactions. In such scenarios, the iterative aggregation across many layers becomes a dominant computational bottleneck, whereas \textsc{HOPSE} maintains its efficiency by utilizing precomputed global or multi-scale descriptors within a shallow, task-independent MLP architecture.

\paragraph{Theoretical Activation Memory Advantage During Training.}
Standard Higher-Order Message Passing (HOMP) models maintain explicit computation graphs across cell hierarchies. To support backpropagation, intermediate message tensors must be saved in GPU memory. This causes the peak activation VRAM to scale with the number of relations in the Hasse neighborhood graph, bounded by:
\[
\mathcal{O}\left(\sum_{r=0}^R | E_{\mathcal{G}_{\mathcal{N}_r}} | \cdot D_r \right)
\]
Here, $R$ is the maximum cell rank, $| E_{\mathcal{G}_{\mathcal{N}_r}} |$ is the number of active structural relations (edges in the Hasse graph), and $D_r$ is the hidden feature dimension for rank $r$.

In contrast, \textsc{HOPSE} decouples topological message passing by applying independent MLPs over precomputed cell encodings. As a result, the autograd engine only caches activation tensors per cell. This yields an activation memory footprint of:
\[
\mathcal{O}\left(\sum_{r=0}^R N_r \cdot D_r \right)
\]
where $N_r$ is the total count of $r$-dimensional cells in the complex. This footprint scales strictly linearly with the cell count $N_r$, completely independent of the higher-order structural density $| E_{\mathcal{G}_{\mathcal{N}_r}} |$.

\paragraph{Practical VRAM Consumption and Capacity Overhead.}
While \textsc{HOPSE} offers an asymptotic activation advantage, practical experiments show that it can exhibit higher peak VRAM utilization in certain configurations (see \cref{app:extended_results}).

The empirical behavior of higher peak VRAM in specific \textsc{HOPSE} configurations occurs because the top-performing hyperparameter configurations select multi-scale encodings across $K$ distinct neighborhood operators (e.g., lower boundary, upper coboundary, Laplacians) over $T$ scale steps. This introduces an encoding multiplicity factor of $TK + 1$ (accounting for $T$ diffusion scales across $K$ operators plus 1 base attribute layer), leading to a parameter footprint of:
\[
\mathcal{O}\left(\sum_{r=0}^R D_r^2 (TK+1) L \right)
\]
where $L$ is the total number of network layers. Furthermore, precomputed encoding matrices requiring $\mathcal{O}(\sum_{r=0}^R N_r \cdot D_{k'})$ space, where $D_{k'}$ is the feature dimension per encoding channel, are typically cached directly in GPU VRAM during training to avoid CPU-to-GPU data transfer bottlenecks.

Because baseline HOMP models are prone to backpropagation Out-Of-Memory (OOM) errors on dense graph connections, their hyperparameter tuning is forced into narrower feature widths $D_r$ and fewer neighborhood relations. In contrast, \textsc{HOPSE} can safely allocate higher static feature memory without risking structural OOM crashes as complex density grows.

\paragraph{Summary.} 
This analysis demonstrates that \textsc{HOPSE} achieves high expressive power while maintaining computational and activation memory complexities that scale linearly with the number of cells. The experimental results in \cref{tbl:best_rerun_runtime} and \cref{tbl:best_rerun_epoch_time} confirm this efficiency, particularly in datasets where the Hasse graph decomposition allows for rapid parallel MLP evaluations across cell sets.
\section{Proofs}\label{app:proofs}

In this section, we aim to prove \Cref{thm:theorem_ccwl,thm:permutation_eq}. We begin by formalizing the notion of rooted refinement trees, which are the fundamental structure in a graph that WL-type algorithms operate on.

\begin{definition}[Graph Rooted Refinement Tree]\label{def:rooted-refinement-tree} Let $\mathcal{G} = (\cV,\mathcal{E}, F_\cV, F_\mathcal{E})$ be a featured graph, and let $\mathcal{N}: \cV \to \mathcal{P}(\cV)$ be a neighborhood function. For each node $v \in \cV$ and depth $K \in \mathbb{N}$, the rooted refinement tree at depth $K$ is defined recursively:
\begin{align}
T_v^{(0)} &:= F_\cV(v), \\
T_v^{(K)} &:= \left( T_v^{(K-1)}, \left\{\!\!\left\{ T_u^{(K-1)} \mid u \in \mathcal{N}(v) \right\}\!\!\right\} \right) \;\;\text{for} \;K\geq1,
\end{align}
where $\{\!\!\{\cdot\}\!\!\}$ denotes a multiset.

The tree $T_{v}^{(K)}$ is a nested structure encoding the $K$-hop neighborhood of $v$: at each level $\ell \leq K$, it contains the subtree $T_{v}^{(\ell)}$ and the multiset of subtrees from the direct neighbors at that level.

Two rooted trees $T_{x}^{(K)}$ and $T_{y}^{(K)}$ are isomorphic, denoted $T_{x}^{(K)} \cong T_{y}^{(K)}$, if there exists a bijection between their nodes preserving the recursive structure, i.e., root features and multisets of children.
\end{definition}
Similarly, rooted refinement trees can be defined for combinatorial complexes as follows:
\begin{definition}[Combinatorial Complex Rooted Refinement Tree]
    Let $(\cV, \cC, rk)$ be a combinatorial complex, with the neighborhood function $\mathcal{N}: \cC \to \mathcal{P}(\cC)$, and let $\mathcal{F} = \{F_r\}_{r \geq 0}$ be a signal over this combinatorial complex. For each cell $\sigma \in \mathcal{C}$ and depth $K \in \mathbb{N}$, the rooted refinement tree $T_{\sigma}^{(K)}$ is defined recursively as follows:
    \begin{align}
        T_{\sigma}^{(0)} &:=  F_{\mathrm{rk}(\sigma)}(\sigma),\\
        T_{\sigma}^{(K)} &:= \left( T_\sigma^{(K-1)}, \left\{\!\!\left\{ T_\delta^{(K-1)} \mid \delta \in \mathcal{N}(\sigma) \right\}\!\!\right\} \right)
    \end{align}
    Moreover, we denote $\mathcal{T}_K$ as the set of all possible rooted refinement trees at depth $K$ up to isomorphism. 
\end{definition}
\begin{lemma}[Rooted Tree Characterization of Weisfeiler-Leman]\label{lemma:wl-rooted-refinement-tree}
    Let $\mathcal{G} = (\cV,\mathcal{E},F_\cV,F_\mathcal{E})$ be a featured graph. For any node $v \in \cV$ and depth $K \geq 0$, there exists an injective function $\Phi_K: \mathcal{T}_K \to \mathbb{N}$ such that 
    \begin{align}
        c^{(K)}(v) = \Phi_K(T_v^{(K)}),
    \end{align}
    where $c^{(K)}(v)$ denotes the $K$-step Weisfeiler-Leman color of node $v$.
\end{lemma}
\begin{proof}
We proceed by induction on $K$.

\paragraph{Base case ($K=0$).} By WL initialization, we know $c^{(0)}(v) = F_\cV(v)$. Thus, $\Phi_0$ can be the trivially injective identity function. Then, $c^{(0)}(v) = \Phi_0(T_v^{(0)})$.

\paragraph{Induction step.} Assume the claim holds for all depths $\ell < K$. Formally, for each $\ell < K$, there exists an injective function $\Phi_\ell: \mathcal{T}_\ell \to \mathbb{N}$ such that $c^{(\ell)}(v) = \Phi_\ell(T_{v}^{(\ell)})$ for all $v \in \cV$. By the WL update rule at step $K$, the color of node $v$ is updated as
\begin{align}
    \label{eq:wl-update}
c^{(K)}(v) = \HASH\left( c^{(K-1)}(v), \left\{\!\!\left\{ c^{(K-1)}(u) \mid u \in \mathcal{N}(v) \right\}\!\!\right\} \right),
\end{align}
where $\HASH$ is a perfect hash function. By the inductive hypothesis, we know
\begin{align}
    c^{(K-1)}(v) &= \Phi_{K-1}(T_v^{(K-1)}), \\
c^{(K-1)}(u) &= \Phi_{K-1}(T_u^{(K-1)}) \quad \text{for all } u \in \mathcal{N}(v).
\end{align}
Therefore, by substituting into \eqref{eq:wl-update}: 
\begin{align}
    \label{eq:wl-subst}
c^{(K)}(v) = \HASH\left( \Phi_{K-1}(T_v^{(K-1)}), \left\{\!\!\left\{ \Phi_{K-1}(T_u^{(K-1)}) \mid u \in \mathcal{N}(v) \right\}\!\!\right\} \right).
\end{align}
Then, we can define $\Phi_K: \mathcal{T}_K \to \mathbb{N}$ as follows. For any tree $\mathcal{T} = (T, \mathcal{M}) \in \mathcal{T}_K$, where $T\in \mathcal{T}_{K-1}$ and $\mathcal{M}$ is a multiset of trees in $\mathcal{T}_{K-1}$, we set:
\begin{align}
    \label{eq:phi-k-def}
\Phi_K\left( (T, \mathcal{M}) \right) := \HASH\left( \Phi_{K-1}(T), \left\{\!\!\left\{ \Phi_{K-1}(S) \mid S \in \mathcal{M} \right\}\!\!\right\} \right).
\end{align}
Note that in \eqref{eq:phi-k-def}, $\mathcal{M}$ is a multiset of \emph{trees} in $\mathcal{T}_{K-1}$ and when we write $\{\!\!\{\Phi_{K-1}(S) \mid S \in \mathcal{M}\}\!\!\}$, each $S$ is already a rooted tree at depth $K-1$. Therefore, we have $\mathcal{M} = \{\!\!\{ T_u^{(K-1)} \mid u \in \mathcal{N}(v) \}\!\!\}$, so each $S \in \mathcal{M}$ is one of the trees $T_u^{(K-1)}$.

To prove that $\Phi_K$ is injective, suppose $\Phi_K\left((T_1, \mathcal{M}_1)\right) = \Phi_K\left((T_2, \mathcal{M}_2)\right)$ for $(T_1, \mathcal{M}_1), (T_2, \mathcal{M}_2) \in \mathcal{T}_K$. Then by definition \ref{eq:phi-k-def}:
\begin{align}
    \HASH\left( \Phi_{K{-}1}(T_1), \{\!\!\{ \Phi_{K{-}1}(S) \!\!\mid\!\! S \in \mathcal{M}_1 \}\!\!\} \right) = \HASH\left( \Phi_{K{-}1}(T_2), \{\!\!\{ \Phi_{K{-}1}(S) \!\!\mid\!\! S \in \mathcal{M}_2 \}\!\!\} \right).
\end{align}
Since $\HASH$ is injective, we have
\begin{align}
    \Phi_{K-1}(T_1) &= \Phi_{K-1}(T_2), \\
\{\!\!\{ \Phi_{K-1}(S) \mid S \in \mathcal{M}_1 \}\!\!\} &= \{\!\!\{ \Phi_{K-1}(S) \mid S \in \mathcal{M}_2 \}\!\!\}
\end{align}
Moreover, by induction hypothesis, we know $\Phi_{K-1}$ is injective, then:
\begin{align}
T_1 &= T_2, \\
\mathcal{M}_1 &= \mathcal{M}_2 
\end{align}
Therefore $(T_1, \mathcal{M}_1) = (T_2, \mathcal{M}_2)$, which proves that $\Phi_K$ is injective as well.

Thus, we have:
\begin{equation}
\Phi_K(T_\sigma^{(K)}) = \HASH\left( \Phi_{K-1}(T_v^{(K-1)}), \left\{\!\!\left\{ \Phi_{K-1}(T_u^{(K-1)}) \mid u \in \mathcal{N}(v) \right\}\!\!\right\} \right).
\end{equation}
which results in
\begin{equation}
c^{(K)}(\sigma) = \Phi_K(T_\sigma^{(K)}),
\end{equation}
and shows that the induction step also holds.
\end{proof}
By Lemma~\ref{lemma:wl-rooted-refinement-tree}, WL colors are injective functions of rooted trees. Consequently, any function with WL-equivalent expressivity must also preserve rooted refinement trees via some injective function. Moreover, we can express the similar result for CCWL as well:
\begin{lemma}\label{lemma:ccwl-rooted-refinement-tree}
    Let $(\mathcal{V}, \mathcal{C}, rk)$ be a combinatorial complex with neighborhood function $\mathcal{N}: \mathcal{C} \to \mathcal{P}(\mathcal{C})$, and let $\mathcal{F} = \{F_r\}_{r\geq0}$ be a signal over this combinatorial complex. For any cell $\sigma \in \mathcal{C}$ and depth $K \geq 0$, there exists an injective function $\Phi_K^\mathcal{N}: \mathcal{T}_K \to \mathbb{N}$ such that
    \begin{align}
        c^{(K)}_{\mathcal{N}}(\sigma) = \Phi_K^{\mathcal{N}}\left( T_\sigma^{(K)} \right),
    \end{align}
    where $c^{(K)}_{\mathcal{N}}(\sigma)$ denotes the $K$-step CCWL color of cell $\sigma$ under neighborhood $\mathcal{N}$.
\end{lemma}
\begin{proof}
    The CCWL update rule is: 
    \begin{align}
        c^{(K+1)}_{\mathcal{N}}(\sigma) = \HASH\left( c^{(K)}_{\mathcal{N}}(\sigma), \left\{\!\!\left\{ c^{(K)}_{\mathcal{N}}(\tau) \mid \tau \in \mathcal{N}(\sigma) \right\}\!\!\right\} \right),
    \end{align}
    with initialization $c^{(0)}_{\mathcal{N}}(\sigma) = F_{rk(\sigma)}(\sigma)$.

    The update rule has identical structure to the WL update on graphs, with the graph neighborhood replaced by the complex neighborhood $\mathcal{N}$, and the CC rooted refinement tree is defined similarly to the graph rooted refinement tree. Therefore, Lemma~\ref{lemma:ccwl-rooted-refinement-tree}, can be proved similarly to Lemma~\ref{lemma:wl-rooted-refinement-tree}, by replacing featured graph with signal-augmented CC, graph neighborhood with CC neighborhood, and WL colors with CCWL colors.

    Thus, the induction can be done similarly and it yields an injective function $\Phi_K^{\mathcal{N}}: \mathcal{T}_K\to\mathbb{N}$ such that $c^{(K)}_{\mathcal{N}}(\sigma) = \Phi_K^{\mathcal{N}}(T_\sigma^{(K)})$ for any $\sigma \in \mathcal{C}$.
\end{proof}
We can now connect CCWL equivalence with preserving rooted refinement trees.
\begin{lemma}\label{lemma:ccwl-equivalence}
    Let $g:(\cC, \cJ_\cN) \to \mathbb{R}^d$ be a function equally expressive as the WL test on $\cJ_\cN$. Then there exists an injective function $\Phi: \cT_K \to \R^d$ such that, for all cells $\sigma \in \mathcal{C}$: 
    \begin{equation}
        g(\sigma, \cJ_\cN)=\Phi(T_\sigma^{(K)}).
    \end{equation}
\end{lemma}
\begin{proof}
    Since $g$ is equally expressive as the WL test on $\cJ_\cN$, we know that $g$ is equal to the WL test up to an injective function. Moreover, according to Proposition B.11 in \cite{papillon2024topotune}, the CCWL test on two CCs is equivalent to the WL test on their associated strictly augmented Hasse graph. Hence, there exists a map $\Phi_1$
    \begin{equation*}
        g(\sigma,\cJ_\cN)=\Phi_1\left(c^{(K)}_{\mathcal{N}}(\sigma)\right),
    \end{equation*}
    with $\Phi_1$ injective. We also know, from Lemma \ref{lemma:ccwl-rooted-refinement-tree} that there exists $\Phi_K^{\cN}$ injective such that
    \begin{equation*}
        c^{(K)}_{\mathcal{N}}(\sigma) = \Phi_K^{\cN}\left( T_\sigma^{(K)} \right).
    \end{equation*}
    By combining the two equations, we obtain
    \begin{equation*}
        g(\sigma,\cJ_\cN)=\Phi_1\left(\Phi_K^{\mathcal{N}}(T_\sigma^{(K)})\right)= \Phi(T_\sigma^{(K)}),
    \end{equation*}
    where $\Phi$ is the combination of two injective functions and thus injective.
\end{proof}

Now the two main theorems of the paper can be proven. 
First, we prove that HOPSE is \textit{as powerful} as the CCWL test, and then we will prove with an example that HOPSE is strictly \textit{more} powerful. 

\begin{theorem}[HOPSE $\geq$ CCWL]
\label{thm:ccwl_as_powerful}
Let HOPSE be configured with a set of neighborhoods $\cN_\cC = \{\cN_1, \ldots, \cN_T\}$ and encodings $\{g_k\}_{k=1}^K$. HOPSE is at least as powerful as the $K$-step CCWL test in distinguishing non-isomorphic combinatorial complexes if the following assumptions hold:

\begin{description}
\item[(A1)] There exists at least one $k^* \in \{1, \ldots, K\}$ and a neighborhood $\cN \in \cN_\cC$ such that the encoding $g_{k^*}(\cdot, \cJ_\cN)$ is equally expressive as the CCWL test on $\cJ_\cN$.

\item[(A2)] The functions $\varepsilon_r(\cdot)$, $\psi_k(\cdot)$, $f_{r,k}(\cdot)$, and $\Theta_r(\cdot)$ are injective for all $k \in \{1, \ldots, K\}$ and $r \in \{0, \ldots, R\}$.
\end{description}

Then HOPSE is at least as powerful as CCWL.
\end{theorem}

\begin{proof}
Let $(\mathcal{V}_1, \mathcal{C}_1, rk_1, F_1)$ and $(\mathcal{V}_2, \mathcal{C}_2,rk_2,F_2)$ be two combinatorial complexes, and let $\sigma_1 \in \cC_1$ and $\sigma_2 \in \cC_2$ be two cells. Suppose HOPSE assigns the same final representation to both cells:
\begin{equation}
\label{eq:hopse-equal}
h_{\sigma_1} = h_{\sigma_2}.
\end{equation}

We will show that this implies $c^{(K)}_{\cN}(\sigma_1) = c^{(K)}_{\cN}(\sigma_2)$ for the CCWL coloring under neighborhood $\cN$.

\paragraph{Step 1:}

By the HOPSE architecture (Equations 4-5 from the main paper), the final representation $h_\sigma$ for rank-$r$ cells is computed as:
\begin{equation}
    \begin{split}
    h_\sigma = \Theta_r\Big( \varepsilon_r(Z_\sigma), &f_{r,1}(\psi_1(\{g_1(\sigma, \cJ_{\cN_i}) \mid \cN_i \in \cN_\cC^{\to r}\})), \ldots, \\ 
    &f_{r,K}(\psi_K(\{g_K(\sigma, \cJ_{\cN_i}) \mid \cN_i \in \cN_\cC^{\to r}\})) \Big),
    \end{split}
\end{equation}
where $Z_\sigma$ are the initial features, $\cN_\cC^{\to r}$ are the rank-$r$-targeted neighborhoods in $\cN_\cC$, and $\psi_k$, $f_{r,k}$, $\varepsilon_r$, $\Theta_r$ are the aggregation and transformation functions.

By assumption (A2), each of these functions is injective. The composition of injective functions is injective. Therefore, $h_\sigma$ is an injective function of the tuple:
\begin{equation}
\label{eq:tuple}
\left( Z_\sigma, g_1(\sigma, \cJ_{\cN_1}), \ldots, g_1(\sigma, \cJ_{\cN_M}), g_2(\sigma, \cJ_{\cN_1}), \ldots, g_K(\sigma, \cJ_{\cN_M}) \right),
\end{equation}
with $M=|\cN_\cC^{\to r}|$. 

From equation~\ref{eq:hopse-equal}, we conclude that the tuples for $\sigma_1$ and $\sigma_2$ are equal:
\begin{equation}
\label{eq:tuples-equal}
g_k(\sigma_1, \cJ_{\cN_i}) = g_k(\sigma_2, \cJ_{\cN_i}) \quad \text{for all } k \in \{1, \ldots, K\}, \; \cN_i \in \cN_\cC^{\to r}.
\end{equation}

(We also have $Z_{\sigma_1} = Z_{\sigma_2}$, i.e., identical initial features, but this is implicit in what follows.)

\paragraph{Step 2:}

By assumption (A1) and Lemma \ref{lemma:ccwl-equivalence}, there exists $k^* \in \{1, \ldots, K\}$ and $\cN \in \cN_\cC$ such that:
\begin{equation}
g_{k^*}(\sigma, \cJ_\cN) = \Phi(T_\sigma^{(K)})
\end{equation}
for an injective function $\Phi: \cT_K \to \R^d$.

From ~\eqref{eq:tuples-equal} applied to $k = k^*$ and neighborhood $\cN$:
\begin{equation}
g_{k^*}(\sigma_1, \cJ_\cN) = g_{k^*}(\sigma_2, \cJ_\cN).
\end{equation}

Therefore:
\begin{equation}
\Phi(T_{\sigma_1}^{(K)}) = \Phi(T_{\sigma_2}^{(K)}).
\end{equation}

Since $\Phi$ is injective, this implies:
\begin{equation}
\label{eq:trees-iso}
T_{\sigma_1}^{(K)} \cong T_{\sigma_2}^{(K)}.
\end{equation}

That is, the rooted refinement trees of $\sigma_1$ and $\sigma_2$ at depth $K$ are isomorphic.

\paragraph{Step 3:}

By Lemma~\ref{lemma:ccwl-rooted-refinement-tree}, there exists an injective function $\Phi_K^\cN: \cT_K \to \mathbb{N}$ such that:
\begin{align}
c^{(K)}_{\cN}(\sigma_1) &= \Phi_K^\cN(T_{\sigma_1}^{(K)}), \\
c^{(K)}_{\cN}(\sigma_2) &= \Phi_K^\cN(T_{\sigma_2}^{(K)}).
\end{align}

From equation~\ref{eq:trees-iso} and the fact that $\Phi_K^\cN$ is well-defined on isomorphism classes of trees:
\begin{equation}
c^{(K)}_{\cN}(\sigma_1) = \Phi_K^\cN(T_{\sigma_1}^{(K)}) = \Phi_K^\cN(T_{\sigma_2}^{(K)}) = c^{(K)}_{\cN}(\sigma_2).
\end{equation}

Thus, whenever HOPSE fails to distinguish two cells (i.e., $h_{\sigma_1} = h_{\sigma_2}$), the CCWL test also fails to distinguish them (i.e., $c^{(K)}_{\cN}(\sigma_1) = c^{(K)}_{\cN}(\sigma_2)$). This establishes that HOPSE is at least as powerful as CCWL.
\end{proof}

Finally, we prove the main results.
\begin{reptheorem}[\cref{thm:theorem_ccwl}]
HOPSE is more powerful than the CCWL test in distinguishing non-isomorphic combinatorial complexes if the functions $\varepsilon_r(\cdot)$, $\psi_k (\cdot)$, $f_{r,k}(\cdot)$, and $\Theta_{r}(\cdot)$ are injective for all $k \in \{1,\dots, K\}$ and for all $r \in \{0,\dots, R\}$ and $g_k(\cdot)$ is equally expressive as the WL test on the corresponding Hasse graph for at least one $k \in \{1,\dots, K\}$.
\end{reptheorem}

\begin{figure}[t]
    \vspace{-20pt}
    \centering
    \includegraphics[width=\textwidth]{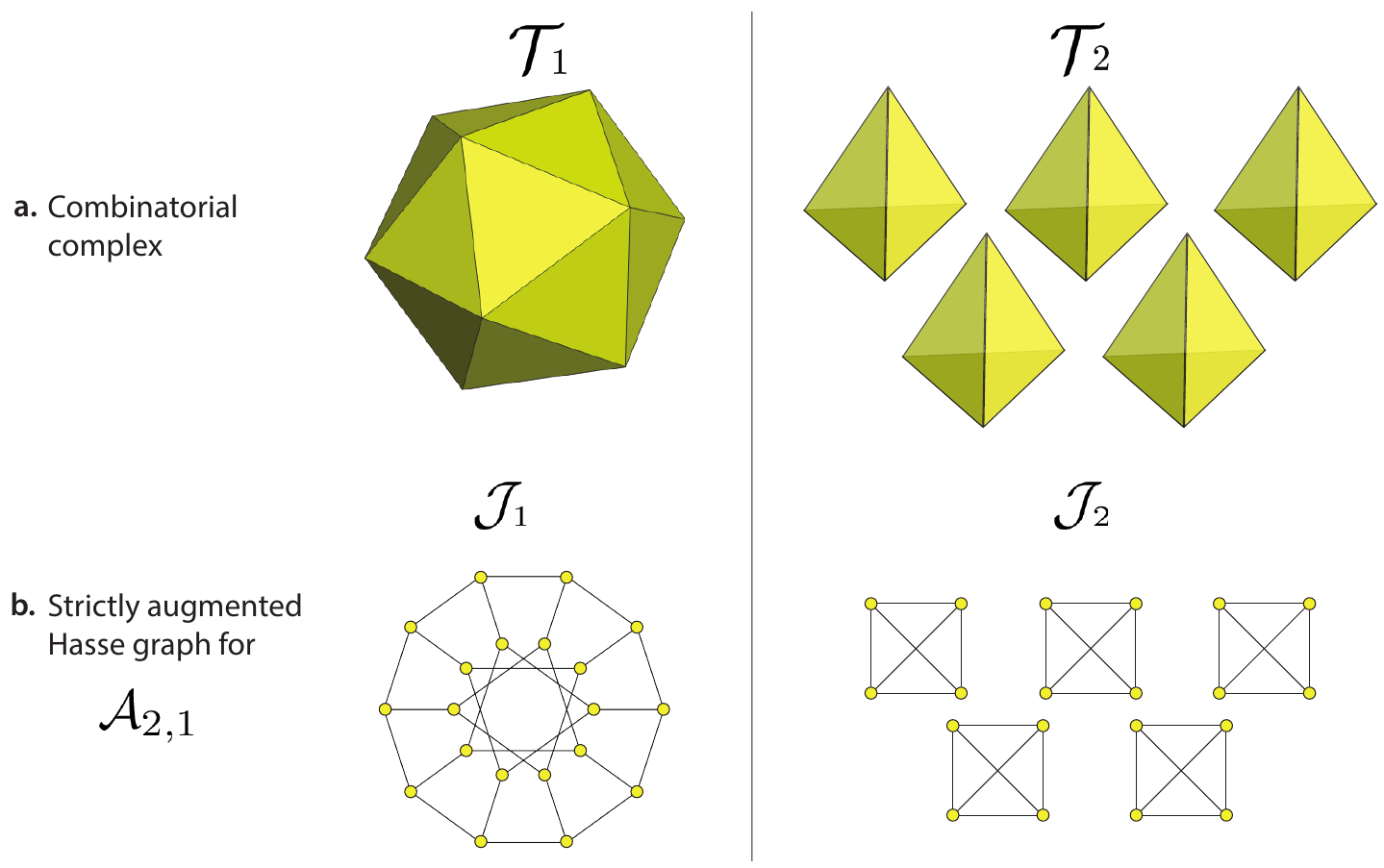}
\caption{\textbf{Combinatorial Complexes example.} The two CCs are not distinguishable by CCWL on $\mathcal{A}_{2,1}$ since both generate Hasse graphs that are 3-regular.  Figure adapted from \citet{papillon2024topotune}.}
\label{fig:ccwl_as_powerful}
\end{figure}

\begin{proof}
Since HOPSE matches the expressive power of CCWL, as established in \cref{thm:ccwl_as_powerful}, to prove this theorem it is sufficient to present a case in which two combinatorial complexes are indistinguishable by the CCWL test but are successfully distinguished by the HOPSE algorithm. The example in \cref{fig:ccwl_as_powerful} illustrates two such CCs. Using the neighborhood $\gA_{2,1}$ (the faces are adjacent if they share an edge), the corresponding Hasse graphs $\cJ_1$ and $\cJ_2$ are constructed, both of which are 3-regular. It is known that the WL test cannot distinguish between regular graphs of the same degree \cite{kiefer2020power}. According to Proposition B.11 in \cite{papillon2024topotune}, the CCWL test on these CCs is equivalent to the WL test on $\cJ_1$ and $\cJ_2$, and thus also fails to distinguish them.
However, the HOPSE algorithm can separate the two cases. In fact, by choosing the function $g_1(\cdot)$ to return the size of the connected component to which a node belongs, the difference between the two HOPSE encodings is apparent.
\end{proof}

\begin{reptheorem}[\cref{thm:permutation_eq}]
    HOPSE is cell permutation equivariant if the functions $\varepsilon_r(\cdot)$, $g_k(\cdot)$, $\psi_k(\cdot)$, $f_{r,k}(\cdot)$, and $\Theta_{r}(\cdot)$ are cell permutation equivariant for all $k \in \{1, \dots, K\}$ and all $r \in \{0, \dots, R\}$.
\end{reptheorem}
\begin{proof}
To demonstrate that HOPSE is cell permutation equivariant, it suffices to show that the composition of permutation equivariant functions remains permutation equivariant. Since, by assumption, all functions that compose HOPSE are permutation equivariant, the result follows directly. Let $\pi(\cdot)$ be the permutation operator. Then $p(\cdot)$ is permutation equivariant if $\pi(p(\cdot))=p(\pi(\cdot))$. Given two permutation equivariant functions $p(\cdot)$ and $q(\cdot)$, their composition $r(\cdot) = (p \circ q) (\cdot)$ is also permutation equivariant:
\[
r(\pi(\cdot)) = (p \circ q) (\pi(\cdot)) = p(q(\pi(\cdot))) = p(\pi(q(\cdot))) = \pi(p(q(\cdot))) = \pi(r(\cdot)).
\]

\end{proof}

\paragraph{Remarks on Expressivity and Triangulated Tori.}
To further clarify the expressivity of \textsc{HOPSE} beyond simple component counting, we introduce an additional family of non-isomorphic combinatorial complexes: connected triangulated tori. We denote a triangulated torus of dimensions $N$ and $M$ as $T(N,M)$, which is constructed by taking a regular $N \times M$ grid embedded on a torus and triangulating each rectangular face to form a 2-dimensional simplicial complex. 

Consider pairs of equal area such as $T(4,9)$ versus $T(6,6)$ and $T(4,12)$ versus $T(6,8)$. Both members of these pairs are connected, meaning their component structure is identical and component counting carries zero distinguishing signal. Furthermore, their underlying 1-skeletons (graphs) are 1-WL-equivalent. Despite this, structural encodings such as \texttt{LapPE} and \texttt{ElectrostaticPE} computed over the higher-order Hasse graphs easily separate these complexes across all eight standard neighborhoods. This separation is driven by the differences in their nonzero Laplacian spectra, which reflect the distinct lengths of the wrap-around cycles inherent to their specific toroidal topologies.

This example highlights that the distinguishing information leveraged by \textsc{HOPSE} is not natively available to a standard Message-Passing Neural Network (MPNN). \textsc{HOPSE} computes encodings on the strictly augmented Hasse graphs of higher-order neighborhoods rather than the underlying graph backbone. For instance, in the example of Figure 4 (the dodecahedral graph versus five disjoint $K_4$ complexes), constructing the face-adjacency structure requires precisely the Hasse decomposition that constitutes the preprocessing step of \textsc{HOPSE}. Providing this explicitly decomposed feature to an MPNN would essentially reduce it to \textsc{HOPSE} paired with a redundant message-passing backbone. 

\section{Framework realization}\label{app:framework_realization}
This section provides an in-depth description of the architectural choices and characterization functions for the Hasse graphs resulting from the decomposition of a combinatorial complex. While the theoretical results assume specific properties for the functions $g_k(\cdot)$ and $\psi_k(\cdot)$, we emphasize that functions which do not strictly fulfill all theoretical guarantees remain empirically relevant for studying representational trade-offs.

\subsection{Precomputation of HOPSE-M (Manual Encodings)}

In the \textbf{HOPSE-M-C} and \textbf{HOPSE-M-F} variants, we utilize a library of manually selected encoding functions. These are partitioned into two distinct categories: those characterizing the \emph{structure} alone, and those also utilizing the \emph{node features}.

Let $\gG=(V,E)$ be an undirected graph with $|V|=n$ nodes and feature matrix $\mathbf{X} \in \R^{n \times d}$. Each characterization function $g_k$ is defined as follows:

\paragraph{Connectivity-only Encodings}
These functions capture topological properties independently of node attributes $\mathbf{X}$:
\begin{enumerate}
    \item $g_1(\cdot)=\mathrm{LapPE}_{i}=[\phi_1,\dots,\phi_i]\in\R^{n\times i}$: The first $i$ nontrivial eigenvectors of the graph Laplacian, providing spectral positional information \citep{dwivedi2021generalization}.
    \item $g_2(\cdot)=\mathrm{RWSE}_{K}\in\R^{n\times K}$: Random Walk Structural Encodings, defined by the diagonal entries of the $t$-step transition matrices $\mathbf{P}^t$ for $t=1,\dots,K$, capturing local cluster structures \citep{dwivedi2022graph}.
    \item $g_3(\cdot)= \mathrm{HKdiagSE}_{K}\in\R^{n\times K}$: Heat Kernel Diagonal Structural Encodings, capturing multiscale diffusion properties via the Laplacian eigenvalues $\lambda_j$ as $[\sum_{j=1}^n \exp(-\lambda_j\,t) \phi_j^2]_{t=1,\dots,K}$ \citep{sun2009concise}.
    \item $g_4(\cdot)=\mathrm{ElectrostaticPE}\in\R^{n\times d_4}$: Electrostatic Potential Positional Encodings, computed by solving the discrete Poisson equation on $\gG$, capturing global relative node positioning \citep{kreuzer2021rethinking}.
\end{enumerate}

\paragraph{Feature-based Encodings}
These functions condition on node features to assess the benefits of coupling topology with node semantics:
\begin{enumerate}
    \setcounter{enumi}{4}
    \item $g_5(\cdot)=\mathrm{HKFE} \in \R^{n\times d_5}$: \textbf{Heat Kernel Feature Encodings}. This function applies a heat kernel diffusion operator $e^{-t\mathbf{L}}$ to the node features $\mathbf{X}$ \citep{xu2019graphheat}. It computes $\mathbf{X}_t = e^{-t\mathbf{L}}\mathbf{X}$ across a range of diffusion scales $t \in [t_{\text{start}}, t_{\text{end}}]$, effectively smoothing features over spectral neighborhoods.
    \item $g_6(\cdot)=\mathrm{KHopFE} \in \R^{n\times d_6}$: \textbf{\textit{K}-hop Feature Encodings}. This encoder propagates node features through $k$ steps of the symmetric normalized adjacency matrix $\mathbf{\hat{A}} = \mathbf{D}^{-1/2}\mathbf{A}\mathbf{D}^{-1/2}$ \citep{feng2022khop}. Following the original implementation, we optimize $K \in \{1, 2, 3\}$, capturing local feature correlations within defined geodesic distances.
    \item $g_7(\cdot)=\mathrm{PPRFE} \in \R^{n\times d_7}$: \textbf{Personalized PageRank Feature Encodings}. This function computes a multiscale version of the Approximate Personalized PageRank \citep{Gasteiger2019appnp}. It balances local feature preservation and global topology by solving $(\mathbf{I} - (1-\alpha)\mathbf{\hat{A}})\mathbf{Z} = \alpha \mathbf{X}$ for various teleport probabilities $\alpha$.
\end{enumerate}

\paragraph{Remark: Directionality of Hasse Graphs.} Some PSEs (e.g., LapPE) require a symmetric Laplacian to guarantee real eigenvalues and orthonormal eigenvectors, so Hasse graphs are made undirected before encoding. This mostly affects incidences, since adjacencies are already symmetric in undirected complexes: e.g., $\mathcal{I}_{0\to1}$ and $\mathcal{I}_{1\to0}$ collapse to the same undirected graph. Although some information is lost, the resulting connectivity still relates cells of the two ranks involved, and HOPSE can readily integrate PSEs that exploit the directionality of Hasse graph expansions.


\subsection{Precomputation of HOPSE-GPSE (General Encodings)}

In contrast to manual selection, \textbf{HOPSE-GPSE} relies on \textbf{GPSE} \citep{canturk2023graph}, a GNN model pre-trained to learn latent representations of positional and structural properties. We evaluate variants pre-trained on \texttt{ZINC} and \texttt{MolPCBA}. 

To maintain consistency with the GPSE pre-training protocol, which utilizes random noise initialization to learn challenging invariants, we modify the Hasse graph initialization for this variant. For all neighborhoods $\gN \in \gN_{C}$, the cell features $F_{r,k} \in \gF_\gN$ of the induced Hasse graph $\gJ_{\gN}{\bigr|}_{k}$ are initialized as random noise: $F_{r, k} \sim N(0 , \mathbf{I}) \in \R^{20}$. 

\subsection{Learnable Embedding Functions}

The mappings $f_{r,k}$, $\varepsilon_r$, and $\Theta_r$ are implemented as $L$-layer multilayer perceptrons (MLPs) with independent parameters. Each layer $ \ell=1,\dots,L$ follows a ResNet-style architecture:

\begin{equation}  
\mathbf{h}_{r,k}^{(\ell)} 
= \sigma\!\Biggl(\mathrm{LayerNorm}\Bigl((\mathbf{h}_{r,k}^{(\ell-1)}\mathbf{W}_{r,k}^{(\ell)} + \mathbf{b}_{r,k}^{(\ell)}) + \mathbf{h}_{r,k}^{(\ell-1)}\Bigr);\,\varepsilon=10^{-6}\Biggr)
\end{equation}

where $\mathbf{W}_{r,k}^{(\ell)}\in\mathbb{R}^{D_r\times D_r}$ and $\mathbf{b}_{r,k}^{(\ell)}\in\mathbb{R}^{D_r}$ are learned, $\mathbf{h}_{r,k}^{(\ell-1)}$ is a skip-connection, $\mathrm{LayerNorm}$ ensures numerical stability, and $\sigma$ is the LeakyReLU activation.

\subsection{Mapping Theoretical Formulation to Codebase Implementation}
\label{app:codebase_mapping}
To ensure reproducibility and provide a clear bridge between theory and practice, this section details the correspondence between the theoretical \textsc{HOPSE} pipeline illustrated in Figure 2 and its modular implementation within the TopoBench framework. To strictly follow the modularity of the library, the execution is decoupled into distinct data transformation, encoder, backbone, and readout stages.

\paragraph{Steps A, B, and C: Preprocessing, Extracting $g_k$, and Aggregation $\psi_k$.}
The construction of Hasse graphs from rank neighborhoods (such as boundaries and coboundaries) and the subsequent extraction of topological descriptors occur entirely offline during the dataset transformation phase. All structural and positional encodings reside in the \texttt{topobench/transforms/data\_manipulations/} directory.

Crucially, the rank-targeted aggregation $\psi_k$ from Equation 2 is also precomputed during these dataset transforms rather than on the fly. The transformation classes concatenate the encoding outputs across different transmission routes targeting the same topological rank, producing the offline $\mathbf{X}_{r,k}$ tensors.

\paragraph{Step D: Initial Features $\varepsilon_r$ and Structural Transformations $f_{r,k}$.}
The processing of initial raw features (Equation 4) and the precomputed structural encodings (Equation 3) are handled by separate neural modules:
\begin{itemize}
    \item \textbf{Initial Feature Encoder ($\varepsilon_r$):} The raw cell attributes $\mathbf{Z}_r$ are embedded in \texttt{topobench/nn/encoders/hopse\_encoder.py}. This module dynamically maps the raw features using MLPs, optionally integrating domain-specific encoders.
    \item \textbf{Structural Backbone ($f_{r,k}$):} The precomputed encodings are transformed using independent MLPs within the \texttt{HOPSE} backbone class, found in \texttt{topobench/nn/backbones/combinatorial/hopse.py}. The backbone processes the complex layer by layer and rank by rank, outputting a nested tuple of tensors where the outer tuple indexes the complex dimension and the inner tuple indexes the feature hops.
\end{itemize}

\paragraph{Steps E and F: Rank-wise Fusion $\Theta_r$ and Global Readout $\varphi$.}
The final fusion and task prediction steps corresponding to Equations 5 and 6 are executed within the readout module. This is implemented in \texttt{topobench/nn/readouts/hopse.py}. 
\begin{itemize}
    \item \textbf{Rank-wise Fusion ($\Theta_r$):} Within the readout forward pass, the model collects the multi-hop embeddings output by the backbone for each specific rank, concatenates them, and passes them through a rank-specific MLP. This operation realizes the fusion function $\Theta_r$ to produce the neighborhood-aware representations $\mathbf{H}_r$.
    \item \textbf{Global Readout ($\varphi$):} Finally, the merged multi-rank representations are globally pooled across the batch and passed through a linear classifier to compute the downstream task logits, fulfilling the role of the global readout $\varphi$.
\end{itemize}

\section{Hardware Details}\label{app:hardware}
All of our experiments were executed on a Linux machine with 256 logical cores, 1.5TB of RAM, and 4 NVIDIA H100
GPUs, each with 94GB of GPU memory.
\section{Hyperparameter Search}\label{app:hyperparameter_search}

Following the standardized protocol established in \texttt{TopoBench} \citep{telyatnikov2025topobench}, we partition each dataset into five random splits (50\% training, 25\% validation, 25\% testing) and optimize the models using the Adam optimizer. 
To ensure a fair comparison, we employ an exhaustive grid-search strategy to optimize architecture hyperparameters across all model-dataset combinations. The common search space is defined as follows:
\begin{itemize}
    \item Encoder hidden dimension: $\{128, 256\}$
    \item Encoder dropout: $\{0.25, 0.5\}$
    \item Number of backbone layers: $\{1, 2, 4\}$
    \item Learning rate: $\{0.01, 0.001\}$
    \item Batch size: $\{128, 256\}$
\end{itemize}

Beyond this shared space, we include model-specific hyperparameters in the grid search where applicable:
\begin{itemize}
    \item \textbf{\textsc{HOPSE-GPSE}:} This variant utilizes \textsc{GPSE} to extract positional and structural information. Because the optimal initialization can be highly task-dependent, we treat the choice of the pre-training corpus (\textsc{ZINC} versus \textsc{MOLPCBA}) as an additional categorical hyperparameter.
    \item \textbf{\textsc{SANN}:} Following its original implementation, \textsc{SANN} requires defining a hop-count for its positional and structural encodings. We optimize this parameter within the discrete set $\{1, 2, 3\}$.
\end{itemize}
\section{Extended results}\label{app:extended_results}
This appendix provides a comprehensive detailing of the empirical evaluations discussed in \cref{sec:experiments}. The following subsections offer insights into training efficiency, preprocessing overhead, model capacity, and the impact of different architectural choices.

\subsection{Training Times}
The computational efficiency of the framework is analyzed through overall runtime and per-epoch speed. \cref{tbl:best_rerun_runtime} presents the end-to-end execution time in seconds, excluding preprocessing, across all datasets and domains. Notably, \textbf{HOPSE-GPSE} often achieves the lowest end-to-end times in complex datasets like \textsc{CYP3A4} and \textsc{NCI109}. Complementing this, \cref{tbl:best_rerun_epoch_time} details the training seconds per epoch. These results highlight that our proposed models maintain competitive per-epoch efficiency compared to established baselines like \textsc{TopoTune}, with \textbf{HOPSE-GPSE} frequently being the fastest model within the cellular and graph domains.

\definecolor{stdblue}{HTML}{C9DAF8}
\definecolor{bestgray}{HTML}{D9D9D9}
\begin{table}[t]
\caption{\textbf{End-to-End Training Times.} Total execution time (in seconds) excluding preprocessing overhead, reported as mean and standard deviation over random seeds. \protect\colorbox{bestgray}{\textbf{Bold}} highlights the lowest mean runtime per dataset within each respective domain. Cells in \protect\colorbox{stdblue}{blue} indicate models whose runtimes are not significantly slower than the optimal model (95\% confidence level, two-sided Z-test).}
\label{tbl:best_rerun_runtime}
\centering
\begin{adjustbox}{width=1.\textwidth}
\renewcommand{\arraystretch}{1.4}
\begin{tabular}{@{}llcccccccccccc@{}}
\toprule
  &  & \multicolumn{8}{c}{\mbox{Graph}} & \multicolumn{4}{c}{\mbox{Simplicial}} \\
\cmidrule(lr){3-10} \cmidrule(lr){11-14}
 & \textbf{Model} & \scriptsize MUTAG & \scriptsize PROTEINS & \scriptsize NCI1 & \scriptsize NCI109 & \scriptsize BBB & \scriptsize CYP3A4 & \scriptsize Cl.Hep. & \scriptsize Caco2 & \scriptsize NAME & \scriptsize ORIENT & \scriptsize $\beta_1$ & \scriptsize $\beta_2$ \\
\midrule
\multirow{3}{*}{\rotatebox[origin=c]{90}{\textbf{Graph}}} & GCN & {\cellcolor{bestgray}{\scriptsize\boldmath $2.00 \pm 0.00$}} & \cellcolor{stdblue}{\scriptsize $6.80 \pm 2.23$} & \cellcolor{stdblue}{\scriptsize $26.20 \pm 1.72$} & \cellcolor{stdblue}{\scriptsize $30.60 \pm 9.83$} & {\cellcolor{bestgray}{\scriptsize\boldmath $7.00 \pm 0.00$}} & {\cellcolor{bestgray}{\scriptsize\boldmath $131.00 \pm 39.34$}} & \cellcolor{stdblue}{\scriptsize $5.00 \pm 0.00$} & {\scriptsize $13.00 \pm 0.00$} & {\cellcolor{bestgray}{\scriptsize\boldmath $704.20 \pm 9.83$}} & {\cellcolor{bestgray}{\scriptsize\boldmath $709.80 \pm 16.34$}} & {\scriptsize $1589.00 \pm 519.71$} & {\scriptsize $1589.00 \pm 519.71$} \\
& GAT & {\cellcolor{bestgray}{\scriptsize\boldmath $2.00 \pm 0.00$}} & {\scriptsize $6.80 \pm 0.75$} & {\cellcolor{bestgray}{\scriptsize\boldmath $25.60 \pm 8.04$}} & \cellcolor{stdblue}{\scriptsize $38.40 \pm 13.71$} & {\scriptsize $9.00 \pm 0.00$} & \cellcolor{stdblue}{\scriptsize $134.80 \pm 20.55$} & {\scriptsize $12.00 \pm 2.45$} & {\scriptsize $13.00 \pm 0.00$} & {\scriptsize $778.60 \pm 38.06$} & {\scriptsize $743.40 \pm 32.20$} & {\cellcolor{bestgray}{\scriptsize\boldmath $778.00 \pm 102.89$}} & {\cellcolor{bestgray}{\scriptsize\boldmath $778.00 \pm 102.89$}} \\
& GIN & {\cellcolor{bestgray}{\scriptsize\boldmath $2.00 \pm 0.00$}} & {\cellcolor{bestgray}{\scriptsize\boldmath $5.60 \pm 1.02$}} & \cellcolor{stdblue}{\scriptsize $28.00 \pm 5.66$} & {\cellcolor{bestgray}{\scriptsize\boldmath $26.40 \pm 3.56$}} & {\scriptsize $18.40 \pm 3.38$} & \cellcolor{stdblue}{\scriptsize $147.20 \pm 27.88$} & {\cellcolor{bestgray}{\scriptsize\boldmath $4.60 \pm 0.80$}} & {\cellcolor{bestgray}{\scriptsize\boldmath $6.20 \pm 0.40$}} & {\scriptsize $725.80 \pm 17.90$} & \cellcolor{stdblue}{\scriptsize $727.80 \pm 45.71$} & {\scriptsize $1429.20 \pm 499.22$} & {\scriptsize $1429.20 \pm 499.22$} \\
\midrule
\multirow{6}{*}{\rotatebox[origin=c]{90}{\textbf{Simplicial}}} & TopoTune & {\scriptsize $10.40 \pm 2.42$} & {\cellcolor{bestgray}{\scriptsize\boldmath $29.20 \pm 3.87$}} & \cellcolor{stdblue}{\scriptsize $136.00 \pm 29.82$} & {\scriptsize $125.80 \pm 11.02$} & {\cellcolor{bestgray}{\scriptsize\boldmath $58.20 \pm 12.80$}} & {\scriptsize $623.40 \pm 61.25$} & {\cellcolor{bestgray}{\scriptsize\boldmath $39.80 \pm 9.26$}} & {\scriptsize $117.60 \pm 24.21$} & {\scriptsize $2416.60 \pm 1379.10$} & {\scriptsize $4086.20 \pm 615.51$} & {\scriptsize $3781.50 \pm 362.50$} & {\scriptsize $3781.50 \pm 362.50$} \\
& SCCNN & {\scriptsize $22.80 \pm 5.31$} & {\scriptsize $126.20 \pm 24.33$} & {\scriptsize $638.75 \pm 196.31$} & {\scriptsize $518.00 \pm 94.81$} & {\scriptsize $428.80 \pm 59.36$} & {\scriptsize $4476.20 \pm 560.78$} & {\scriptsize $130.75 \pm 2.68$} & {\scriptsize $543.60 \pm 26.30$} & {\scriptsize $10145.60 \pm 1374.85$} & {\scriptsize $12956.80 \pm 2198.37$} & {\scriptsize $16188.60 \pm 3117.89$} & {\scriptsize $16188.60 \pm 3117.89$} \\
& SANN & {\scriptsize $23.20 \pm 6.08$} & {\scriptsize $43.40 \pm 6.95$} & {\scriptsize $208.20 \pm 46.05$} & {\scriptsize $163.80 \pm 59.72$} & {\scriptsize $289.00 \pm 34.66$} & {\scriptsize $2675.40 \pm 435.20$} & {\scriptsize $339.60 \pm 123.26$} & {\scriptsize $309.60 \pm 43.63$} & {\scriptsize $1602.40 \pm 491.10$} & {\scriptsize $2170.40 \pm 959.63$} & \cellcolor{stdblue}{\scriptsize $2211.00 \pm 846.57$} & \cellcolor{stdblue}{\scriptsize $2211.00 \pm 846.57$} \\
& \textbf{HOPSE-M-F}  & \cellcolor{stdblue}{\scriptsize $9.00 \pm 3.74$} & {\scriptsize $38.40 \pm 8.80$} & {\scriptsize $289.40 \pm 48.44$} & {\scriptsize $234.20 \pm 54.24$} & \cellcolor{stdblue}{\scriptsize $65.00 \pm 12.03$} & {\scriptsize $597.20 \pm 121.72$} & {\scriptsize $118.60 \pm 6.68$} & \cellcolor{stdblue}{\scriptsize $94.60 \pm 17.55$} & {\cellcolor{bestgray}{\scriptsize\boldmath $718.60 \pm 183.92$}} & {\scriptsize $1920.80 \pm 783.10$} & \cellcolor{stdblue}{\scriptsize $3875.80 \pm 3553.00$} & \cellcolor{stdblue}{\scriptsize $3875.80 \pm 3553.00$} \\
& \textbf{HOPSE-M-C}  & {\scriptsize $8.80 \pm 1.72$} & {\scriptsize $49.80 \pm 13.26$} & {\cellcolor{bestgray}{\scriptsize\boldmath $133.40 \pm 27.23$}} & {\scriptsize $220.60 \pm 59.69$} & {\scriptsize $126.80 \pm 29.24$} & {\scriptsize $566.00 \pm 87.83$} & {\scriptsize $125.80 \pm 47.94$} & {\cellcolor{bestgray}{\scriptsize\boldmath $80.40 \pm 26.09$}} & {\scriptsize $2554.00 \pm 1072.81$} & {\scriptsize $2029.60 \pm 1208.73$} & {\cellcolor{bestgray}{\scriptsize\boldmath $1635.50 \pm 255.50$}} & {\cellcolor{bestgray}{\scriptsize\boldmath $1635.50 \pm 255.50$}} \\
& \textbf{HOPSE-GPSE}  & {\cellcolor{bestgray}{\scriptsize\boldmath $6.20 \pm 0.75$}} & {\scriptsize $41.60 \pm 12.31$} & \cellcolor{stdblue}{\scriptsize $139.00 \pm 18.51$} & {\cellcolor{bestgray}{\scriptsize\boldmath $98.20 \pm 13.12$}} & \cellcolor{stdblue}{\scriptsize $67.20 \pm 16.73$} & {\cellcolor{bestgray}{\scriptsize\boldmath $303.00 \pm 26.15$}} & {\scriptsize $99.80 \pm 16.41$} & {\scriptsize $122.20 \pm 31.82$} & \cellcolor{stdblue}{\scriptsize $808.00 \pm 385.35$} & {\cellcolor{bestgray}{\scriptsize\boldmath $639.00 \pm 202.55$}} & \cellcolor{stdblue}{\scriptsize $2046.00 \pm 1278.00$} & \cellcolor{stdblue}{\scriptsize $2046.00 \pm 1278.00$} \\
\midrule
\multirow{6}{*}{\rotatebox[origin=c]{90}{\textbf{Cell}}} & TopoTune & {\scriptsize $10.20 \pm 2.23$} & {\scriptsize $38.60 \pm 10.59$} & \cellcolor{stdblue}{\scriptsize $170.80 \pm 54.88$} & {\cellcolor{bestgray}{\scriptsize\boldmath $169.40 \pm 46.85$}} & {\cellcolor{bestgray}{\scriptsize\boldmath $44.60 \pm 1.62$}} & {\scriptsize $704.60 \pm 52.02$} & {\cellcolor{bestgray}{\scriptsize\boldmath $29.00 \pm 2.10$}} & {\cellcolor{bestgray}{\scriptsize\boldmath $80.60 \pm 15.27$}} & - & - & - & - \\
& CWN & {\scriptsize $8.80 \pm 1.17$} & {\scriptsize $45.20 \pm 9.09$} & {\scriptsize $340.20 \pm 55.09$} & {\scriptsize $305.60 \pm 46.20$} & {\scriptsize $158.20 \pm 10.85$} & {\scriptsize $1259.20 \pm 192.99$} & {\scriptsize $85.40 \pm 15.13$} & {\scriptsize $180.80 \pm 2.04$} & - & - & - & - \\
& CCCN & {\scriptsize $8.40 \pm 1.62$} & {\scriptsize $46.40 \pm 13.51$} & {\scriptsize $226.00 \pm 25.55$} & {\scriptsize $225.60 \pm 37.57$} & {\scriptsize $81.20 \pm 1.72$} & {\scriptsize $1063.40 \pm 146.32$} & {\scriptsize $46.80 \pm 0.40$} & {\scriptsize $148.20 \pm 2.93$} & - & - & - & - \\
& \textbf{HOPSE-M-F}  & {\scriptsize $9.00 \pm 0.63$} & \cellcolor{stdblue}{\scriptsize $35.00 \pm 9.74$} & {\scriptsize $203.40 \pm 39.15$} & \cellcolor{stdblue}{\scriptsize $169.60 \pm 27.56$} & {\scriptsize $64.80 \pm 5.71$} & \cellcolor{stdblue}{\scriptsize $386.40 \pm 53.13$} & {\scriptsize $62.20 \pm 22.86$} & {\scriptsize $114.40 \pm 14.75$} & - & - & - & - \\
& \textbf{HOPSE-M-C}  & \cellcolor{stdblue}{\scriptsize $6.00 \pm 0.89$} & {\scriptsize $55.20 \pm 10.91$} & {\scriptsize $198.40 \pm 46.99$} & \cellcolor{stdblue}{\scriptsize $263.60 \pm 126.03$} & {\scriptsize $75.80 \pm 6.71$} & {\scriptsize $745.40 \pm 204.94$} & {\scriptsize $44.60 \pm 1.50$} & {\scriptsize $138.20 \pm 44.17$} & - & - & - & - \\
& \textbf{HOPSE-GPSE}  & {\cellcolor{bestgray}{\scriptsize\boldmath $5.20 \pm 0.98$}} & {\cellcolor{bestgray}{\scriptsize\boldmath $28.00 \pm 5.06$}} & {\cellcolor{bestgray}{\scriptsize\boldmath $146.20 \pm 11.36$}} & \cellcolor{stdblue}{\scriptsize $172.00 \pm 23.87$} & {\scriptsize $73.00 \pm 12.73$} & {\cellcolor{bestgray}{\scriptsize\boldmath $370.40 \pm 42.20$}} & {\scriptsize $63.40 \pm 4.59$} & \cellcolor{stdblue}{\scriptsize $87.60 \pm 20.13$} & - & - & - & - \\
\bottomrule
\end{tabular}
\end{adjustbox}
\end{table}
\definecolor{stdblue}{HTML}{C9DAF8}
\definecolor{bestgray}{HTML}{D9D9D9}
\begin{table}[t]
\caption{\textbf{Per-Epoch Training Times.} Mean training time per epoch (in seconds) and standard deviation across datasets and structural domains.\protect\colorbox{bestgray}{\textbf{Bold}} values indicate the lowest mean time per dataset within a given domain (Graph, Simplicial, or Cell). Cells highlighted in \protect\colorbox{stdblue}{blue} denote configurations that are not significantly slower than the best performing model based on a two-sided Z-test at a 95\% confidence level.}
\label{tbl:best_rerun_epoch_time}
\centering
\begin{adjustbox}{width=1.\textwidth}
\renewcommand{\arraystretch}{1.4}
\begin{tabular}{@{}llcccccccccccc@{}}
\toprule
  &  & \multicolumn{8}{c}{\mbox{Graph}} & \multicolumn{4}{c}{\mbox{Simplicial}} \\
\cmidrule(lr){3-10} \cmidrule(lr){11-14}
 & \textbf{Model} & \scriptsize MUTAG & \scriptsize PROTEINS & \scriptsize NCI1 & \scriptsize NCI109 & \scriptsize BBB & \scriptsize CYP3A4 & \scriptsize Cl.Hep. & \scriptsize Caco2 & \scriptsize NAME & \scriptsize ORIENT & \scriptsize $\beta_1$ & \scriptsize $\beta_2$ \\
\midrule
\multirow{3}{*}{\rotatebox[origin=c]{90}{\textbf{Graph}}} & GCN & \cellcolor{stdblue}{\scriptsize $0.02 \pm 0.01$} & \cellcolor{stdblue}{\scriptsize $0.07 \pm 0.02$} & \cellcolor{stdblue}{\scriptsize $0.27 \pm 0.04$} & \cellcolor{stdblue}{\scriptsize $0.26 \pm 0.05$} & \cellcolor{stdblue}{\scriptsize $0.14 \pm 0.04$} & {\cellcolor{bestgray}{\scriptsize\boldmath $0.85 \pm 0.07$}} & \cellcolor{stdblue}{\scriptsize $0.08 \pm 0.01$} & \cellcolor{stdblue}{\scriptsize $0.07 \pm 0.02$} & {\cellcolor{bestgray}{\scriptsize\boldmath $13.54 \pm 2.34$}} & {\cellcolor{bestgray}{\scriptsize\boldmath $13.25 \pm 2.32$}} & \cellcolor{stdblue}{\scriptsize $14.40 \pm 2.62$} & \cellcolor{stdblue}{\scriptsize $14.40 \pm 2.62$} \\
& GAT & {\cellcolor{bestgray}{\scriptsize\boldmath $0.02 \pm 0.01$}} & \cellcolor{stdblue}{\scriptsize $0.07 \pm 0.02$} & \cellcolor{stdblue}{\scriptsize $0.27 \pm 0.05$} & \cellcolor{stdblue}{\scriptsize $0.34 \pm 0.06$} & {\cellcolor{bestgray}{\scriptsize\boldmath $0.13 \pm 0.05$}} & \cellcolor{stdblue}{\scriptsize $1.04 \pm 0.08$} & \cellcolor{stdblue}{\scriptsize $0.12 \pm 0.01$} & \cellcolor{stdblue}{\scriptsize $0.07 \pm 0.01$} & \cellcolor{stdblue}{\scriptsize $14.62 \pm 2.52$} & \cellcolor{stdblue}{\scriptsize $14.30 \pm 2.57$} & {\cellcolor{bestgray}{\scriptsize\boldmath $14.33 \pm 2.65$}} & {\cellcolor{bestgray}{\scriptsize\boldmath $14.33 \pm 2.65$}} \\
& GIN & \cellcolor{stdblue}{\scriptsize $0.02 \pm 0.01$} & {\cellcolor{bestgray}{\scriptsize\boldmath $0.06 \pm 0.02$}} & {\cellcolor{bestgray}{\scriptsize\boldmath $0.23 \pm 0.05$}} & {\cellcolor{bestgray}{\scriptsize\boldmath $0.23 \pm 0.04$}} & \cellcolor{stdblue}{\scriptsize $0.19 \pm 0.03$} & \cellcolor{stdblue}{\scriptsize $0.88 \pm 0.07$} & {\cellcolor{bestgray}{\scriptsize\boldmath $0.08 \pm 0.03$}} & {\cellcolor{bestgray}{\scriptsize\boldmath $0.05 \pm 0.02$}} & \cellcolor{stdblue}{\scriptsize $14.55 \pm 2.59$} & \cellcolor{stdblue}{\scriptsize $13.34 \pm 2.91$} & \cellcolor{stdblue}{\scriptsize $14.37 \pm 2.59$} & \cellcolor{stdblue}{\scriptsize $14.37 \pm 2.59$} \\
\midrule
\multirow{6}{*}{\rotatebox[origin=c]{90}{\textbf{Simplicial}}} & TopoTune & \cellcolor{stdblue}{\scriptsize $0.06 \pm 0.03$} & {\cellcolor{bestgray}{\scriptsize\boldmath $0.28 \pm 0.05$}} & \cellcolor{stdblue}{\scriptsize $0.74 \pm 0.13$} & \cellcolor{stdblue}{\scriptsize $0.73 \pm 0.13$} & \cellcolor{stdblue}{\scriptsize $0.53 \pm 0.05$} & {\scriptsize $3.34 \pm 0.21$} & {\cellcolor{bestgray}{\scriptsize\boldmath $0.33 \pm 0.06$}} & \cellcolor{stdblue}{\scriptsize $0.32 \pm 0.04$} & \cellcolor{stdblue}{\scriptsize $8.91 \pm 1.61$} & \cellcolor{stdblue}{\scriptsize $10.75 \pm 1.78$} & {\scriptsize $13.82 \pm 1.98$} & {\scriptsize $13.82 \pm 1.98$} \\
& SCCNN & {\scriptsize $0.17 \pm 0.06$} & {\scriptsize $1.18 \pm 0.22$} & {\scriptsize $3.71 \pm 0.95$} & {\scriptsize $3.43 \pm 0.83$} & {\scriptsize $2.68 \pm 0.52$} & {\scriptsize $16.17 \pm 2.05$} & {\scriptsize $1.38 \pm 0.37$} & {\scriptsize $1.02 \pm 0.17$} & {\scriptsize $56.64 \pm 11.38$} & {\scriptsize $51.79 \pm 9.68$} & {\scriptsize $71.57 \pm 10.90$} & {\scriptsize $71.57 \pm 10.90$} \\
& SANN & \cellcolor{stdblue}{\scriptsize $0.16 \pm 0.06$} & \cellcolor{stdblue}{\scriptsize $0.35 \pm 0.09$} & {\scriptsize $1.18 \pm 0.21$} & {\scriptsize $1.26 \pm 0.22$} & {\scriptsize $2.28 \pm 0.51$} & {\scriptsize $15.49 \pm 2.37$} & {\scriptsize $1.40 \pm 0.37$} & {\scriptsize $0.95 \pm 0.16$} & {\scriptsize $14.49 \pm 2.54$} & {\scriptsize $14.48 \pm 2.78$} & {\scriptsize $15.77 \pm 2.96$} & {\scriptsize $15.77 \pm 2.96$} \\
& \textbf{HOPSE-M-F}  & \cellcolor{stdblue}{\scriptsize $0.06 \pm 0.02$} & \cellcolor{stdblue}{\scriptsize $0.28 \pm 0.08$} & {\scriptsize $1.43 \pm 0.24$} & {\scriptsize $1.14 \pm 0.19$} & \cellcolor{stdblue}{\scriptsize $0.58 \pm 0.09$} & {\scriptsize $3.38 \pm 0.23$} & \cellcolor{stdblue}{\scriptsize $0.40 \pm 0.05$} & {\cellcolor{bestgray}{\scriptsize\boldmath $0.25 \pm 0.04$}} & \cellcolor{stdblue}{\scriptsize $9.04 \pm 1.44$} & {\scriptsize $13.10 \pm 1.79$} & \cellcolor{stdblue}{\scriptsize $14.47 \pm 4.44$} & \cellcolor{stdblue}{\scriptsize $14.47 \pm 4.44$} \\
& \textbf{HOPSE-M-C}  & \cellcolor{stdblue}{\scriptsize $0.06 \pm 0.03$} & \cellcolor{stdblue}{\scriptsize $0.36 \pm 0.10$} & \cellcolor{stdblue}{\scriptsize $0.89 \pm 0.15$} & \cellcolor{stdblue}{\scriptsize $0.95 \pm 0.20$} & {\scriptsize $0.76 \pm 0.07$} & {\scriptsize $4.22 \pm 0.38$} & {\scriptsize $0.49 \pm 0.05$} & \cellcolor{stdblue}{\scriptsize $0.27 \pm 0.04$} & \cellcolor{stdblue}{\scriptsize $13.76 \pm 4.53$} & {\scriptsize $13.91 \pm 2.28$} & \cellcolor{stdblue}{\scriptsize $11.92 \pm 2.41$} & \cellcolor{stdblue}{\scriptsize $11.92 \pm 2.41$} \\
& \textbf{HOPSE-GPSE}  & {\cellcolor{bestgray}{\scriptsize\boldmath $0.04 \pm 0.02$}} & \cellcolor{stdblue}{\scriptsize $0.38 \pm 0.08$} & {\cellcolor{bestgray}{\scriptsize\boldmath $0.60 \pm 0.14$}} & {\cellcolor{bestgray}{\scriptsize\boldmath $0.56 \pm 0.13$}} & {\cellcolor{bestgray}{\scriptsize\boldmath $0.45 \pm 0.04$}} & {\cellcolor{bestgray}{\scriptsize\boldmath $2.38 \pm 0.27$}} & \cellcolor{stdblue}{\scriptsize $0.33 \pm 0.07$} & \cellcolor{stdblue}{\scriptsize $0.31 \pm 0.04$} & {\cellcolor{bestgray}{\scriptsize\boldmath $6.74 \pm 1.11$}} & {\cellcolor{bestgray}{\scriptsize\boldmath $6.67 \pm 1.33$}} & {\cellcolor{bestgray}{\scriptsize\boldmath $8.94 \pm 1.37$}} & {\cellcolor{bestgray}{\scriptsize\boldmath $8.94 \pm 1.37$}} \\
\midrule
\multirow{6}{*}{\rotatebox[origin=c]{90}{\textbf{Cell}}} & TopoTune & \cellcolor{stdblue}{\scriptsize $0.07 \pm 0.02$} & \cellcolor{stdblue}{\scriptsize $0.28 \pm 0.05$} & \cellcolor{stdblue}{\scriptsize $0.93 \pm 0.16$} & \cellcolor{stdblue}{\scriptsize $0.94 \pm 0.16$} & \cellcolor{stdblue}{\scriptsize $0.54 \pm 0.05$} & {\scriptsize $4.13 \pm 0.25$} & \cellcolor{stdblue}{\scriptsize $0.31 \pm 0.06$} & \cellcolor{stdblue}{\scriptsize $0.23 \pm 0.03$} & - & - & - & - \\
& CWN & \cellcolor{stdblue}{\scriptsize $0.06 \pm 0.02$} & \cellcolor{stdblue}{\scriptsize $0.35 \pm 0.09$} & \cellcolor{stdblue}{\scriptsize $1.35 \pm 0.21$} & {\scriptsize $1.42 \pm 0.22$} & {\scriptsize $0.81 \pm 0.06$} & {\scriptsize $5.19 \pm 0.28$} & {\scriptsize $0.47 \pm 0.04$} & {\scriptsize $0.36 \pm 0.04$} & - & - & - & - \\
& CCCN & \cellcolor{stdblue}{\scriptsize $0.07 \pm 0.02$} & \cellcolor{stdblue}{\scriptsize $0.32 \pm 0.06$} & \cellcolor{stdblue}{\scriptsize $1.33 \pm 0.22$} & \cellcolor{stdblue}{\scriptsize $1.26 \pm 0.21$} & {\scriptsize $0.79 \pm 0.06$} & {\scriptsize $4.50 \pm 0.28$} & {\scriptsize $0.45 \pm 0.08$} & {\scriptsize $0.32 \pm 0.03$} & - & - & - & - \\
& \textbf{HOPSE-M-F}  & \cellcolor{stdblue}{\scriptsize $0.06 \pm 0.02$} & \cellcolor{stdblue}{\scriptsize $0.34 \pm 0.09$} & \cellcolor{stdblue}{\scriptsize $1.11 \pm 0.20$} & \cellcolor{stdblue}{\scriptsize $1.19 \pm 0.23$} & {\cellcolor{bestgray}{\scriptsize\boldmath $0.51 \pm 0.05$}} & \cellcolor{stdblue}{\scriptsize $2.92 \pm 0.25$} & {\scriptsize $0.42 \pm 0.07$} & \cellcolor{stdblue}{\scriptsize $0.23 \pm 0.04$} & - & - & - & - \\
& \textbf{HOPSE-M-C}  & \cellcolor{stdblue}{\scriptsize $0.05 \pm 0.02$} & \cellcolor{stdblue}{\scriptsize $0.38 \pm 0.11$} & \cellcolor{stdblue}{\scriptsize $1.00 \pm 0.20$} & \cellcolor{stdblue}{\scriptsize $1.42 \pm 0.35$} & \cellcolor{stdblue}{\scriptsize $0.59 \pm 0.09$} & {\scriptsize $4.61 \pm 0.27$} & \cellcolor{stdblue}{\scriptsize $0.36 \pm 0.07$} & {\scriptsize $0.40 \pm 0.05$} & - & - & - & - \\
& \textbf{HOPSE-GPSE}  & {\cellcolor{bestgray}{\scriptsize\boldmath $0.04 \pm 0.01$}} & {\cellcolor{bestgray}{\scriptsize\boldmath $0.25 \pm 0.07$}} & {\cellcolor{bestgray}{\scriptsize\boldmath $0.85 \pm 0.15$}} & {\cellcolor{bestgray}{\scriptsize\boldmath $0.85 \pm 0.19$}} & \cellcolor{stdblue}{\scriptsize $0.60 \pm 0.09$} & {\cellcolor{bestgray}{\scriptsize\boldmath $2.72 \pm 0.19$}} & {\cellcolor{bestgray}{\scriptsize\boldmath $0.23 \pm 0.06$}} & {\cellcolor{bestgray}{\scriptsize\boldmath $0.22 \pm 0.03$}} & - & - & - & - \\
\bottomrule
\end{tabular}
\end{adjustbox}
\end{table}

\subsection{Preprocessing Times}
\cref{tbl:preprocess_ablation_preproc_time} outlines the preprocessing times required for different neighborhood configurations in both the cellular and simplicial domains. As expected, increasing the complexity of the neighborhood function (e.g., from $\emph{Adj-1}$ to $\emph{Adj-3}$ or $\emph{Inc-2}$) generally leads to higher preprocessing overhead. 

\definecolor{bestgray}{HTML}{D9D9D9}
\begin{table}[t]
\caption{\textbf{Preprocessing Time Overhead.} Total preprocessing time (in seconds) required for various neighborhood configurations across Graph and Simplicial domains. \protect\colorbox{bestgray}{\textbf{Bold}} entries represent the fastest preprocessing configuration for each dataset within its respective domain.}
\label{tbl:preprocess_ablation_preproc_time}
\centering
\begin{adjustbox}{width=1.\textwidth}
\begin{tabular}{@{}lllcccccccccccc@{}}
\toprule
 & & & \multicolumn{8}{c}{\mbox{Graph}} & \multicolumn{4}{c}{\mbox{Simplicial}} \\
\cmidrule(lr){4-11} \cmidrule(lr){12-15}
 & \textbf{Model} & \textbf{Neigh.} & \scriptsize MUTAG & \scriptsize PROTEINS & \scriptsize NCI1 & \scriptsize NCI109 & \scriptsize BBB & \scriptsize CYP3A4 & \scriptsize Cl.Hep. & \scriptsize Caco2 & \scriptsize NAME & \scriptsize ORIENT & \scriptsize $\beta_1$ & \scriptsize $\beta_2$ \\
\midrule
\multirow{15}{*}{\rotatebox[origin=c]{90}{\textbf{Simplicial}}} & \multirow{5}{*}{\textbf{HOPSE-M-C}} & $\emph{Adj-1}$ & {\scriptsize 2.23} & {\scriptsize 20.45} & {\scriptsize 41.86} & {\scriptsize 41.71} & {\cellcolor{bestgray}{\scriptsize\textbf{20.24}}} & {\cellcolor{bestgray}{\scriptsize\textbf{120.73}}} & {\scriptsize 12.95} & {\scriptsize 10.07} & {\cellcolor{bestgray}{\scriptsize\textbf{203.74}}} & {\cellcolor{bestgray}{\scriptsize\textbf{203.24}}} & {\cellcolor{bestgray}{\scriptsize\textbf{203.81}}} & {\cellcolor{bestgray}{\scriptsize\textbf{203.81}}} \\
 &  & $\emph{Adj-2}$ & {\scriptsize 2.45} & {\scriptsize 26.07} & {\scriptsize 47.18} & {\scriptsize 47.22} & {\scriptsize 22.65} & {\scriptsize 136.20} & {\scriptsize 14.68} & {\scriptsize 11.19} & {\scriptsize 373.93} & {\scriptsize 378.69} & {\scriptsize 372.46} & {\scriptsize 372.46} \\
 &  & $\emph{Adj-3}$ & {\scriptsize 3.37} & {\scriptsize 52.61} & {\scriptsize 70.39} & {\scriptsize 70.40} & {\scriptsize 33.46} & {\scriptsize 204.20} & {\scriptsize 21.51} & {\scriptsize 16.58} & {\scriptsize 1141.67} & {\scriptsize 1143.81} & {\scriptsize 1138.54} & {\scriptsize 1138.54} \\
 &  & $\emph{Inc-1}$ & {\scriptsize 2.29} & {\scriptsize 120.36} & {\scriptsize 78.79} & {\scriptsize 80.29} & {\scriptsize 24.93} & {\scriptsize 133.20} & {\scriptsize 15.96} & {\scriptsize 11.57} & {\scriptsize 395.12} & {\scriptsize 394.77} & {\scriptsize 393.15} & {\scriptsize 393.15} \\
 &  & $\emph{Inc-2}$ & {\scriptsize 3.48} & {\scriptsize 355.84} & {\scriptsize 142.45} & {\scriptsize 144.25} & {\scriptsize 42.48} & {\scriptsize 222.65} & {\scriptsize 26.65} & {\scriptsize 19.09} & {\scriptsize 1114.51} & {\scriptsize 1110.81} & {\scriptsize 1112.50} & {\scriptsize 1112.50} \\
\arrayrulecolor{black!30}\cmidrule[0.25pt](lr){2-15}\arrayrulecolor{black}
 & \multirow{5}{*}{\textbf{HOPSE-M-F}} & $\emph{Adj-1}$ & {\scriptsize 2.21} & {\cellcolor{bestgray}{\scriptsize\textbf{18.97}}} & {\cellcolor{bestgray}{\scriptsize\textbf{41.59}}} & {\cellcolor{bestgray}{\scriptsize\textbf{41.35}}} & {\scriptsize 20.51} & {\scriptsize 122.00} & {\cellcolor{bestgray}{\scriptsize\textbf{12.82}}} & {\cellcolor{bestgray}{\scriptsize\textbf{9.92}}} & {\scriptsize 218.66} & {\scriptsize 217.29} & {\scriptsize 217.72} & {\scriptsize 217.72} \\
 &  & $\emph{Adj-2}$ & {\scriptsize 2.41} & {\scriptsize 24.05} & {\scriptsize 46.34} & {\scriptsize 46.61} & {\scriptsize 22.74} & {\scriptsize 137.62} & {\scriptsize 14.76} & {\scriptsize 11.14} & {\scriptsize 404.62} & {\scriptsize 404.11} & {\scriptsize 403.82} & {\scriptsize 403.82} \\
 &  & $\emph{Adj-3}$ & {\scriptsize 3.37} & {\scriptsize 43.39} & {\scriptsize 68.56} & {\scriptsize 69.27} & {\scriptsize 33.76} & {\scriptsize 203.63} & {\scriptsize 21.15} & {\scriptsize 16.35} & {\scriptsize 1148.18} & {\scriptsize 1155.15} & {\scriptsize 1153.25} & {\scriptsize 1153.25} \\
 &  & $\emph{Inc-1}$ & {\scriptsize 2.29} & {\scriptsize 24.09} & {\scriptsize 43.43} & {\scriptsize 43.50} & {\scriptsize 21.53} & {\scriptsize 128.96} & {\scriptsize 13.90} & {\scriptsize 10.54} & {\scriptsize 414.73} & {\scriptsize 417.95} & {\scriptsize 417.93} & {\scriptsize 417.93} \\
 &  & $\emph{Inc-2}$ & {\scriptsize 3.48} & {\scriptsize 43.90} & {\scriptsize 71.42} & {\scriptsize 71.81} & {\scriptsize 35.43} & {\scriptsize 214.91} & {\scriptsize 22.29} & {\scriptsize 16.96} & {\scriptsize 1173.56} & {\scriptsize 1175.86} & {\scriptsize 1173.30} & {\scriptsize 1173.30} \\
\arrayrulecolor{black!30}\cmidrule[0.25pt](lr){2-15}\arrayrulecolor{black}
 & \multirow{5}{*}{\textbf{HOPSE-GPSE}} & $\emph{Adj-1}$ & {\cellcolor{bestgray}{\scriptsize\textbf{2.11}}} & {\scriptsize 20.91} & {\scriptsize 48.55} & {\scriptsize 48.34} & {\scriptsize 23.45} & {\scriptsize 143.45} & {\scriptsize 14.65} & {\scriptsize 11.12} & {\scriptsize 274.30} & {\scriptsize 271.71} & {\scriptsize 272.62} & {\scriptsize 272.62} \\
 &  & $\emph{Adj-2}$ & {\scriptsize 3.11} & {\scriptsize 28.11} & {\scriptsize 72.43} & {\scriptsize 72.16} & {\scriptsize 34.43} & {\scriptsize 212.86} & {\scriptsize 22.07} & {\scriptsize 16.34} & {\scriptsize 520.88} & {\scriptsize 517.87} & {\scriptsize 516.67} & {\scriptsize 516.67} \\
 &  & $\emph{Adj-3}$ & {\scriptsize 5.20} & {\scriptsize 55.58} & {\scriptsize 120.30} & {\scriptsize 120.82} & {\scriptsize 57.90} & {\scriptsize 358.40} & {\scriptsize 35.98} & {\scriptsize 27.17} & {\scriptsize 1600.21} & {\scriptsize 1604.60} & {\scriptsize 1609.09} & {\scriptsize 1609.09} \\
 &  & $\emph{Inc-1}$ & {\scriptsize 2.22} & {\scriptsize 28.22} & {\scriptsize 51.74} & {\scriptsize 51.48} & {\scriptsize 24.77} & {\scriptsize 152.30} & {\scriptsize 16.09} & {\scriptsize 12.04} & {\scriptsize 577.57} & {\scriptsize 573.48} & {\scriptsize 575.92} & {\scriptsize 575.92} \\
 &  & $\emph{Inc-2}$ & {\scriptsize 5.32} & {\scriptsize 56.71} & {\scriptsize 122.73} & {\scriptsize 123.85} & {\scriptsize 59.58} & {\scriptsize 365.62} & {\scriptsize 37.47} & {\scriptsize 28.06} & {\scriptsize 1690.07} & {\scriptsize 1667.08} & {\scriptsize 1674.69} & {\scriptsize 1674.69} \\
 \midrule
\multirow{15}{*}{\rotatebox[origin=c]{90}{\textbf{Cell}}} & \multirow{5}{*}{\textbf{HOPSE-M-C}} & $\emph{Adj-1}$ & {\scriptsize 2.49} & {\scriptsize 17.42} & {\scriptsize 46.40} & {\scriptsize 46.39} & {\cellcolor{bestgray}{\scriptsize\textbf{22.12}}} & {\scriptsize 137.33} & {\scriptsize 14.12} & {\scriptsize 11.03} & {\scriptsize -} & {\scriptsize -} & {\scriptsize -} & {\scriptsize -} \\
 &  & $\emph{Adj-2}$ & {\scriptsize 3.26} & {\scriptsize 23.25} & {\scriptsize 63.34} & {\scriptsize 63.37} & {\scriptsize 31.22} & {\scriptsize 180.49} & {\scriptsize 19.12} & {\scriptsize 14.90} & {\scriptsize -} & {\scriptsize -} & {\scriptsize -} & {\scriptsize -} \\
 &  & $\emph{Adj-3}$ & {\scriptsize 5.83} & {\scriptsize 52.12} & {\scriptsize 122.03} & {\scriptsize 123.01} & {\scriptsize 58.90} & {\scriptsize 350.10} & {\scriptsize 36.17} & {\scriptsize 26.35} & {\scriptsize -} & {\scriptsize -} & {\scriptsize -} & {\scriptsize -} \\
 &  & $\emph{Inc-1}$ & {\scriptsize 3.30} & {\scriptsize 125.37} & {\scriptsize 98.22} & {\scriptsize 100.07} & {\scriptsize 35.29} & {\scriptsize 187.67} & {\scriptsize 21.39} & {\scriptsize 14.96} & {\scriptsize -} & {\scriptsize -} & {\scriptsize -} & {\scriptsize -} \\
 &  & $\emph{Inc-2}$ & {\scriptsize 6.25} & {\scriptsize 385.47} & {\scriptsize 200.07} & {\scriptsize 203.05} & {\scriptsize 72.41} & {\scriptsize 385.85} & {\scriptsize 43.53} & {\scriptsize 30.09} & {\scriptsize -} & {\scriptsize -} & {\scriptsize -} & {\scriptsize -} \\
\arrayrulecolor{black!30}\cmidrule[0.25pt](lr){2-15}\arrayrulecolor{black}
 & \multirow{5}{*}{\textbf{HOPSE-M-F}} & $\emph{Adj-1}$ & {\scriptsize 2.46} & {\cellcolor{bestgray}{\scriptsize\textbf{15.99}}} & {\cellcolor{bestgray}{\scriptsize\textbf{45.91}}} & {\cellcolor{bestgray}{\scriptsize\textbf{46.24}}} & {\scriptsize 22.68} & {\cellcolor{bestgray}{\scriptsize\textbf{135.49}}} & {\cellcolor{bestgray}{\scriptsize\textbf{14.08}}} & {\cellcolor{bestgray}{\scriptsize\textbf{10.62}}} & {\scriptsize -} & {\scriptsize -} & {\scriptsize -} & {\scriptsize -} \\
 &  & $\emph{Adj-2}$ & {\scriptsize 3.33} & {\scriptsize 20.99} & {\scriptsize 63.87} & {\scriptsize 63.87} & {\scriptsize 31.42} & {\scriptsize 189.15} & {\scriptsize 19.60} & {\scriptsize 14.36} & {\scriptsize -} & {\scriptsize -} & {\scriptsize -} & {\scriptsize -} \\
 &  & $\emph{Adj-3}$ & {\scriptsize 6.21} & {\scriptsize 39.96} & {\scriptsize 125.51} & {\scriptsize 127.02} & {\scriptsize 61.18} & {\scriptsize 372.06} & {\scriptsize 38.19} & {\scriptsize 27.46} & {\scriptsize -} & {\scriptsize -} & {\scriptsize -} & {\scriptsize -} \\
 &  & $\emph{Inc-1}$ & {\scriptsize 3.35} & {\scriptsize 21.15} & {\scriptsize 64.28} & {\scriptsize 65.04} & {\scriptsize 31.90} & {\scriptsize 190.69} & {\scriptsize 19.77} & {\scriptsize 14.60} & {\scriptsize -} & {\scriptsize -} & {\scriptsize -} & {\scriptsize -} \\
 &  & $\emph{Inc-2}$ & {\scriptsize 6.61} & {\scriptsize 40.71} & {\scriptsize 135.26} & {\scriptsize 136.59} & {\scriptsize 65.96} & {\scriptsize 405.77} & {\scriptsize 41.34} & {\scriptsize 29.73} & {\scriptsize -} & {\scriptsize -} & {\scriptsize -} & {\scriptsize -} \\
\arrayrulecolor{black!30}\cmidrule[0.25pt](lr){2-15}\arrayrulecolor{black}
 & \multirow{5}{*}{\textbf{HOPSE-GPSE}} & $\emph{Adj-1}$ & {\cellcolor{bestgray}{\scriptsize\textbf{2.36}}} & {\scriptsize 18.06} & {\scriptsize 52.74} & {\scriptsize 53.06} & {\scriptsize 26.01} & {\scriptsize 155.48} & {\scriptsize 15.91} & {\scriptsize 11.88} & {\scriptsize -} & {\scriptsize -} & {\scriptsize -} & {\scriptsize -} \\
 &  & $\emph{Adj-2}$ & {\scriptsize 3.53} & {\scriptsize 25.25} & {\scriptsize 79.38} & {\scriptsize 79.05} & {\scriptsize 38.58} & {\scriptsize 232.78} & {\scriptsize 23.55} & {\scriptsize 17.44} & {\scriptsize -} & {\scriptsize -} & {\scriptsize -} & {\scriptsize -} \\
 &  & $\emph{Adj-3}$ & {\scriptsize 7.59} & {\scriptsize 52.88} & {\scriptsize 167.96} & {\scriptsize 169.83} & {\scriptsize 81.77} & {\scriptsize 507.12} & {\scriptsize 51.65} & {\scriptsize 37.22} & {\scriptsize -} & {\scriptsize -} & {\scriptsize -} & {\scriptsize -} \\
 &  & $\emph{Inc-1}$ & {\scriptsize 3.61} & {\scriptsize 25.46} & {\scriptsize 79.44} & {\scriptsize 79.68} & {\scriptsize 38.18} & {\scriptsize 234.28} & {\scriptsize 23.99} & {\scriptsize 17.37} & {\scriptsize -} & {\scriptsize -} & {\scriptsize -} & {\scriptsize -} \\
 &  & $\emph{Inc-2}$ & {\scriptsize 8.13} & {\scriptsize 54.16} & {\scriptsize 178.47} & {\scriptsize 180.03} & {\scriptsize 87.05} & {\scriptsize 534.79} & {\scriptsize 53.70} & {\scriptsize 39.10} & {\scriptsize -} & {\scriptsize -} & {\scriptsize -} & {\scriptsize -} \\
\bottomrule
\end{tabular}
\end{adjustbox}
\end{table}

\subsection{Model Parameters}
\cref{tbl:model_parameters} reports the parameter counts for the optimal model configurations identified via exhaustive grid search. We observe that baseline GNNs often select configurations with significantly fewer parameters than HOPSE or high-order baselines across several datasets.  

While this creates a numerical discrepancy, the comparison remains fundamentally fair because each model was evaluated at its own performance-optimized capacity. The fact that GNNs "prefer" smaller configurations suggests a limit in their effective capacity; for these specific datasets, increasing the parameter count in a standard GNN does not necessarily translate to better generalization, likely due to the inherent over-smoothing or structural bottlenecks present in lower-order message passing. 

In contrast, while HOPSE models can exhibit larger parameter counts, they stay within comparable scales to other high-capacity topological models like \textsc{SCCNN} or simplicial \textsc{TopoTune}. This indicates that the added complexity of HOPSE is effectively utilized to capture higher-order features that simpler models cannot leverage, even if their parameter budgets were increased. 

\begin{table}[t]
\caption{\textbf{Model Parameter Capacity.} Number of parameters for the optimal model configurations identified via exhaustive grid search across all evaluated datasets and domains. We report $\beta_1$ and $\beta_2$ together since they are obtained from the same model.}
\label{tbl:model_parameters}
\centering
\begin{adjustbox}{width=1.\textwidth}
\renewcommand{\arraystretch}{1.2}
\begin{tabular}{@{}llccccccccccc@{}}
\toprule
 & & \multicolumn{8}{c}{\mbox{Graph}} & \multicolumn{3}{c}{\mbox{Simplicial}} \\
\cmidrule(lr){3-10} \cmidrule(lr){11-13}
 & Model & \scriptsize MUTAG & \scriptsize PROTEINS & \scriptsize NCI1 & \scriptsize NCI109 & \scriptsize BBB & \scriptsize CYP3A4 & \scriptsize Cl.Hep. & \scriptsize Caco2 & \scriptsize NAME & \scriptsize ORIENT & \scriptsize $\beta_1,\beta_2$ \\
\midrule
\multirow{3}{*}{\rotatebox[origin=c]{90}{\textbf{Graph}}} 
 & GCN & \scriptsize 81.1k & \scriptsize 17.5k & \scriptsize 77.7k & \scriptsize 77.8k & \scriptsize 74.0k & \scriptsize 143.5k & \scriptsize 274.8k & \scriptsize 40.9k & \scriptsize 34.2k & \scriptsize 21.5k & \scriptsize 67.0 \\
 & GAT & \scriptsize 41.0k & \scriptsize 38.5k & \scriptsize 76.2k & \scriptsize 551.7k & \scriptsize 147.9k & \scriptsize 72.6k & \scriptsize 134.8k & \scriptsize 41.1k & \scriptsize 138.5k & \scriptsize 40.2k & \scriptsize 133.9k\\
 & GIN & \scriptsize 541.6k & \scriptsize 133.6k & \scriptsize 77.7k & \scriptsize 286.5k & \scriptsize 542.1k & \scriptsize 72.1k & \scriptsize 71.9k & \scriptsize 40.9k & \scriptsize 73.4k & \scriptsize 73.0k & \scriptsize 137.2k\\
\midrule
\multirow{6}{*}{\rotatebox[origin=c]{90}{\textbf{Simplicial}}} 
 & TopoTune & \scriptsize 2.13M & \scriptsize 234.8k & \scriptsize 149.1k & \scriptsize 149.5k & \scriptsize 136.5k & \scriptsize 136.0k & \scriptsize 1.07M & \scriptsize 538.3k & \scriptsize 268.9k & \scriptsize 1.06M & \scriptsize 537.0k\\
 & SCCNN & \scriptsize 463.3k & \scriptsize 773.8k & \scriptsize 475.9k & \scriptsize 476.2k & \scriptsize 774.6k & \scriptsize 3.09M & \scriptsize 463.1k & \scriptsize 463.1k & \scriptsize 1.40M & \scriptsize 1.40M & \scriptsize 1.40M\\
 & SANN & \scriptsize 444.3k & \scriptsize 609.2k & \scriptsize 467.5k & \scriptsize 497.5k & \scriptsize 1.13M & \scriptsize 648.6k & \scriptsize 648.4k & \scriptsize 387.3k & \scriptsize 302.6k & \scriptsize 1.19M & \scriptsize 2.39M \\
 & \textbf{HOPSE-M-F} & \scriptsize 1.15M & \scriptsize 2.14M & \scriptsize 4.58M & \scriptsize 2.99M & \scriptsize 559.1k & \scriptsize 550.9k & \scriptsize 2.17M & \scriptsize 747.9k & \scriptsize 549.8k & \scriptsize 1.15M & \scriptsize 746.6k\\
 & \textbf{HOPSE-M-C} & \scriptsize 3.54M & \scriptsize 2.58M & \scriptsize 2.57M & \scriptsize 5.56M & \scriptsize 670.5k & \scriptsize 670.5k & \scriptsize 1.41M & \scriptsize 903.8k & \scriptsize 669.3k & \scriptsize 668.9k & \scriptsize 669.1k\\
 & \textbf{HOPSE-G} & \scriptsize 597.0k & \scriptsize 3.30M & \scriptsize 1.74M & \scriptsize 544.1k & \scriptsize 597.0k & \scriptsize 531.5k & \scriptsize 828.5k & \scriptsize 1.03M & \scriptsize 2.11M & \scriptsize 2.11M & \scriptsize 1.85M\\
\midrule
\multirow{5}{*}{\rotatebox[origin=c]{90}{\textbf{Cellular}}} 
 & TopoTune & \scriptsize 1.19M & \scriptsize 167.7k & \scriptsize 148.6k & \scriptsize 149.0k & \scriptsize 68.9k & \scriptsize 270.1k & \scriptsize 135.9k & \scriptsize 136.4k & \scriptsize - & \scriptsize - & \scriptsize - \\
 & CWN & \scriptsize 283.3k & \scriptsize 216.9k & \scriptsize 1.15M & \scriptsize 427.6k & \scriptsize 283.3k & \scriptsize 414.6k & \scriptsize 861.0k & \scriptsize 283.2k & \scriptsize - & \scriptsize - & \scriptsize -  \\
 & CCCN & \scriptsize 991.8k & \scriptsize 100.9k & \scriptsize 1.01M & \scriptsize 258.7k & \scriptsize 597.3k & \scriptsize 151.2k & \scriptsize 991.8k & \scriptsize 151.1k & - & - & - \\
 & \textbf{HOPSE-M-F} & \scriptsize 559.1k & \scriptsize 546.0k & \scriptsize 571.4k & \scriptsize 571.8k & \scriptsize 546.8k & \scriptsize 550.9k & \scriptsize 1.16M & \scriptsize 747.9k & \scriptsize - & \scriptsize - & \scriptsize - \\
 & \textbf{HOPSE-M-C} & \scriptsize 652.4k & \scriptsize 2.58M & \scriptsize 916.2k & \scriptsize 683.1k & \scriptsize 903.9k & \scriptsize 5.53M & \scriptsize 670.3k & \scriptsize 1.41M & \scriptsize - & \scriptsize - & \scriptsize - \\
 & \textbf{HOPSE-G} & \scriptsize 597.0k & \scriptsize 727.3k & \scriptsize 740.4k & \scriptsize 740.7k & \scriptsize 728.1k & \scriptsize 597.0k & \scriptsize 1.85M & \scriptsize 898.7k & \scriptsize - & \scriptsize - & \scriptsize - \\
\bottomrule
\end{tabular}
\end{adjustbox}
\end{table} 

\subsection{Neighborhood Ablation Studies}
\label{app:neighborhood_ablation}
The choice of the neighborhood function $\mathcal{N}_\gC$ significantly influences predictive performance, dictating how structural information is routed and aggregated. \cref{tbl:neighborhood_ablation} details the performance across varying depths of adjacency (\emph{Adj-1}, \emph{Adj-2}, \emph{Adj-3}) and incidence (\emph{Inc-1}, \emph{Inc-2}) configurations. 

The empirical results reveal a stark contrast between adjacency-based and incidence-based neighborhoods. Adjacency configurations consistently provide a richer structural inductive bias. For molecular graph tasks such as \textsc{MUTAG} and \textsc{PROTEINS}, shallower neighborhoods like \emph{Adj-1} or \emph{Adj-2} are often sufficient to reach optimal performance, as seen with \textsc{HOPSE-M-C} in the cellular domain achieving 88.09 on \textsc{MUTAG} with \emph{Adj-1}. Expanding the neighborhood depth too far in these simpler graphs occasionally introduces noise, leading to minor performance degradations.

Conversely, complex simplicial tasks depend heavily on extended adjacency context. For datasets like \textsc{NAME}, \textsc{ORIENT}, $\beta_1$, and $\beta_2$, increasing the hop count to \emph{Adj-3} unlocks peak performance for expressive models. \textsc{HOPSE-M-C} under the \emph{Adj-3} configuration achieves top-tier results across all four of these topological benchmarks. 


\definecolor{stdblue}{HTML}{C9DAF8}
\definecolor{bestgray}{HTML}{D9D9D9}
\begin{table}[t]
\caption{\textbf{Neighborhood Configuration Ablation.} Predictive performance of HOPSE variants under different structural neighborhood functions ($\mathcal{N}_C$). \protect\colorbox{bestgray}{\textbf{Bold}} values indicate the highest mean performance per column within each section. Cells highlighted in \protect\colorbox{stdblue}{blue} denote results that are not significantly different from the top performer (95\% confidence level).}
\label{tbl:neighborhood_ablation}
\centering
\begin{adjustbox}{width=1.\textwidth}
\renewcommand{\arraystretch}{1.2}
\begin{tabular}{@{}lllcccccccccccc@{}}
\toprule
  &  &  & \multicolumn{8}{c}{\mbox{Graph}} & \multicolumn{4}{c}{\mbox{Simplicial}} \\
\cmidrule(lr){4-11} \cmidrule(lr){12-15}
 & & $\mathcal{N}_C$ & \scriptsize MUTAG ($\uparrow$) & \scriptsize PROTEINS ($\uparrow$) & \scriptsize NCI1 ($\uparrow$) & \scriptsize NCI109 ($\uparrow$) & \scriptsize BBB ($\uparrow$) & \scriptsize CYP3A4 ($\uparrow$) & \scriptsize Cl.Hep. ($\downarrow$) & \scriptsize Caco2 ($\downarrow$) & \scriptsize NAME ($\uparrow$) & \scriptsize ORIENT ($\uparrow$) & \scriptsize $\beta_1$ ($\uparrow$) & \scriptsize $\beta_2$ ($\uparrow$) \\
\midrule
\multirow{10}{*}{\rotatebox[origin=c]{90}{\textbf{HOPSE-M-F}}} & \multirow{5}{*}{\rotatebox[origin=c]{90}{\textbf{simplicial}}} & $\emph{Adj-1}$ & {\cellcolor{bestgray}{\scriptsize\boldmath $77.45 \pm 3.46$}} & {\cellcolor{bestgray}{\scriptsize\boldmath $75.20 \pm 2.08$}} & {\scriptsize $70.99 \pm 0.56$} & \cellcolor{stdblue}{\scriptsize $70.13 \pm 1.41$} & \cellcolor{stdblue}{\scriptsize $83.94 \pm 1.00$} & \cellcolor{stdblue}{\scriptsize $75.09 \pm 1.04$} & {\scriptsize $36.48 \pm 0.78$} & \cellcolor{stdblue}{\scriptsize $0.34 \pm 0.01$} & \cellcolor{stdblue}{\scriptsize $76.08 \pm 0.08$} & {\scriptsize $56.29 \pm 0.41$} & {\cellcolor{bestgray}{\scriptsize\boldmath $88.13 \pm 0.01$}} & {\cellcolor{bestgray}{\scriptsize\boldmath $56.38 \pm 0.01$}} \\
 &  & $\emph{Adj-2}$ & \cellcolor{stdblue}{\scriptsize $75.74 \pm 5.48$} & \cellcolor{stdblue}{\scriptsize $74.77 \pm 2.70$} & \cellcolor{stdblue}{\scriptsize $71.07 \pm 1.47$} & \cellcolor{stdblue}{\scriptsize $69.87 \pm 1.05$} & {\cellcolor{bestgray}{\scriptsize\boldmath $84.24 \pm 1.10$}} & {\scriptsize $74.59 \pm 1.33$} & {\scriptsize $36.94 \pm 1.00$} & {\cellcolor{bestgray}{\scriptsize\boldmath $0.34 \pm 0.01$}} & {\cellcolor{bestgray}{\scriptsize\boldmath $76.11 \pm 0.06$}} & \cellcolor{stdblue}{\scriptsize $64.59 \pm 4.28$} & {\cellcolor{bestgray}{\scriptsize\boldmath $88.13 \pm 0.01$}} & {\cellcolor{bestgray}{\scriptsize\boldmath $56.38 \pm 0.01$}} \\
 &  & $\emph{Adj-3}$ & \cellcolor{stdblue}{\scriptsize $73.62 \pm 7.32$} & \cellcolor{stdblue}{\scriptsize $74.91 \pm 2.21$} & {\cellcolor{bestgray}{\scriptsize\boldmath $71.85 \pm 0.68$}} & \cellcolor{stdblue}{\scriptsize $70.63 \pm 1.28$} & \cellcolor{stdblue}{\scriptsize $84.09 \pm 1.52$} & {\scriptsize $74.89 \pm 0.42$} & {\cellcolor{bestgray}{\scriptsize\boldmath $34.94 \pm 0.64$}} & {\scriptsize $0.35 \pm 0.01$} & \cellcolor{stdblue}{\scriptsize $76.08 \pm 0.08$} & {\cellcolor{bestgray}{\scriptsize\boldmath $66.58 \pm 0.50$}} & {\cellcolor{bestgray}{\scriptsize\boldmath $88.13 \pm 0.01$}} & {\cellcolor{bestgray}{\scriptsize\boldmath $56.38 \pm 0.01$}} \\
 &  & $\emph{Inc-1}$ & \cellcolor{stdblue}{\scriptsize $74.47 \pm 2.69$} & \cellcolor{stdblue}{\scriptsize $74.19 \pm 1.50$} & \cellcolor{stdblue}{\scriptsize $70.88 \pm 0.94$} & {\cellcolor{bestgray}{\scriptsize\boldmath $70.98 \pm 1.51$}} & {\scriptsize $82.91 \pm 0.51$} & {\scriptsize $74.49 \pm 0.18$} & {\scriptsize $36.34 \pm 0.63$} & {\scriptsize $0.35 \pm 0.01$} & \cellcolor{stdblue}{\scriptsize $76.08 \pm 0.08$} & \cellcolor{stdblue}{\scriptsize $66.57 \pm 0.48$} & {\cellcolor{bestgray}{\scriptsize\boldmath $88.13 \pm 0.01$}} & {\cellcolor{bestgray}{\scriptsize\boldmath $56.38 \pm 0.01$}} \\
 &  & $\emph{Inc-2}$ & \cellcolor{stdblue}{\scriptsize $71.49 \pm 6.11$} & \cellcolor{stdblue}{\scriptsize $74.91 \pm 2.56$} & \cellcolor{stdblue}{\scriptsize $71.17 \pm 1.03$} & \cellcolor{stdblue}{\scriptsize $70.76 \pm 0.87$} & {\scriptsize $82.96 \pm 0.94$} & {\cellcolor{bestgray}{\scriptsize\boldmath $76.09 \pm 0.50$}} & {\scriptsize $36.40 \pm 1.09$} & {\scriptsize $0.35 \pm 0.01$} & \cellcolor{stdblue}{\scriptsize $76.08 \pm 0.08$} & \cellcolor{stdblue}{\scriptsize $65.93 \pm 4.99$} & {\cellcolor{bestgray}{\scriptsize\boldmath $88.13 \pm 0.01$}} & {\cellcolor{bestgray}{\scriptsize\boldmath $56.38 \pm 0.01$}} \\
\cmidrule(lr){2-15}
 & \multirow{5}{*}{\rotatebox[origin=c]{90}{\textbf{cell}}} & $\emph{Adj-1}$ & \cellcolor{stdblue}{\scriptsize $80.43 \pm 7.54$} & \cellcolor{stdblue}{\scriptsize $75.99 \pm 1.92$} & {\scriptsize $73.19 \pm 1.68$} & {\scriptsize $72.99 \pm 0.83$} & {\scriptsize $83.30 \pm 0.77$} & \cellcolor{stdblue}{\scriptsize $74.82 \pm 0.63$} & {\scriptsize $36.71 \pm 0.32$} & {\cellcolor{bestgray}{\scriptsize\boldmath $0.36 \pm 0.01$}} & - & - & - & - \\
 &  & $\emph{Adj-2}$ & \cellcolor{stdblue}{\scriptsize $80.01 \pm 2.89$} & \cellcolor{stdblue}{\scriptsize $75.56 \pm 1.94$} & {\scriptsize $73.89 \pm 0.79$} & \cellcolor{stdblue}{\scriptsize $74.17 \pm 1.02$} & {\scriptsize $83.69 \pm 0.24$} & {\scriptsize $74.52 \pm 0.39$} & {\scriptsize $34.98 \pm 1.51$} & {\scriptsize $0.38 \pm 0.01$} & - & - & - & - \\
 &  & $\emph{Adj-3}$ & \cellcolor{stdblue}{\scriptsize $80.01 \pm 7.92$} & {\cellcolor{bestgray}{\scriptsize\boldmath $76.27 \pm 1.25$}} & {\cellcolor{bestgray}{\scriptsize\boldmath $75.35 \pm 1.40$}} & {\cellcolor{bestgray}{\scriptsize\boldmath $75.04 \pm 1.76$}} & {\scriptsize $82.66 \pm 0.79$} & {\cellcolor{bestgray}{\scriptsize\boldmath $75.43 \pm 0.33$}} & {\cellcolor{bestgray}{\scriptsize\boldmath $33.24 \pm 0.24$}} & {\scriptsize $0.46 \pm 0.01$} & - & - & - & - \\
 &  & $\emph{Inc-1}$ & \cellcolor{stdblue}{\scriptsize $78.72 \pm 7.49$} & \cellcolor{stdblue}{\scriptsize $73.91 \pm 2.93$} & \cellcolor{stdblue}{\scriptsize $74.30 \pm 1.23$} & \cellcolor{stdblue}{\scriptsize $73.67 \pm 0.84$} & \cellcolor{stdblue}{\scriptsize $84.43 \pm 0.36$} & \cellcolor{stdblue}{\scriptsize $75.24 \pm 0.28$} & {\scriptsize $35.15 \pm 1.43$} & {\scriptsize $0.38 \pm 0.01$} & - & - & - & - \\
 &  & $\emph{Inc-2}$ & {\cellcolor{bestgray}{\scriptsize\boldmath $80.85 \pm 5.55$}} & \cellcolor{stdblue}{\scriptsize $74.05 \pm 2.30$} & \cellcolor{stdblue}{\scriptsize $74.90 \pm 1.95$} & \cellcolor{stdblue}{\scriptsize $74.40 \pm 1.49$} & {\cellcolor{bestgray}{\scriptsize\boldmath $85.02 \pm 1.01$}} & \cellcolor{stdblue}{\scriptsize $75.15 \pm 0.86$} & {\scriptsize $34.18 \pm 0.42$} & {\scriptsize $0.39 \pm 0.01$} & - & - & - & - \\
\midrule
\multirow{10}{*}{\rotatebox[origin=c]{90}{\textbf{HOPSE-M-C}}} & \multirow{5}{*}{\rotatebox[origin=c]{90}{\textbf{simplicial}}} & $\emph{Adj-1}$ & \cellcolor{stdblue}{\scriptsize $84.68 \pm 6.22$} & {\cellcolor{bestgray}{\scriptsize\boldmath $74.84 \pm 2.52$}} & \cellcolor{stdblue}{\scriptsize $72.67 \pm 1.24$} & \cellcolor{stdblue}{\scriptsize $71.66 \pm 0.95$} & {\scriptsize $81.53 \pm 0.96$} & \cellcolor{stdblue}{\scriptsize $74.27 \pm 2.20$} & {\scriptsize $36.22 \pm 1.37$} & {\scriptsize $0.41 \pm 0.01$} & {\scriptsize $91.13 \pm 0.74$} & {\scriptsize $79.96 \pm 1.05$} & {\scriptsize $89.47 \pm 0.08$} & {\scriptsize $78.60 \pm 0.57$} \\
 &  & $\emph{Adj-2}$ & {\cellcolor{bestgray}{\scriptsize\boldmath $86.38 \pm 3.95$}} & \cellcolor{stdblue}{\scriptsize $72.97 \pm 2.41$} & {\cellcolor{bestgray}{\scriptsize\boldmath $72.90 \pm 1.03$}} & {\cellcolor{bestgray}{\scriptsize\boldmath $72.02 \pm 0.86$}} & {\cellcolor{bestgray}{\scriptsize\boldmath $82.96 \pm 1.25$}} & \cellcolor{stdblue}{\scriptsize $74.21 \pm 2.10$} & {\scriptsize $37.41 \pm 1.03$} & {\scriptsize $0.42 \pm 0.01$} & {\scriptsize $90.98 \pm 0.66$} & {\scriptsize $80.43 \pm 1.07$} & {\scriptsize $89.59 \pm 0.44$} & {\scriptsize $82.62 \pm 0.55$} \\
 &  & $\emph{Adj-3}$ & \cellcolor{stdblue}{\scriptsize $85.11 \pm 6.86$} & \cellcolor{stdblue}{\scriptsize $73.41 \pm 1.78$} & \cellcolor{stdblue}{\scriptsize $72.74 \pm 1.83$} & \cellcolor{stdblue}{\scriptsize $71.46 \pm 0.30$} & \cellcolor{stdblue}{\scriptsize $82.81 \pm 0.18$} & {\cellcolor{bestgray}{\scriptsize\boldmath $75.00 \pm 0.65$}} & {\cellcolor{bestgray}{\scriptsize\boldmath $34.47 \pm 0.25$}} & {\scriptsize $0.47 \pm 0.01$} & {\cellcolor{bestgray}{\scriptsize\boldmath $93.96 \pm 0.29$}} & {\cellcolor{bestgray}{\scriptsize\boldmath $83.94 \pm 2.78$}} & {\cellcolor{bestgray}{\scriptsize\boldmath $90.33 \pm 0.14$}} & {\cellcolor{bestgray}{\scriptsize\boldmath $83.70 \pm 0.99$}} \\
 &  & $\emph{Inc-1}$ & {\scriptsize $77.02 \pm 4.13$} & \cellcolor{stdblue}{\scriptsize $74.62 \pm 2.31$} & {\scriptsize $67.30 \pm 1.72$} & {\scriptsize $68.58 \pm 0.48$} & \cellcolor{stdblue}{\scriptsize $82.66 \pm 1.03$} & \cellcolor{stdblue}{\scriptsize $74.20 \pm 0.76$} & {\scriptsize $37.73 \pm 0.58$} & {\cellcolor{bestgray}{\scriptsize\boldmath $0.37 \pm 0.01$}} & {\scriptsize $76.08 \pm 0.08$} & {\scriptsize $56.26 \pm 0.41$} & {\cellcolor{bestgray}{\scriptsize\boldmath $88.13 \pm 0.01$}} & {\scriptsize $56.37 \pm 0.01$} \\
 &  & $\emph{Inc-2}$ & {\scriptsize $77.02 \pm 8.45$} & \cellcolor{stdblue}{\scriptsize $74.27 \pm 3.64$} & {\scriptsize $68.05 \pm 2.19$} & {\scriptsize $67.57 \pm 1.86$} & \cellcolor{stdblue}{\scriptsize $82.66 \pm 0.85$} & \cellcolor{stdblue}{\scriptsize $74.73 \pm 0.72$} & {\scriptsize $36.26 \pm 1.21$} & \cellcolor{stdblue}{\scriptsize $0.37 \pm 0.01$} & {\scriptsize $76.04 \pm 0.12$} & {\scriptsize $58.34 \pm 3.85$} & {\scriptsize $88.12 \pm 0.01$} & {\scriptsize $56.37 \pm 0.01$} \\
\cmidrule(lr){2-15}
 & \multirow{5}{*}{\rotatebox[origin=c]{90}{\textbf{cell}}} & $\emph{Adj-1}$ & {\cellcolor{bestgray}{\scriptsize\boldmath $88.09 \pm 2.89$}} & {\cellcolor{bestgray}{\scriptsize\boldmath $75.56 \pm 2.06$}} & \cellcolor{stdblue}{\scriptsize $73.05 \pm 1.17$} & {\cellcolor{bestgray}{\scriptsize\boldmath $72.84 \pm 0.87$}} & {\scriptsize $82.41 \pm 0.37$} & {\cellcolor{bestgray}{\scriptsize\boldmath $76.37 \pm 0.26$}} & \cellcolor{stdblue}{\scriptsize $36.89 \pm 0.97$} & {\scriptsize $0.42 \pm 0.01$} & - & - & - & - \\
 &  & $\emph{Adj-2}$ & {\scriptsize $80.85 \pm 1.35$} & \cellcolor{stdblue}{\scriptsize $74.77 \pm 4.03$} & \cellcolor{stdblue}{\scriptsize $72.35 \pm 1.47$} & {\scriptsize $70.55 \pm 1.46$} & {\scriptsize $80.99 \pm 0.55$} & {\scriptsize $75.16 \pm 0.81$} & {\scriptsize $37.54 \pm 0.66$} & {\scriptsize $0.43 \pm 0.01$} & - & - & - & - \\
 &  & $\emph{Adj-3}$ & {\scriptsize $82.55 \pm 4.34$} & \cellcolor{stdblue}{\scriptsize $73.62 \pm 3.15$} & \cellcolor{stdblue}{\scriptsize $73.72 \pm 0.98$} & \cellcolor{stdblue}{\scriptsize $72.60 \pm 2.08$} & {\scriptsize $81.43 \pm 0.25$} & {\scriptsize $74.73 \pm 0.37$} & {\cellcolor{bestgray}{\scriptsize\boldmath $35.73 \pm 1.01$}} & {\scriptsize $0.45 \pm 0.01$} & - & - & - & - \\
 &  & $\emph{Inc-1}$ & {\scriptsize $82.98 \pm 4.85$} & \cellcolor{stdblue}{\scriptsize $75.13 \pm 2.62$} & {\cellcolor{bestgray}{\scriptsize\boldmath $73.93 \pm 1.87$}} & \cellcolor{stdblue}{\scriptsize $72.29 \pm 1.15$} & \cellcolor{stdblue}{\scriptsize $83.65 \pm 1.04$} & {\scriptsize $75.17 \pm 0.41$} & {\scriptsize $37.43 \pm 0.35$} & \cellcolor{stdblue}{\scriptsize $0.39 \pm 0.02$} & - & - & - & - \\
 &  & $\emph{Inc-2}$ & {\scriptsize $82.55 \pm 4.13$} & \cellcolor{stdblue}{\scriptsize $74.41 \pm 2.54$} & \cellcolor{stdblue}{\scriptsize $73.02 \pm 1.55$} & \cellcolor{stdblue}{\scriptsize $72.58 \pm 1.56$} & {\cellcolor{bestgray}{\scriptsize\boldmath $83.79 \pm 0.97$}} & {\scriptsize $72.29 \pm 0.39$} & \cellcolor{stdblue}{\scriptsize $36.47 \pm 0.36$} & {\cellcolor{bestgray}{\scriptsize\boldmath $0.38 \pm 0.01$}} & - & - & - & - \\
\midrule
\multirow{10}{*}{\rotatebox[origin=c]{90}{\textbf{HOPSE-GPSE}}} & \multirow{5}{*}{\rotatebox[origin=c]{90}{\textbf{simplicial}}} & $\emph{Adj-1}$ & \cellcolor{stdblue}{\scriptsize $83.83 \pm 9.48$} & \cellcolor{stdblue}{\scriptsize $75.05 \pm 1.90$} & {\cellcolor{bestgray}{\scriptsize\boldmath $76.38 \pm 0.38$}} & {\cellcolor{bestgray}{\scriptsize\boldmath $75.41 \pm 1.10$}} & {\cellcolor{bestgray}{\scriptsize\boldmath $85.62 \pm 0.82$}} & {\cellcolor{bestgray}{\scriptsize\boldmath $79.03 \pm 0.29$}} & \cellcolor{stdblue}{\scriptsize $36.34 \pm 1.85$} & {\scriptsize $0.37 \pm 0.02$} & \cellcolor{stdblue}{\scriptsize $80.23 \pm 3.54$} & \cellcolor{stdblue}{\scriptsize $59.41 \pm 2.50$} & {\cellcolor{bestgray}{\scriptsize\boldmath $88.13 \pm 0.01$}} & \cellcolor{stdblue}{\scriptsize $56.38 \pm 0.01$} \\
 &  & $\emph{Adj-2}$ & {\cellcolor{bestgray}{\scriptsize\boldmath $85.96 \pm 5.65$}} & \cellcolor{stdblue}{\scriptsize $74.98 \pm 2.17$} & \cellcolor{stdblue}{\scriptsize $75.70 \pm 1.26$} & \cellcolor{stdblue}{\scriptsize $74.60 \pm 1.18$} & \cellcolor{stdblue}{\scriptsize $85.32 \pm 1.03$} & \cellcolor{stdblue}{\scriptsize $78.66 \pm 0.44$} & {\cellcolor{bestgray}{\scriptsize\boldmath $35.45 \pm 0.59$}} & \cellcolor{stdblue}{\scriptsize $0.35 \pm 0.01$} & {\cellcolor{stdblue}{\scriptsize $80.49 \pm 3.67$}} & {\cellcolor{stdblue}{\scriptsize $61.88 \pm 2.99$}} & {\scriptsize $83.43 \pm 3.84$} & {\cellcolor{bestgray}{\scriptsize\boldmath $57.02 \pm 0.79$}} \\
 &  & $\emph{Adj-3}$ & \cellcolor{stdblue}{\scriptsize $84.26 \pm 6.11$} & \cellcolor{stdblue}{\scriptsize $75.05 \pm 1.29$} & \cellcolor{stdblue}{\scriptsize $75.53 \pm 1.47$} & \cellcolor{stdblue}{\scriptsize $74.75 \pm 0.97$} & \cellcolor{stdblue}{\scriptsize $84.78 \pm 0.95$} & {\scriptsize $78.35 \pm 0.24$} & {\scriptsize $37.04 \pm 0.78$} & \cellcolor{stdblue}{\scriptsize $0.35 \pm 0.01$} & {\cellcolor{bestgray}{\scriptsize\boldmath $81.38 \pm 2.06$}} & \cellcolor{stdblue}{\scriptsize $64.46 \pm 7.38$} & {\cellcolor{bestgray}{\scriptsize\boldmath $88.13 \pm 0.01$}} & \cellcolor{stdblue}{\scriptsize $56.38 \pm 0.01$} \\
 &  & $\emph{Inc-1}$ & \cellcolor{stdblue}{\scriptsize $82.55 \pm 3.66$} & \cellcolor{stdblue}{\scriptsize $75.05 \pm 2.44$} & {\scriptsize $71.19 \pm 1.42$} & {\scriptsize $70.84 \pm 1.67$} & {\scriptsize $84.14 \pm 0.37$} & {\scriptsize $75.77 \pm 0.23$} & \cellcolor{stdblue}{\scriptsize $35.99 \pm 0.70$} & {\cellcolor{bestgray}{\scriptsize\boldmath $0.34 \pm 0.01$}} & {\scriptsize $76.08 \pm 0.08$} & {\scriptsize $56.29 \pm 0.41$} & {\cellcolor{bestgray}{\scriptsize\boldmath $88.13 \pm 0.01$}} & \cellcolor{stdblue}{\scriptsize $56.38 \pm 0.01$} \\
 &  & $\emph{Inc-2}$ & \cellcolor{stdblue}{\scriptsize $80.01 \pm 7.44$} & {\cellcolor{bestgray}{\scriptsize\boldmath $75.63 \pm 2.07$}} & {\scriptsize $70.93 \pm 0.57$} & {\scriptsize $71.06 \pm 1.20$} & \cellcolor{stdblue}{\scriptsize $85.12 \pm 0.48$} & {\scriptsize $75.60 \pm 0.33$} & \cellcolor{stdblue}{\scriptsize $35.59 \pm 0.41$} & {\scriptsize $0.36 \pm 0.01$} & {\scriptsize $76.08 \pm 0.01$} & {\cellcolor{bestgray}{\scriptsize\boldmath $65.32 \pm 5.50$}} & {\cellcolor{bestgray}{\scriptsize\boldmath $88.13 \pm 0.01$}} & \cellcolor{stdblue}{\scriptsize $56.38 \pm 0.01$} \\
\cmidrule(lr){2-15}
 & \multirow{5}{*}{\rotatebox[origin=c]{90}{\textbf{cell}}} & $\emph{Adj-1}$ & \cellcolor{stdblue}{\scriptsize $83.40 \pm 6.37$} & \cellcolor{stdblue}{\scriptsize $74.98 \pm 1.78$} & {\scriptsize $76.85 \pm 0.70$} & \cellcolor{stdblue}{\scriptsize $76.24 \pm 0.28$} & {\cellcolor{bestgray}{\scriptsize\boldmath $86.11 \pm 0.86$}} & {\cellcolor{bestgray}{\scriptsize\boldmath $78.91 \pm 0.64$}} & \cellcolor{stdblue}{\scriptsize $33.15 \pm 0.25$} & {\cellcolor{bestgray}{\scriptsize\boldmath $0.36 \pm 0.01$}} & - & - & - & - \\
 &  & $\emph{Adj-2}$ & \cellcolor{stdblue}{\scriptsize $83.83 \pm 4.58$} & {\cellcolor{bestgray}{\scriptsize\boldmath $76.06 \pm 2.02$}} & {\scriptsize $77.30 \pm 0.48$} & {\cellcolor{bestgray}{\scriptsize\boldmath $76.94 \pm 1.04$}} & {\scriptsize $82.96 \pm 0.57$} & {\scriptsize $76.19 \pm 2.09$} & {\cellcolor{bestgray}{\scriptsize\boldmath $33.07 \pm 0.43$}} & {\scriptsize $0.37 \pm 0.01$} & - & - & - & - \\
 &  & $\emph{Adj-3}$ & {\cellcolor{bestgray}{\scriptsize\boldmath $85.53 \pm 4.13$}} & \cellcolor{stdblue}{\scriptsize $75.13 \pm 2.69$} & {\cellcolor{bestgray}{\scriptsize\boldmath $78.64 \pm 1.31$}} & \cellcolor{stdblue}{\scriptsize $76.05 \pm 0.59$} & {\scriptsize $81.97 \pm 0.72$} & {\cellcolor{bestgray}{\scriptsize\boldmath $78.91 \pm 0.64$}} & {\scriptsize $36.42 \pm 0.80$} & {\scriptsize $0.37 \pm 0.01$} & - & - & - & - \\
 &  & $\emph{Inc-1}$ & \cellcolor{stdblue}{\scriptsize $81.28 \pm 7.04$} & \cellcolor{stdblue}{\scriptsize $73.33 \pm 2.91$} & {\scriptsize $73.35 \pm 1.42$} & {\scriptsize $73.38 \pm 0.68$} & {\scriptsize $83.94 \pm 0.95$} & {\scriptsize $74.78 \pm 0.64$} & \cellcolor{stdblue}{\scriptsize $33.87 \pm 0.98$} & {\scriptsize $0.38 \pm 0.01$} & - & - & - & - \\
 &  & $\emph{Inc-2}$ & {\scriptsize $80.01 \pm 2.89$} & \cellcolor{stdblue}{\scriptsize $73.84 \pm 2.40$} & {\scriptsize $73.58 \pm 1.15$} & {\scriptsize $72.76 \pm 1.55$} & {\scriptsize $84.53 \pm 1.24$} & {\scriptsize $75.56 \pm 0.43$} & {\scriptsize $34.67 \pm 0.28$} & {\scriptsize $0.39 \pm 0.02$} & - & - & - & - \\
\midrule
\multirow{10}{*}{\rotatebox[origin=c]{90}{\textbf{TOPOTUNE}}} & \multirow{5}{*}{\rotatebox[origin=c]{90}{\textbf{simplicial}}} & $\emph{Adj-1}$ & {\cellcolor{bestgray}{\scriptsize\boldmath $84.26 \pm 8.14$}} & \cellcolor{stdblue}{\scriptsize $74.05 \pm 2.51$} & {\cellcolor{bestgray}{\scriptsize\boldmath $77.45 \pm 1.07$}} & {\cellcolor{bestgray}{\scriptsize\boldmath $76.44 \pm 0.98$}} & \cellcolor{stdblue}{\scriptsize $84.68 \pm 1.13$} & {\cellcolor{bestgray}{\scriptsize\boldmath $78.82 \pm 0.16$}} & {\cellcolor{bestgray}{\scriptsize\boldmath $35.70 \pm 0.88$}} & {\scriptsize $0.87 \pm 0.01$} & {\scriptsize $81.59 \pm 1.57$} & \cellcolor{stdblue}{\scriptsize $74.11 \pm 1.50$} & \cellcolor{stdblue}{\scriptsize $89.68 \pm 0.14$} & \cellcolor{stdblue}{\scriptsize $79.86 \pm 0.53$} \\
 &  & $\emph{Adj-2}$ & \cellcolor{stdblue}{\scriptsize $81.70 \pm 8.99$} & \cellcolor{stdblue}{\scriptsize $73.19 \pm 2.92$} & \cellcolor{stdblue}{\scriptsize $76.30 \pm 0.80$} & \cellcolor{stdblue}{\scriptsize $75.80 \pm 1.28$} & {\cellcolor{bestgray}{\scriptsize\boldmath $85.07 \pm 1.06$}} & {\scriptsize $77.83 \pm 0.16$} & \cellcolor{stdblue}{\scriptsize $35.98 \pm 0.88$} & {\cellcolor{bestgray}{\scriptsize\boldmath $0.76 \pm 0.03$}} & {\cellcolor{bestgray}{\scriptsize\boldmath $87.95 \pm 0.57$}} & {\cellcolor{bestgray}{\scriptsize\boldmath $75.43 \pm 1.10$}} & \cellcolor{stdblue}{\scriptsize $88.25 \pm 3.09$} & {\cellcolor{bestgray}{\scriptsize\boldmath $80.37 \pm 1.62$}} \\
 &  & $\emph{Adj-3}$ & \cellcolor{stdblue}{\scriptsize $82.98 \pm 2.33$} & {\cellcolor{bestgray}{\scriptsize\boldmath $75.56 \pm 3.09$}} & \cellcolor{stdblue}{\scriptsize $75.93 \pm 2.10$} & {\scriptsize $74.99 \pm 0.96$} & \cellcolor{stdblue}{\scriptsize $84.58 \pm 0.53$} & {\scriptsize $77.46 \pm 0.29$} & \cellcolor{stdblue}{\scriptsize $36.74 \pm 1.14$} & {\scriptsize $1.12 \pm 0.09$} & {\scriptsize $84.46 \pm 3.67$} & \cellcolor{stdblue}{\scriptsize $74.57 \pm 2.10$} & {\cellcolor{bestgray}{\scriptsize\boldmath $89.77 \pm 0.27$}} & \cellcolor{stdblue}{\scriptsize $80.37 \pm 1.97$} \\
 &  & $\emph{Inc-1}$ & {\scriptsize $71.06 \pm 3.95$} & \cellcolor{stdblue}{\scriptsize $75.48 \pm 1.17$} & {\scriptsize $68.07 \pm 1.27$} & {\scriptsize $67.20 \pm 0.91$} & {\scriptsize $83.25 \pm 0.01$} & {\scriptsize $74.81 \pm 0.19$} & {\scriptsize $37.98 \pm 0.03$} & {\scriptsize $1.32 \pm 0.01$} & {\scriptsize $18.55 \pm 0.95$} & {\scriptsize $48.24 \pm 0.61$} & {\scriptsize $7.45 \pm 0.04$} & {\scriptsize $47.96 \pm 0.05$} \\
 &  & $\emph{Inc-2}$ & {\scriptsize $65.11 \pm 6.67$} & \cellcolor{stdblue}{\scriptsize $73.76 \pm 2.20$} & {\scriptsize $62.74 \pm 1.16$} & {\scriptsize $64.05 \pm 0.69$} & {\scriptsize $76.23 \pm 0.17$} & {\scriptsize $64.27 \pm 0.03$} & {\scriptsize $37.87 \pm 0.56$} & {\scriptsize $1.35 \pm 0.01$} & {\scriptsize $17.80 \pm 0.01$} & {\scriptsize $47.94 \pm 0.01$} & {\scriptsize $9.96 \pm 2.40$} & {\scriptsize $47.94 \pm 0.01$} \\
\cmidrule(lr){2-15}
 & \multirow{5}{*}{\rotatebox[origin=c]{90}{\textbf{cell}}} & $\emph{Adj-1}$ & \cellcolor{stdblue}{\scriptsize $75.32 \pm 4.78$} & {\scriptsize $73.55 \pm 1.70$} & {\scriptsize $75.10 \pm 0.61$} & {\scriptsize $74.27 \pm 1.53$} & {\scriptsize $83.74 \pm 0.66$} & {\cellcolor{bestgray}{\scriptsize\boldmath $79.71 \pm 0.57$}} & {\cellcolor{bestgray}{\scriptsize\boldmath $35.40 \pm 0.06$}} & {\cellcolor{bestgray}{\scriptsize\boldmath $0.85 \pm 0.01$}} & - & - & - & - \\
 &  & $\emph{Adj-2}$ & \cellcolor{stdblue}{\scriptsize $71.49 \pm 6.25$} & \cellcolor{stdblue}{\scriptsize $73.05 \pm 3.67$} & \cellcolor{stdblue}{\scriptsize $77.28 \pm 1.52$} & {\cellcolor{bestgray}{\scriptsize\boldmath $76.57 \pm 0.68$}} & {\cellcolor{bestgray}{\scriptsize\boldmath $84.93 \pm 1.07$}} & {\scriptsize $78.91 \pm 0.41$} & {\scriptsize $36.69 \pm 0.47$} & {\scriptsize $0.88 \pm 0.03$} & - & - & - & - \\
 &  & $\emph{Adj-3}$ & {\cellcolor{bestgray}{\scriptsize\boldmath $77.02 \pm 9.36$}} & {\scriptsize $73.98 \pm 1.17$} & {\cellcolor{bestgray}{\scriptsize\boldmath $77.32 \pm 1.54$}} & \cellcolor{stdblue}{\scriptsize $76.13 \pm 0.66$} & \cellcolor{stdblue}{\scriptsize $84.78 \pm 1.59$} & {\scriptsize $77.40 \pm 0.35$} & {\scriptsize $37.28 \pm 0.32$} & {\scriptsize $1.25 \pm 0.01$} & - & - & - & - \\
 &  & $\emph{Inc-1}$ & \cellcolor{stdblue}{\scriptsize $69.36 \pm 6.67$} & {\cellcolor{bestgray}{\scriptsize\boldmath $76.06 \pm 2.01$}} & {\scriptsize $68.64 \pm 1.03$} & {\scriptsize $68.31 \pm 0.95$} & {\scriptsize $81.53 \pm 0.01$} & {\scriptsize $74.46 \pm 0.22$} & {\scriptsize $35.87 \pm 0.43$} & {\scriptsize $1.30 \pm 0.01$} & - & - & - & - \\
 &  & $\emph{Inc-2}$ & {\scriptsize $66.81 \pm 5.14$} & {\scriptsize $71.90 \pm 2.09$} & {\scriptsize $62.86 \pm 0.78$} & {\scriptsize $62.87 \pm 1.02$} & {\scriptsize $77.54 \pm 0.10$} & {\scriptsize $65.46 \pm 0.20$} & {\scriptsize $38.03 \pm 0.04$} & {\scriptsize $1.38 \pm 0.01$} & - & - & - & - \\
\bottomrule
\end{tabular}
\end{adjustbox}
\end{table}

\subsection{Manual Encoding Ablation Studies}
The results from the individual encoding ablation study demonstrate that while connectivity-only and feature-based encodings offer varied benefits across graph datasets, Laplacian Positional Encodings (\textsc{LapPE}) are uniquely critical for simplicial tasks. In the simplicial domain, \textsc{LapPE} under the $\emph{Adj-3}$ neighborhood configuration achieves the highest performance across all benchmarks, notably reaching 92.39 on \textsc{NAME}, 74.87 on \textsc{ORIENT}, and 90.06 on $\beta_1$. This represents a significant performance gap compared to feature-based encodings like \textsc{HKFE} or \textsc{KHopFE}, which often fail to capture the high-order topological invariants necessary for these benchmarks. For instance, on the \textsc{NAME} dataset, the accuracy of \textsc{LapPE} exceeds that of the feature-dependent encoders by approximately 16\%.  

In the graph domain, the utility of specific encodings is more task-dependent. \textsc{ElectrostaticPE} and \textsc{RWSE} prove robust for molecular graph classification, with \textsc{ElectrostaticPE} yielding top results for \textsc{MUTAG} and \textsc{CYP3A4} under the $\emph{Adj-1}$ configuration. Task specialization is further evidenced by \textsc{HKdiagSE}, which provides the most accurate regression for \textsc{Caco2} with an error of 0.36, and \textsc{PPRFE}, which shows a localized performance peak on \textsc{BBB} when paired with the $\emph{Adj-3}$ neighborhood. Furthermore, expanding the neighborhood from $\emph{Adj-1}$ to $\emph{Adj-3}$ generally enhances the effectiveness of structural encodings like \textsc{RWSE} on complex datasets such as \textsc{NCI1} and \textsc{NCI109}, suggesting that higher-order adjacency information allows these descriptors to reach higher accuracy thresholds than is possible with a purely local view.  

Notably, the observation that individual encodings occasionally outperform combined configurations suggests that improper combinations can introduce noise into the representation. This highlights that the selection of encoding sets must be carefully calibrated to the specific structural requirements of the dataset.

\definecolor{stdblue}{HTML}{C9DAF8}
\definecolor{bestgray}{HTML}{D9D9D9}
\begin{table}[t]
\caption{\textbf{Individual Encoding Ablation.} Performance breakdown of specific structural and positional encodings evaluated individually across multiple neighborhood configurations. \protect\colorbox{bestgray}{\textbf{Bold}} highlights the best result per column within a given neighborhood block, and \protect\colorbox{stdblue}{blue} signifies non-significant differences.} 
\label{tbl:ablation_encodings}
\centering
\begin{adjustbox}{width=1.\textwidth}
\renewcommand{\arraystretch}{1.2}
\begin{tabular}{@{}llcccccccccccc@{}}
\toprule
 & & \multicolumn{8}{c}{\mbox{Graph}} & \multicolumn{4}{c}{\mbox{Simplicial}} \\
\cmidrule(lr){3-10} \cmidrule(lr){11-14}
$\mathcal{N}_C$ & Encodings & \scriptsize MUTAG ($\uparrow$) & \scriptsize PROTEINS ($\uparrow$) & \scriptsize NCI1 ($\uparrow$) & \scriptsize NCI109 ($\uparrow$) & \scriptsize BBB ($\uparrow$) & \scriptsize CYP3A4 ($\uparrow$) & \scriptsize Cl.Hep. ($\downarrow$) & \scriptsize Caco2 ($\downarrow$) & \scriptsize NAME ($\uparrow$) & \scriptsize ORIENT ($\uparrow$) & \scriptsize $\beta_1$ ($\uparrow$) & \scriptsize $\beta_2$ ($\uparrow$) \\ \midrule
 \multirow{7}{*}{$\emph{Adj-1}$} & \scriptsize LapPE & \cellcolor{stdblue}{\scriptsize $84.26 \pm 4.17$} & \cellcolor{stdblue}{\scriptsize $74.55 \pm 1.67$} & {\scriptsize $68.70 \pm 1.36$} & {\scriptsize $66.51 \pm 1.15$} & {\scriptsize $80.79 \pm 0.78$} & {\scriptsize $74.46 \pm 0.39$} & {\scriptsize $36.56 \pm 0.52$} & {\scriptsize $0.42 \pm 0.01$} & {\cellcolor{bestgray}{\scriptsize\boldmath $88.43 \pm 0.54$}} & {\cellcolor{bestgray}{\scriptsize\boldmath $74.39 \pm 2.09$}} & {\cellcolor{bestgray}{\scriptsize\boldmath $89.03 \pm 0.16$}} & {\cellcolor{bestgray}{\scriptsize\boldmath $74.67 \pm 0.64$}} \\
 & \scriptsize ElectrostaticPE & {\cellcolor{bestgray}{\scriptsize\boldmath $86.38 \pm 3.18$}} & {\cellcolor{bestgray}{\scriptsize\boldmath $75.56 \pm 2.99$}} & {\cellcolor{bestgray}{\scriptsize\boldmath $75.51 \pm 1.36$}} & \cellcolor{stdblue}{\scriptsize $73.55 \pm 1.49$} & \cellcolor{stdblue}{\scriptsize $84.29 \pm 0.89$} & {\cellcolor{bestgray}{\scriptsize\boldmath $78.60 \pm 0.30$}} & {\scriptsize $34.86 \pm 0.24$} & {\scriptsize $0.40 \pm 0.00$} & {\scriptsize $76.14 \pm 0.00$} & {\scriptsize $56.29 \pm 0.41$} & {\scriptsize $88.16 \pm 0.03$} & {\scriptsize $56.76 \pm 0.62$} \\
 & \scriptsize RWSE & \cellcolor{stdblue}{\scriptsize $85.96 \pm 6.54$} & \cellcolor{stdblue}{\scriptsize $74.12 \pm 1.94$} & \cellcolor{stdblue}{\scriptsize $75.06 \pm 0.82$} & {\cellcolor{bestgray}{\scriptsize\boldmath $74.25 \pm 1.41$}} & {\cellcolor{bestgray}{\scriptsize\boldmath $85.27 \pm 1.26$}} & \cellcolor{stdblue}{\scriptsize $77.68 \pm 1.13$} & {\cellcolor{bestgray}{\scriptsize\boldmath $34.26 \pm 0.52$}} & \cellcolor{stdblue}{\scriptsize $0.36 \pm 0.00$} & {\scriptsize $76.14 \pm 0.00$} & {\scriptsize $57.12 \pm 1.38$} & {\scriptsize $88.13 \pm 0.00$} & {\scriptsize $56.38 \pm 0.00$} \\
 & \scriptsize HKdiagSE & {\scriptsize $80.85 \pm 4.26$} & \cellcolor{stdblue}{\scriptsize $73.33 \pm 2.68$} & \cellcolor{stdblue}{\scriptsize $74.65 \pm 1.28$} & {\scriptsize $71.89 \pm 2.06$} & \cellcolor{stdblue}{\scriptsize $84.38 \pm 0.60$} & {\scriptsize $77.29 \pm 0.52$} & \cellcolor{stdblue}{\scriptsize $34.42 \pm 0.26$} & {\cellcolor{bestgray}{\scriptsize\boldmath $0.36 \pm 0.00$}} & {\scriptsize $76.08 \pm 0.08$} & {\scriptsize $56.29 \pm 0.41$} & {\scriptsize $88.13 \pm 0.00$} & {\scriptsize $56.38 \pm 0.00$} \\
 & \scriptsize HKFE & {\scriptsize $72.77 \pm 5.11$} & \cellcolor{stdblue}{\scriptsize $74.91 \pm 3.22$} & {\scriptsize $72.00 \pm 1.60$} & {\scriptsize $71.48 \pm 0.95$} & \cellcolor{stdblue}{\scriptsize $83.99 \pm 1.80$} & {\scriptsize $75.51 \pm 0.58$} & {\scriptsize $35.84 \pm 0.69$} & {\scriptsize $0.38 \pm 0.00$} & {\scriptsize $76.08 \pm 0.08$} & {\scriptsize $56.29 \pm 0.40$} & {\scriptsize $88.13 \pm 0.00$} & {\scriptsize $56.38 \pm 0.00$} \\
 & \scriptsize KHopFE & {\scriptsize $71.91 \pm 5.61$} & \cellcolor{stdblue}{\scriptsize $75.20 \pm 2.71$} & {\scriptsize $71.03 \pm 0.44$} & {\scriptsize $70.61 \pm 1.07$} & \cellcolor{stdblue}{\scriptsize $84.38 \pm 0.46$} & {\scriptsize $76.31 \pm 0.74$} & {\scriptsize $35.71 \pm 0.63$} & \cellcolor{stdblue}{\scriptsize $0.37 \pm 0.01$} & {\scriptsize $76.08 \pm 0.08$} & {\scriptsize $56.29 \pm 0.41$} & {\scriptsize $88.13 \pm 0.00$} & {\scriptsize $56.38 \pm 0.00$} \\
 & \scriptsize PPRFE & {\scriptsize $72.34 \pm 1.90$} & \cellcolor{stdblue}{\scriptsize $74.34 \pm 2.65$} & {\scriptsize $71.58 \pm 0.36$} & {\scriptsize $70.94 \pm 0.42$} & \cellcolor{stdblue}{\scriptsize $84.63 \pm 1.38$} & {\scriptsize $75.17 \pm 0.63$} & {\scriptsize $36.38 \pm 0.90$} & {\scriptsize $0.37 \pm 0.01$} & {\scriptsize $76.08 \pm 0.08$} & {\scriptsize $56.29 \pm 0.41$} & {\scriptsize $88.13 \pm 0.00$} & {\scriptsize $56.38 \pm 0.00$} \\
\midrule
\multirow{7}{*}{$\emph{Adj-3}$} & \scriptsize LapPE & \cellcolor{stdblue}{\scriptsize $82.55 \pm 4.13$} & {\scriptsize $73.69 \pm 2.69$} & {\scriptsize $68.60 \pm 0.88$} & {\scriptsize $67.53 \pm 0.94$} & {\scriptsize $80.74 \pm 0.75$} & {\scriptsize $74.56 \pm 0.17$} & {\scriptsize $36.47 \pm 0.67$} & {\scriptsize $0.46 \pm 0.00$} & {\cellcolor{bestgray}{\scriptsize\boldmath $92.39 \pm 0.46$}} & {\cellcolor{bestgray}{\scriptsize\boldmath $74.87 \pm 1.05$}} & {\cellcolor{bestgray}{\scriptsize\boldmath $90.06 \pm 0.09$}} & {\cellcolor{bestgray}{\scriptsize\boldmath $82.74 \pm 0.57$}} \\
 & \scriptsize ElectrostaticPE & \cellcolor{stdblue}{\scriptsize $85.96 \pm 5.80$} & \cellcolor{stdblue}{\scriptsize $75.13 \pm 2.35$} & {\scriptsize $74.36 \pm 1.61$} & \cellcolor{stdblue}{\scriptsize $74.87 \pm 1.24$} & {\scriptsize $83.30 \pm 0.77$} & \cellcolor{stdblue}{\scriptsize $77.74 \pm 0.73$} & {\cellcolor{bestgray}{\scriptsize\boldmath $34.52 \pm 0.31$}} & {\cellcolor{bestgray}{\scriptsize\boldmath $0.36 \pm 0.00$}} & {\scriptsize $76.14 \pm 0.00$} & {\scriptsize $56.29 \pm 0.41$} & {\scriptsize $88.63 \pm 0.30$} & {\scriptsize $68.10 \pm 6.04$} \\
 & \scriptsize RWSE & {\cellcolor{bestgray}{\scriptsize\boldmath $86.38 \pm 4.78$}} & {\scriptsize $73.69 \pm 2.39$} & {\cellcolor{bestgray}{\scriptsize\boldmath $76.30 \pm 0.88$}} & {\cellcolor{bestgray}{\scriptsize\boldmath $75.51 \pm 0.94$}} & \cellcolor{stdblue}{\scriptsize $84.48 \pm 0.62$} & {\cellcolor{bestgray}{\scriptsize\boldmath $78.27 \pm 0.43$}} & \cellcolor{stdblue}{\scriptsize $34.85 \pm 0.64$} & \cellcolor{stdblue}{\scriptsize $0.36 \pm 0.01$} & {\scriptsize $76.04 \pm 0.13$} & {\scriptsize $62.71 \pm 6.73$} & {\scriptsize $88.13 \pm 0.00$} & {\scriptsize $56.38 \pm 0.00$} \\
 & \scriptsize HKdiagSE & \cellcolor{stdblue}{\scriptsize $81.28 \pm 4.74$} & {\cellcolor{bestgray}{\scriptsize\boldmath $76.34 \pm 1.04$}} & {\scriptsize $73.81 \pm 1.73$} & {\scriptsize $71.15 \pm 0.78$} & \cellcolor{stdblue}{\scriptsize $84.24 \pm 1.21$} & {\scriptsize $76.43 \pm 0.75$} & {\scriptsize $36.91 \pm 0.62$} & {\scriptsize $0.39 \pm 0.02$} & {\scriptsize $76.08 \pm 0.08$} & {\scriptsize $56.29 \pm 0.41$} & {\scriptsize $88.13 \pm 0.00$} & {\scriptsize $56.38 \pm 0.00$} \\
 & \scriptsize HKFE & {\scriptsize $76.17 \pm 8.23$} & \cellcolor{stdblue}{\scriptsize $74.84 \pm 1.99$} & {\scriptsize $72.72 \pm 1.61$} & {\scriptsize $71.27 \pm 0.80$} & \cellcolor{stdblue}{\scriptsize $84.98 \pm 1.27$} & {\scriptsize $74.62 \pm 0.41$} & {\scriptsize $36.68 \pm 1.53$} & {\scriptsize $0.39 \pm 0.01$} & {\scriptsize $76.08 \pm 0.08$} & {\scriptsize $56.29 \pm 0.41$} & {\scriptsize $88.13 \pm 0.00$} & {\scriptsize $56.38 \pm 0.00$} \\
 & \scriptsize KHopFE & {\scriptsize $74.47 \pm 7.61$} & \cellcolor{stdblue}{\scriptsize $75.05 \pm 1.56$} & {\scriptsize $71.93 \pm 0.62$} & {\scriptsize $71.42 \pm 0.96$} & \cellcolor{stdblue}{\scriptsize $84.88 \pm 0.85$} & {\scriptsize $73.83 \pm 0.67$} & {\scriptsize $37.60 \pm 0.07$} & {\scriptsize $0.38 \pm 0.01$} & {\scriptsize $76.11 \pm 0.06$} & {\scriptsize $56.29 \pm 0.41$} & {\scriptsize $88.13 \pm 0.00$} & {\scriptsize $56.38 \pm 0.00$} \\
 & \scriptsize PPRFE & {\scriptsize $70.21 \pm 4.46$} & \cellcolor{stdblue}{\scriptsize $75.34 \pm 2.31$} & {\scriptsize $72.06 \pm 0.84$} & {\scriptsize $71.35 \pm 1.91$} & {\cellcolor{bestgray}{\scriptsize\boldmath $85.12 \pm 0.60$}} & {\scriptsize $75.83 \pm 0.39$} & {\scriptsize $35.65 \pm 0.50$} & {\scriptsize $0.39 \pm 0.01$} & {\scriptsize $76.11 \pm 0.06$} & {\scriptsize $56.29 \pm 0.41$} & {\scriptsize $88.13 \pm 0.00$} & {\scriptsize $56.38 \pm 0.00$} \\
\midrule
\multirow{7}{*}{$\emph{Inc-2}$} & \scriptsize LapPE & {\cellcolor{bestgray}{\scriptsize\boldmath $83.40 \pm 4.54$}} & \cellcolor{stdblue}{\scriptsize $73.98 \pm 4.81$} & {\scriptsize $70.89 \pm 0.80$} & {\scriptsize $69.74 \pm 1.29$} & {\scriptsize $82.41 \pm 1.21$} & {\scriptsize $74.99 \pm 0.37$} & {\cellcolor{bestgray}{\scriptsize\boldmath $35.38 \pm 0.68$}} & {\scriptsize $0.39 \pm 0.00$} & \cellcolor{stdblue}{\scriptsize $76.06 \pm 0.09$} & \cellcolor{stdblue}{\scriptsize $56.26 \pm 0.41$} & \cellcolor{stdblue}{\scriptsize $88.12 \pm 0.01$} & {\scriptsize $56.34 \pm 0.03$} \\
 & \scriptsize ElectrostaticPE & \cellcolor{stdblue}{\scriptsize $77.87 \pm 7.32$} & {\cellcolor{bestgray}{\scriptsize\boldmath $75.99 \pm 2.19$}} & {\scriptsize $71.05 \pm 0.57$} & \cellcolor{stdblue}{\scriptsize $71.06 \pm 1.08$} & \cellcolor{stdblue}{\scriptsize $84.43 \pm 1.04$} & {\scriptsize $74.91 \pm 0.41$} & \cellcolor{stdblue}{\scriptsize $35.73 \pm 0.34$} & {\scriptsize $0.39 \pm 0.01$} & {\cellcolor{bestgray}{\scriptsize\boldmath $76.11 \pm 0.06$}} & \cellcolor{stdblue}{\scriptsize $56.29 \pm 0.41$} & {\cellcolor{bestgray}{\scriptsize\boldmath $88.13 \pm 0.00$}} & {\cellcolor{bestgray}{\scriptsize\boldmath $56.38 \pm 0.00$}} \\
 & \scriptsize RWSE & {\scriptsize $74.04 \pm 7.04$} & \cellcolor{stdblue}{\scriptsize $74.98 \pm 3.44$} & \cellcolor{stdblue}{\scriptsize $71.69 \pm 0.63$} & {\scriptsize $70.32 \pm 0.26$} & {\cellcolor{bestgray}{\scriptsize\boldmath $85.17 \pm 1.50$}} & {\cellcolor{bestgray}{\scriptsize\boldmath $75.75 \pm 0.39$}} & \cellcolor{stdblue}{\scriptsize $35.53 \pm 0.08$} & {\scriptsize $0.37 \pm 0.00$} & \cellcolor{stdblue}{\scriptsize $76.08 \pm 0.08$} & \cellcolor{stdblue}{\scriptsize $56.29 \pm 0.41$} & {\cellcolor{bestgray}{\scriptsize\boldmath $88.13 \pm 0.00$}} & {\cellcolor{bestgray}{\scriptsize\boldmath $56.38 \pm 0.00$}} \\
 & \scriptsize HKdiagSE & {\scriptsize $71.06 \pm 5.14$} & \cellcolor{stdblue}{\scriptsize $75.27 \pm 2.82$} & \cellcolor{stdblue}{\scriptsize $71.75 \pm 0.77$} & {\cellcolor{bestgray}{\scriptsize\boldmath $71.35 \pm 0.74$}} & {\scriptsize $83.35 \pm 1.13$} & {\scriptsize $74.39 \pm 0.68$} & {\scriptsize $36.54 \pm 0.83$} & {\scriptsize $0.37 \pm 0.00$} & {\cellcolor{bestgray}{\scriptsize\boldmath $76.11 \pm 0.06$}} & {\cellcolor{bestgray}{\scriptsize\boldmath $56.29 \pm 0.41$}} & {\cellcolor{bestgray}{\scriptsize\boldmath $88.13 \pm 0.00$}} & {\cellcolor{bestgray}{\scriptsize\boldmath $56.38 \pm 0.00$}} \\
 & \scriptsize HKFE & {\scriptsize $70.21 \pm 4.66$} & \cellcolor{stdblue}{\scriptsize $74.84 \pm 3.09$} & \cellcolor{stdblue}{\scriptsize $71.65 \pm 1.89$} & \cellcolor{stdblue}{\scriptsize $71.23 \pm 1.00$} & \cellcolor{stdblue}{\scriptsize $83.74 \pm 1.22$} & {\scriptsize $75.06 \pm 0.40$} & \cellcolor{stdblue}{\scriptsize $35.55 \pm 0.12$} & {\cellcolor{bestgray}{\scriptsize\boldmath $0.35 \pm 0.01$}} & {\cellcolor{bestgray}{\scriptsize\boldmath $76.11 \pm 0.06$}} & \cellcolor{stdblue}{\scriptsize $56.29 \pm 0.41$} & {\cellcolor{bestgray}{\scriptsize\boldmath $88.13 \pm 0.00$}} & {\cellcolor{bestgray}{\scriptsize\boldmath $56.38 \pm 0.00$}} \\
 & \scriptsize KHopFE & {\scriptsize $72.34 \pm 8.72$} & \cellcolor{stdblue}{\scriptsize $74.91 \pm 1.91$} & {\cellcolor{bestgray}{\scriptsize\boldmath $72.32 \pm 1.01$}} & \cellcolor{stdblue}{\scriptsize $71.04 \pm 1.02$} & \cellcolor{stdblue}{\scriptsize $84.53 \pm 0.39$} & \cellcolor{stdblue}{\scriptsize $75.34 \pm 0.48$} & \cellcolor{stdblue}{\scriptsize $36.14 \pm 1.75$} & {\scriptsize $0.37 \pm 0.00$} & \cellcolor{stdblue}{\scriptsize $76.08 \pm 0.08$} & {\cellcolor{bestgray}{\scriptsize\boldmath $56.29 \pm 0.41$}} & {\cellcolor{bestgray}{\scriptsize\boldmath $88.13 \pm 0.00$}} & {\cellcolor{bestgray}{\scriptsize\boldmath $56.38 \pm 0.00$}} \\
 & \scriptsize PPRFE & {\scriptsize $75.32 \pm 7.32$} & \cellcolor{stdblue}{\scriptsize $74.98 \pm 2.74$} & \cellcolor{stdblue}{\scriptsize $71.11 \pm 1.67$} & \cellcolor{stdblue}{\scriptsize $71.09 \pm 1.56$} & \cellcolor{stdblue}{\scriptsize $84.53 \pm 1.03$} & \cellcolor{stdblue}{\scriptsize $75.59 \pm 0.47$} & {\scriptsize $36.15 \pm 0.42$} & {\scriptsize $0.38 \pm 0.01$} & \cellcolor{stdblue}{\scriptsize $76.08 \pm 0.08$} & \cellcolor{stdblue}{\scriptsize $56.29 \pm 0.41$} & {\cellcolor{bestgray}{\scriptsize\boldmath $88.13 \pm 0.00$}} & {\cellcolor{bestgray}{\scriptsize\boldmath $56.38 \pm 0.00$}} \\
\bottomrule
\end{tabular}
\end{adjustbox}
\end{table}

\subsection{GPSE Pretrained Models}

The performance of \textbf{HOPSE-GPSE} is conditioned on the structural and positional priors learned by the underlying \textsc{GPSE} model during its pre-training phase \citep{canturk2023graph}. We evaluated two distinct \textsc{GPSE} variants, pre-trained on the MolPCBA and ZINC datasets, respectively. The comparative results across the benchmark suite are summarized in \cref{tbl:gpse_pretrained_comparison}.
The empirical evidence suggests that while both pre-trained models provide high-quality structural encodings, their effectiveness is sensitive to the downstream task. This task-dependency justifies treating the choice of the \textsc{GPSE} pre-trained model as a categorical hyperparameter within our exhaustive search strategy, as no single pre-trained initialization serves as a universal optimum across all evaluated domains.

\definecolor{stdblue}{HTML}{C9DAF8}
\definecolor{bestgray}{HTML}{D9D9D9}

\begin{table}[t]
\caption{\textbf{GPSE Pre-training Corpus Ablation.} Comparative predictive performance of the HOPSE-GPSE architecture when initialized with structural encodings from GPSE models pre-trained on MolPCBA versus ZINC datasets. \protect\colorbox{bestgray}{\textbf{Bold}} indicates the highest mean performance, while \protect\colorbox{stdblue}{blue} indicates results without statistically significant differences.}
\label{tbl:gpse_pretrained_comparison}
\centering
\begin{adjustbox}{width=1.\textwidth}
\renewcommand{\arraystretch}{1.2}
\begin{tabular}{@{}lcccccccccccc@{}}
\toprule
 & \multicolumn{8}{c}{\mbox{Graph}} & \multicolumn{4}{c}{\mbox{Simplicial}} \\
\cmidrule(lr){2-9} \cmidrule(lr){10-13}
Pretrained Model & \scriptsize MUTAG ($\uparrow$) & \scriptsize PROTEINS ($\uparrow$) & \scriptsize NCI1 ($\uparrow$) & \scriptsize NCI109 ($\uparrow$) & \scriptsize BBB ($\uparrow$) & \scriptsize CYP3A4 ($\uparrow$) & \scriptsize Cl.Hep. ($\downarrow$) & \scriptsize Caco2 ($\downarrow$) & \scriptsize NAME ($\uparrow$) & \scriptsize ORIENT ($\uparrow$) & \scriptsize $\beta_1$ ($\uparrow$) & \scriptsize $\beta_2$ ($\uparrow$) \\ 
\midrule
MolPCBA & 
\cellcolor{bestgray}{\scriptsize\boldmath $83.83 \pm 4.58$} & 
\cellcolor{stdblue}{\scriptsize $75.13 \pm 2.69$} & 
\cellcolor{stdblue}{\scriptsize $77.45 \pm 1.40$} & 
\cellcolor{bestgray}{\scriptsize\boldmath $76.05 \pm 0.59$} & 
\cellcolor{bestgray}{\scriptsize\boldmath $84.68 \pm 1.07$} & 
\cellcolor{stdblue}{\scriptsize $78.35 \pm 0.24$} & 
\cellcolor{bestgray}{\scriptsize\boldmath $33.07 \pm 0.43$} & 
\cellcolor{bestgray}{\scriptsize\boldmath $0.36 \pm 0.01$} & 
\cellcolor{bestgray}{\scriptsize\boldmath $80.23 \pm 3.54$} & 
\cellcolor{bestgray}{\scriptsize\boldmath $61.88 \pm 2.99$} & 
\cellcolor{bestgray}{\scriptsize\boldmath $88.13 \pm 0.00$} & 
\cellcolor{bestgray}{\scriptsize\boldmath $56.38 \pm 0.00$} \\

ZINC & 
\cellcolor{stdblue}{\scriptsize $81.28 \pm 7.89$} & 
\cellcolor{bestgray}{\scriptsize\boldmath $75.63 \pm 2.07$} & 
\cellcolor{bestgray}{\scriptsize\boldmath $78.64 \pm 1.31$} & 
\cellcolor{stdblue}{\scriptsize $75.41 \pm 1.10$} & 
\cellcolor{stdblue}{\scriptsize $84.53 \pm 1.24$} & 
\cellcolor{bestgray}{\scriptsize\boldmath $79.03 \pm 0.29$} & 
{\scriptsize $37.04 \pm 0.78$} & 
\cellcolor{stdblue}{\scriptsize $0.37 \pm 0.01$} & 
\cellcolor{stdblue}{\scriptsize $79.98 \pm 3.72$} & 
\cellcolor{stdblue}{\scriptsize $59.41 \pm 2.50$} & 
\cellcolor{bestgray}{\scriptsize\boldmath $88.13 \pm 0.00$} & 
\cellcolor{bestgray}{\scriptsize\boldmath $56.38 \pm 0.00$} \\
\bottomrule
\end{tabular}
\end{adjustbox}
\end{table}

\subsection{GNN Encoding Ablation}
In this section, we examine the effect of integrating encodings into baseline GNN architectures. \cref{tbl:gnn_ablation} compares \textsc{GCN}, \textsc{GAT}, and \textsc{GIN} with no additional encodings (\textsc{NO-ENC}) against configurations using feature (\textsc{F}) or topological encodings (\textsc{C}). The results consistently show that augmenting baseline GNNs with \texttt{C} or feature encodings boosts performance across multiple datasets, such as \textsc{MUTAG} and \textsc{NCI1}. However, there is no single dominant encoding type across all models and datasets; for instance, while \textsc{C} provides the most significant gains for \textsc{GAT} and \textsc{GIN} on \textsc{MUTAG}, feature encodings (\textsc{F}) are more effective for \textsc{GCN} on the same dataset. This suggests that while encodings are generally beneficial, there is no clear universal trend, and the optimal choice must be determined on a case-by-case basis depending on the specific model-dataset combination.

\definecolor{stdblue}{HTML}{C9DAF8}
\definecolor{bestgray}{HTML}{D9D9D9}
\begin{table}[t]
\caption{\textbf{GNN Baseline Encoding Ablation.} Performance impact of integrating positional and structural encodings into standard GNN architectures. The configurations represent base models with no additional encodings ($\times$), augmented with feature encodings (F), and augmented with connectivity encodings (C). \protect\colorbox{bestgray}{\textbf{Bold}} signifies the highest mean performance per model and dataset, with \protect\colorbox{stdblue}{blue} indicating non-significant deviations.}
\label{tbl:gnn_ablation}
\centering
\begin{adjustbox}{width=1.\textwidth}
\renewcommand{\arraystretch}{1.2}
\begin{tabular}{@{}llcccccccccccc@{}}
\toprule
 & & \multicolumn{8}{c}{\mbox{Graph}} & \multicolumn{4}{c}{\mbox{Simplicial}} \\
\cmidrule(lr){3-10} \cmidrule(lr){11-14}
Model & Enc. & \scriptsize MUTAG ($\uparrow$) & \scriptsize PROTEINS ($\uparrow$) & \scriptsize NCI1 ($\uparrow$) & \scriptsize NCI109 ($\uparrow$) & \scriptsize BBB ($\uparrow$) & \scriptsize CYP3A4 ($\uparrow$) & \scriptsize Cl.Hep. ($\downarrow$) & \scriptsize Caco2 ($\downarrow$) & \scriptsize NAME ($\uparrow$) & \scriptsize ORIENT ($\uparrow$) & \scriptsize $\beta_1$ ($\uparrow$) & \scriptsize $\beta_2$ ($\uparrow$) \\
\midrule
\multirow{3}{*}{\textbf{GAT}} & $\times$ & {\scriptsize $69.79 \pm 7.17$} & {\cellcolor{bestgray}{\scriptsize\boldmath $75.05 \pm 1.08$}} & {\scriptsize $72.53 \pm 1.00$} & {\scriptsize $72.89 \pm 2.17$} & {\cellcolor{bestgray}{\scriptsize\boldmath $83.60 \pm 0.12$}} & {\scriptsize $77.58 \pm 0.39$} & {\scriptsize $36.54 \pm 1.22$} & {\scriptsize $1.07 \pm 0.00$} & \cellcolor{stdblue}{\scriptsize $17.92 \pm 0.23$} & {\cellcolor{bestgray}{\scriptsize\boldmath $47.94 \pm 0.00$}} & \cellcolor{stdblue}{\scriptsize $7.36 \pm 0.16$} & {\cellcolor{bestgray}{\scriptsize\boldmath $47.94 \pm 0.00$}} \\
 & F & \cellcolor{stdblue}{\scriptsize $83.83 \pm 4.38$} & \cellcolor{stdblue}{\scriptsize $73.48 \pm 3.02$} & {\cellcolor{bestgray}{\scriptsize\boldmath $74.94 \pm 2.05$}} & {\cellcolor{bestgray}{\scriptsize\boldmath $75.57 \pm 1.80$}} & {\scriptsize $82.51 \pm 0.16$} & {\scriptsize $77.19 \pm 0.15$} & {\cellcolor{bestgray}{\scriptsize\boldmath $33.94 \pm 0.85$}} & {\scriptsize $0.97 \pm 0.03$} & {\cellcolor{bestgray}{\scriptsize\boldmath $18.38 \pm 1.17$}} & {\cellcolor{bestgray}{\scriptsize\boldmath $47.94 \pm 0.00$}} & \cellcolor{stdblue}{\scriptsize $7.45 \pm 0.04$} & {\cellcolor{bestgray}{\scriptsize\boldmath $47.94 \pm 0.00$}} \\
 & C & {\cellcolor{bestgray}{\scriptsize\boldmath $85.11 \pm 4.66$}} & \cellcolor{stdblue}{\scriptsize $72.19 \pm 3.55$} & \cellcolor{stdblue}{\scriptsize $74.75 \pm 0.95$} & \cellcolor{stdblue}{\scriptsize $74.62 \pm 1.17$} & {\scriptsize $83.40 \pm 0.12$} & {\cellcolor{bestgray}{\scriptsize\boldmath $78.61 \pm 0.08$}} & {\scriptsize $36.98 \pm 1.23$} & {\cellcolor{bestgray}{\scriptsize\boldmath $0.65 \pm 0.00$}} & \cellcolor{stdblue}{\scriptsize $18.09 \pm 0.58$} & {\cellcolor{bestgray}{\scriptsize\boldmath $47.94 \pm 0.00$}} & {\cellcolor{bestgray}{\scriptsize\boldmath $7.45 \pm 0.04$}} & {\cellcolor{bestgray}{\scriptsize\boldmath $47.94 \pm 0.00$}} \\
\midrule
\multirow{3}{*}{\textbf{GCN}} & $\times$ & {\scriptsize $68.94 \pm 7.68$} & \cellcolor{stdblue}{\scriptsize $75.13 \pm 2.47$} & {\scriptsize $70.86 \pm 0.72$} & {\scriptsize $71.21 \pm 1.59$} & {\cellcolor{bestgray}{\scriptsize\boldmath $85.47 \pm 0.16$}} & {\scriptsize $77.42 \pm 0.91$} & {\scriptsize $37.42 \pm 0.01$} & {\scriptsize $1.21 \pm 0.00$} & {\cellcolor{bestgray}{\scriptsize\boldmath $22.80 \pm 3.18$}} & {\cellcolor{bestgray}{\scriptsize\boldmath $47.94 \pm 0.00$}} & {\cellcolor{bestgray}{\scriptsize\boldmath $36.67 \pm 10.87$}} & \cellcolor{stdblue}{\scriptsize $47.94 \pm 0.00$} \\
 & F & {\cellcolor{bestgray}{\scriptsize\boldmath $85.53 \pm 4.54$}} & \cellcolor{stdblue}{\scriptsize $74.05 \pm 4.10$} & {\scriptsize $72.51 \pm 1.57$} & {\cellcolor{bestgray}{\scriptsize\boldmath $73.44 \pm 1.07$}} & {\scriptsize $83.69 \pm 0.36$} & {\cellcolor{bestgray}{\scriptsize\boldmath $78.68 \pm 0.10$}} & {\cellcolor{bestgray}{\scriptsize\boldmath $34.76 \pm 0.03$}} & {\scriptsize $1.03 \pm 0.02$} & \cellcolor{stdblue}{\scriptsize $21.59 \pm 1.05$} & {\cellcolor{bestgray}{\scriptsize\boldmath $47.94 \pm 0.00$}} & \cellcolor{stdblue}{\scriptsize $29.01 \pm 0.70$} & {\cellcolor{bestgray}{\scriptsize\boldmath $48.09 \pm 0.31$}} \\
 & C & \cellcolor{stdblue}{\scriptsize $85.11 \pm 5.55$} & {\cellcolor{bestgray}{\scriptsize\boldmath $75.41 \pm 1.81$}} & {\cellcolor{bestgray}{\scriptsize\boldmath $74.69 \pm 1.90$}} & \cellcolor{stdblue}{\scriptsize $73.34 \pm 1.11$} & {\scriptsize $79.61 \pm 1.87$} & {\scriptsize $77.42 \pm 0.37$} & {\scriptsize $36.88 \pm 0.58$} & {\cellcolor{bestgray}{\scriptsize\boldmath $0.81 \pm 0.14$}} & \cellcolor{stdblue}{\scriptsize $21.76 \pm 4.89$} & {\cellcolor{bestgray}{\scriptsize\boldmath $47.94 \pm 0.00$}} & {\scriptsize $19.71 \pm 0.79$} & \cellcolor{stdblue}{\scriptsize $47.94 \pm 0.00$} \\
\midrule
\multirow{3}{*}{\textbf{GIN}} & $\times$ & {\scriptsize $75.32 \pm 7.32$} & {\cellcolor{bestgray}{\scriptsize\boldmath $75.84 \pm 1.82$}} & {\scriptsize $70.62 \pm 1.34$} & {\cellcolor{bestgray}{\scriptsize\boldmath $73.38 \pm 0.99$}} & {\cellcolor{bestgray}{\scriptsize\boldmath $84.38 \pm 0.94$}} & \cellcolor{stdblue}{\scriptsize $78.05 \pm 0.54$} & {\scriptsize $36.76 \pm 0.16$} & {\scriptsize $1.13 \pm 0.07$} & - & - & - & - \\
 & F & \cellcolor{stdblue}{\scriptsize $83.40 \pm 3.90$} & {\scriptsize $71.97 \pm 3.86$} & {\cellcolor{bestgray}{\scriptsize\boldmath $73.97 \pm 1.62$}} & \cellcolor{stdblue}{\scriptsize $72.82 \pm 2.41$} & \cellcolor{stdblue}{\scriptsize $83.89 \pm 0.20$} & {\cellcolor{bestgray}{\scriptsize\boldmath $78.82 \pm 0.73$}} & {\cellcolor{bestgray}{\scriptsize\boldmath $33.93 \pm 0.60$}} & {\scriptsize $1.08 \pm 0.00$} & - & - & - & - \\
 & C & {\cellcolor{bestgray}{\scriptsize\boldmath $83.83 \pm 3.95$}} & \cellcolor{stdblue}{\scriptsize $72.33 \pm 4.00$} & \cellcolor{stdblue}{\scriptsize $73.83 \pm 1.13$} & \cellcolor{stdblue}{\scriptsize $73.07 \pm 1.25$} & {\scriptsize $82.91 \pm 0.30$} & {\scriptsize $77.64 \pm 0.41$} & \cellcolor{stdblue}{\scriptsize $34.15 \pm 0.00$} & {\cellcolor{bestgray}{\scriptsize\boldmath $0.88 \pm 0.00$}} & - & - & - & - \\
\bottomrule
\end{tabular}
\end{adjustbox}
\end{table}

\subsection{HOMP Baselines with Hasse-Graph Encodings}
\label{app:homp_hasse_ablation}

Evaluating HOMP baselines enriched with Hasse-graph Positional and Structural Encodings (PSEs) provides a valuable ablation to determine whether encodings act as a complement to or substitute for higher-order message passing. Consequently, we performed a controlled experiment in which the initial rank-0 features of the HOMP baselines were augmented with CombinedPE features derived from graph-based adjacency. To isolate the effect of the additional encodings, we used the best-validation hyperparameters of each original model and reran it with CombinedPE under the same evaluation protocol.

The benefit of adding PSEs to HOMP architectures is dataset-dependent rather than systematic. As shown in \cref{tbl:homp_hasse_encodings}, the augmented variants improve the reported mean for all five models on Caco2 and four of the five models on MUTAG, but improvements are not consistently observed on the other datasets. Across all model--dataset pairs, adding PSEs improves the mean in 13 of 40 comparisons. Moreover, the best-performing method in each corresponding column of our main results remains unchanged. These results indicate that while PSEs can complement higher-order message passing on particular tasks, they do not provide a predictable or universal improvement when added to an existing HOMP architecture.

We note, however, that existing HOMP baselines are structurally designed to accept fixed, uniform input feature dimensions per rank. Handling arbitrary collections of higher-order neighborhoods---which induce distinct Hasse graphs and potentially variable feature dimensions per target rank---is a unique design characteristic of \textsc{HOPSE}'s target-rank aggregation logic ($\psi_k$). As a result, fully applying \textsc{HOPSE}'s multi-neighborhood encoding scheme directly to standard HOMP architectures requires non-trivial structural adaptations that are unique to the \textsc{HOPSE} design.

\begin{table}[htbp]
\caption{\textbf{HOMP Baseline Ablation with Hasse-Graph Encodings.} Test scores (mean $\pm$ std over seeds) comparing the best-validation sweep configurations (Original) against the same hyperparameters augmented with CombinedPE (+All PSEs). \protect\colorbox{bestgray}{\textbf{Bold}} signifies the highest mean performance per model and dataset, with \protect\colorbox{stdblue}{blue} indicating non-significant deviations.}
\label{tbl:homp_hasse_encodings}
\centering
\begin{adjustbox}{width=1.\textwidth}
\renewcommand{\arraystretch}{1.4}
\begin{tabular}{@{}llcccccccc@{}}
\toprule
\textbf{Model} & \textbf{Setting} & \scriptsize MUTAG ($\uparrow$) & \scriptsize PROTEINS ($\uparrow$) & \scriptsize NCI1 ($\uparrow$) & \scriptsize NCI109 ($\uparrow$) & \scriptsize BBB ($\uparrow$) & \scriptsize CYP3A4 ($\uparrow$) & \scriptsize Cl.Hep. ($\downarrow$) & \scriptsize Caco2 ($\downarrow$) \\
\midrule
\multirow{2}{*}{\textbf{CCCN}} 
 & Original & \cellcolor{stdblue}{\scriptsize $77.87 \pm 3.95$} & {\cellcolor{bestgray}{\scriptsize\boldmath $73.69 \pm 2.18$}} & {\cellcolor{bestgray}{\scriptsize\boldmath $76.73 \pm 1.83$}} & {\cellcolor{bestgray}{\scriptsize\boldmath $74.97 \pm 1.03$}} & {\scriptsize $83.25 \pm 0.00$} & {\cellcolor{bestgray}{\scriptsize\boldmath $76.24 \pm 0.15$}} & {\cellcolor{bestgray}{\scriptsize\boldmath $35.58 \pm 0.09$}} & {\scriptsize $0.93 \pm 0.01$} \\
 & +All PSEs & {\cellcolor{bestgray}{\scriptsize\boldmath $81.70 \pm 5.80$}} & \cellcolor{stdblue}{\scriptsize $72.83 \pm 3.72$} & \cellcolor{stdblue}{\scriptsize $74.59 \pm 1.17$} & \cellcolor{stdblue}{\scriptsize $73.80 \pm 1.71$} & {\cellcolor{bestgray}{\scriptsize\boldmath $84.53 \pm 0.98$}} & {\scriptsize $75.30 \pm 0.35$} & {\scriptsize $36.33 \pm 0.06$} & {\cellcolor{bestgray}{\scriptsize\boldmath $0.75 \pm 0.01$}} \\
\midrule
\multirow{2}{*}{\textbf{CWN}} 
 & Original & \cellcolor{stdblue}{\scriptsize $81.70 \pm 7.32$} & {\cellcolor{bestgray}{\scriptsize\boldmath $76.20 \pm 2.05$}} & {\cellcolor{bestgray}{\scriptsize\boldmath $73.77 \pm 1.37$}} & {\cellcolor{bestgray}{\scriptsize\boldmath $73.86 \pm 1.20$}} & {\cellcolor{bestgray}{\scriptsize\boldmath $82.96 \pm 1.24$}} & {\cellcolor{bestgray}{\scriptsize\boldmath $76.84 \pm 0.28$}} & {\scriptsize $35.49 \pm 0.27$} & {\scriptsize $1.00 \pm 0.01$} \\
 & +All PSEs & {\cellcolor{bestgray}{\scriptsize\boldmath $84.68 \pm 5.28$}} & \cellcolor{stdblue}{\scriptsize $72.69 \pm 1.69$} & {\scriptsize $71.38 \pm 0.85$} & {\scriptsize $68.79 \pm 1.46$} & \cellcolor{stdblue}{\scriptsize $82.27 \pm 0.85$} & {\scriptsize $76.22 \pm 0.20$} & {\cellcolor{bestgray}{\scriptsize\boldmath $34.87 \pm 0.01$}} & {\cellcolor{bestgray}{\scriptsize\boldmath $0.76 \pm 0.01$}} \\
\midrule
\multirow{2}{*}{\textbf{SCCNN}} 
 & Original & {\scriptsize $74.47 \pm 3.01$} & {\cellcolor{bestgray}{\scriptsize\boldmath $77.06 \pm 2.04$}} & \cellcolor{stdblue}{\scriptsize $71.40 \pm 1.45$} & {\cellcolor{bestgray}{\scriptsize\boldmath $71.87 \pm 0.93$}} & {\cellcolor{bestgray}{\scriptsize\boldmath $82.96 \pm 0.55$}} & {\cellcolor{bestgray}{\scriptsize\boldmath $77.29 \pm 0.08$}} & {\cellcolor{bestgray}{\scriptsize\boldmath $36.40 \pm 0.01$}} & {\scriptsize $1.04 \pm 0.01$} \\
 & +All PSEs & {\cellcolor{bestgray}{\scriptsize\boldmath $82.98 \pm 2.69$}} & \cellcolor{stdblue}{\scriptsize $73.41 \pm 1.61$} & {\cellcolor{bestgray}{\scriptsize\boldmath $72.88 \pm 1.26$}} & \cellcolor{stdblue}{\scriptsize $70.86 \pm 1.14$} & \cellcolor{stdblue}{\scriptsize $82.57 \pm 0.75$} & {\scriptsize $73.32 \pm 0.27$} & {\scriptsize $37.32 \pm 0.13$} & {\cellcolor{bestgray}{\scriptsize\boldmath $0.71 \pm 0.01$}} \\
\midrule
\multirow{2}{*}{\textbf{TopoTune (cell)}} 
 & Original & \cellcolor{stdblue}{\scriptsize $77.02 \pm 9.36$} & {\cellcolor{bestgray}{\scriptsize\boldmath $76.06 \pm 2.01$}} & {\cellcolor{bestgray}{\scriptsize\boldmath $77.28 \pm 1.52$}} & {\cellcolor{bestgray}{\scriptsize\boldmath $76.57 \pm 0.68$}} & {\cellcolor{bestgray}{\scriptsize\boldmath $84.93 \pm 1.07$}} & {\cellcolor{bestgray}{\scriptsize\boldmath $78.91 \pm 0.41$}} & {\cellcolor{bestgray}{\scriptsize\boldmath $36.69 \pm 0.47$}} & {\scriptsize $0.85 \pm 0.01$} \\
 & +All PSEs & {\cellcolor{bestgray}{\scriptsize\boldmath $82.13 \pm 7.32$}} & \cellcolor{stdblue}{\scriptsize $73.05 \pm 2.25$} & \cellcolor{stdblue}{\scriptsize $76.38 \pm 0.60$} & {\scriptsize $74.40 \pm 1.09$} & {\scriptsize $82.71 \pm 0.48$} & {\scriptsize $76.85 \pm 0.62$} & \cellcolor{stdblue}{\scriptsize $37.01 \pm 1.01$} & {\cellcolor{bestgray}{\scriptsize\boldmath $0.71 \pm 0.01$}} \\
\midrule
\multirow{2}{*}{\textbf{TopoTune (simplicial)}} 
 & Original & {\cellcolor{bestgray}{\scriptsize\boldmath $84.26 \pm 8.14$}} & {\cellcolor{bestgray}{\scriptsize\boldmath $75.48 \pm 1.17$}} & {\cellcolor{bestgray}{\scriptsize\boldmath $77.45 \pm 1.07$}} & {\cellcolor{bestgray}{\scriptsize\boldmath $76.44 \pm 0.98$}} & {\cellcolor{bestgray}{\scriptsize\boldmath $84.68 \pm 1.13$}} & \cellcolor{stdblue}{\scriptsize $77.83 \pm 0.16$} & {\cellcolor{bestgray}{\scriptsize\boldmath $35.70 \pm 0.88$}} & {\scriptsize $0.76 \pm 0.03$} \\
 & +All PSEs & \cellcolor{stdblue}{\scriptsize $81.28 \pm 7.54$} & \cellcolor{stdblue}{\scriptsize $70.47 \pm 3.88$} & \cellcolor{stdblue}{\scriptsize $75.51 \pm 1.27$} & {\scriptsize $74.02 \pm 0.39$} & \cellcolor{stdblue}{\scriptsize $82.86 \pm 1.35$} & {\cellcolor{bestgray}{\scriptsize\boldmath $77.89 \pm 0.66$}} & \cellcolor{stdblue}{\scriptsize $36.27 \pm 1.43$} & {\cellcolor{bestgray}{\scriptsize\boldmath $0.65 \pm 0.02$}} \\
\bottomrule
\end{tabular}
\end{adjustbox}
\end{table}

\subsection{VRAM Consumption vs. Baselines}
\label{app:vram_baselines}

\begin{figure}[h]
    \centering
    \includegraphics[width=0.9\linewidth]{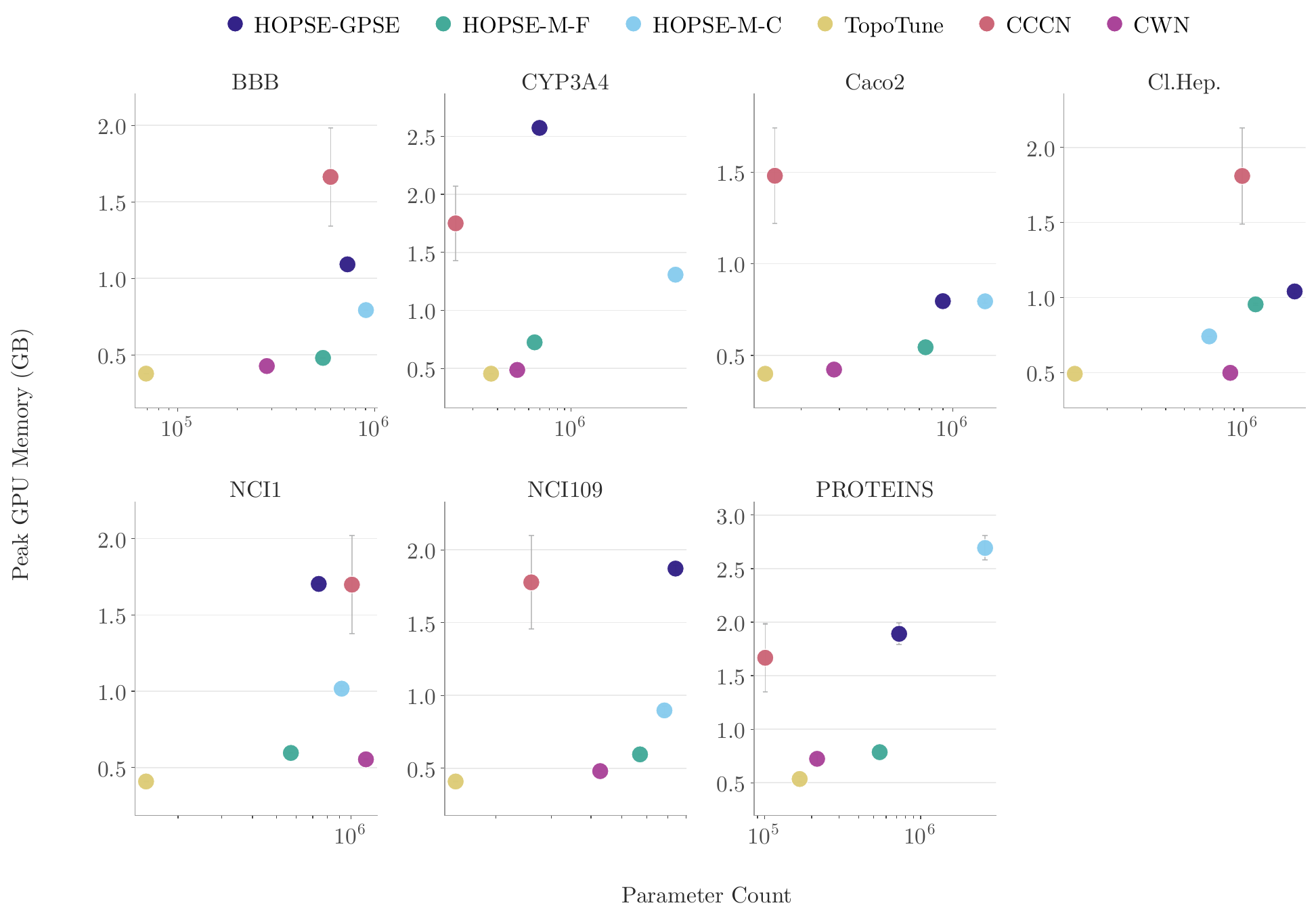}
    \caption{Peak GPU memory vs. parameter count, Cell domain.}
    \label{fig:vram_cell}
\end{figure}

\begin{figure}[h]
    \centering
    \includegraphics[width=0.9\linewidth]{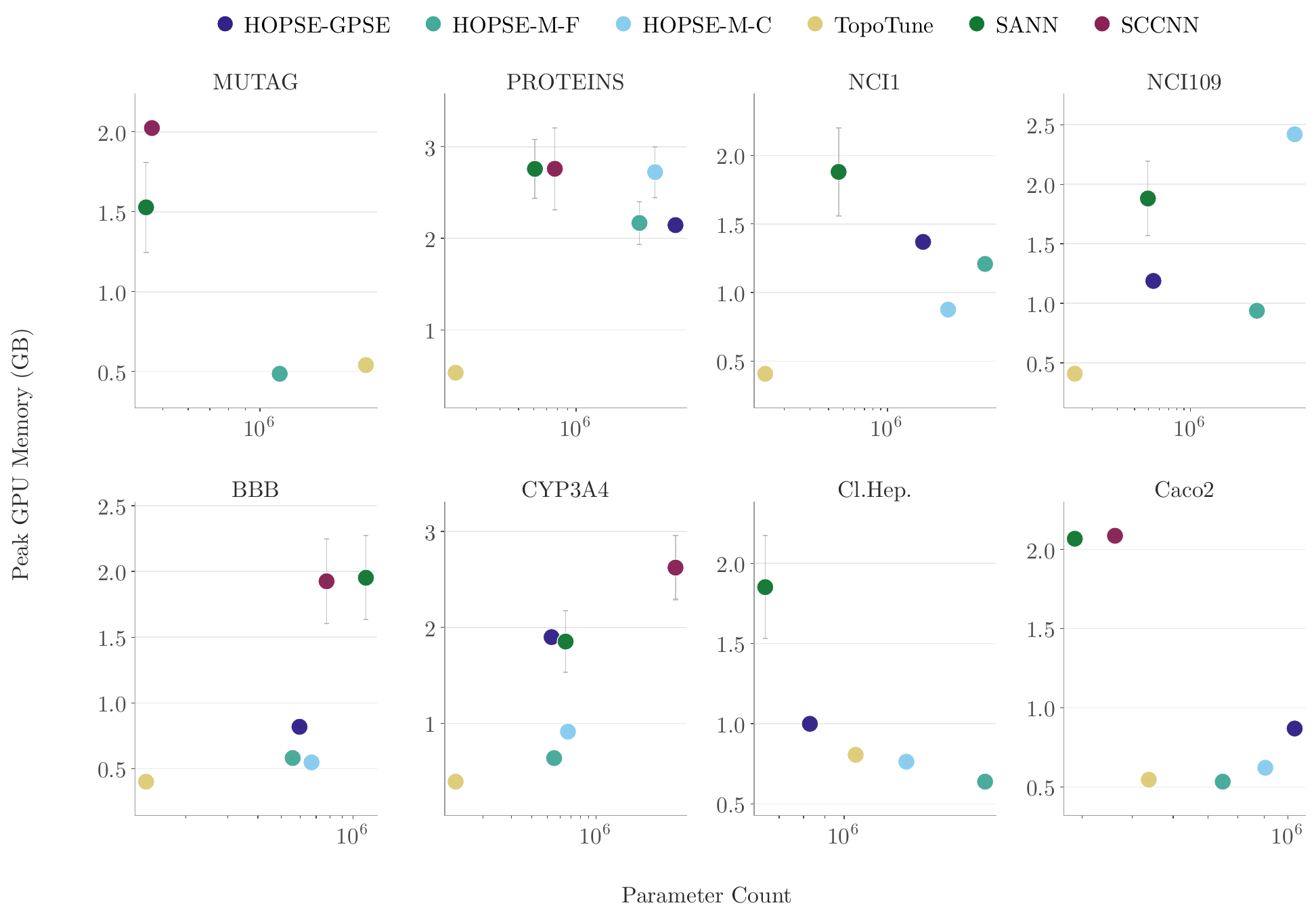}
    \caption{Peak GPU memory vs. parameter count, Simplicial domain.}
    \label{fig:vram_graph}
\end{figure}

Complementing the theoretical memory analysis in \cref{app:complexity}, \cref{fig:vram_cell,fig:vram_graph} report the peak GPU memory usage of the best-validation configuration for each model, plotted against its total parameter count, across the Cell and Simplicial domains. Consistent with \cref{app:complexity}, HOPSE variants do not universally minimize VRAM usage relative to HOMP baselines; rather, peak memory scales with the selected encoding multiplicity and parameter footprint. Notably, HOPSE-M-C and HOPSE-GPSE remain within a comparable memory range to SANN and Topotune despite their higher parameter counts, and consistently avoid the memory spikes exhibited by CCCN and SCCNN.

\subsection{CWN Baseline Discrepancy with TopoBench}
\label{app:cwn_protocol}

\cref{tbl:cwn_protocol_diff} details the protocol differences between the original CWN evaluation~\citep{bodnar2021weisfeiler} and the standardized TopoBench~\citep{telyatnikov2025topobench} setup used throughout this paper. Our reported CWN numbers align with official TopoBench reproducibility reports (e.g., MUTAG $80.43 \pm 1.78$, NCI1 $73.93 \pm 1.87$), within one standard deviation. This shift is not specific to CWN: standard GNN baselines such as GIN exhibit a comparable performance drop under the same transition, while CWN's relative margin over GIN is preserved, indicating the discrepancy is protocol-driven rather than an artifact of our reproduction. As a targeted check, reverting only the NCI1 data split from TopoBench's 50/25/25 partition to CWN's original 10-fold cross-validation (90\% train) raised accuracy from $74.51$ to $79.08 \pm 2.50$, recovering most of the gap to the originally reported $83.6$. The remaining difference is attributable to TopoBench's shared hyperparameter grid, fixed learning-rate schedule, and common encoders/readouts across all baselines. Because these constraints apply uniformly to HOPSE and to every baseline, the relative ordering in \cref{tbl:hopse_wandb_graph_sim_compact} and our efficiency--performance conclusions in \cref{sec:experiments} remain unaffected.

\begin{table}[h]
\centering
\caption{Protocol differences between the original CWN evaluation and TopoBench.}
\label{tbl:cwn_protocol_diff}
\begin{tabular}{lll}
\toprule
Protocol aspect & Original CWN & TopoBench (this paper) \\
\midrule
Data split & 10-fold CV (90\% train) & 50/25/25 train/val/test \\
Hyperparameter search & Per-dataset, bespoke & Shared 5-axis grid, all models \\
Learning rate schedule & Model-specific & Fixed across models \\
Encoders / readouts & Model-specific & Shared across models \\
\bottomrule
\end{tabular}
\end{table}
\section{Extended Training Scalability Analysis}
\label{app:scalability}
We analyze how training time scales with the size and topological complexity of datasets. We define \textbf{topological size} as the total count of topological structures in the training split: for simplicial complexes, this is the sum of undirected edges plus triangles; for cell complexes, this is undirected edges plus 2-cells. Higher topological size indicates both more data and more complex higher-order structure.

\begin{figure}[t]
    \centering
    \includegraphics[width=0.9\textwidth]{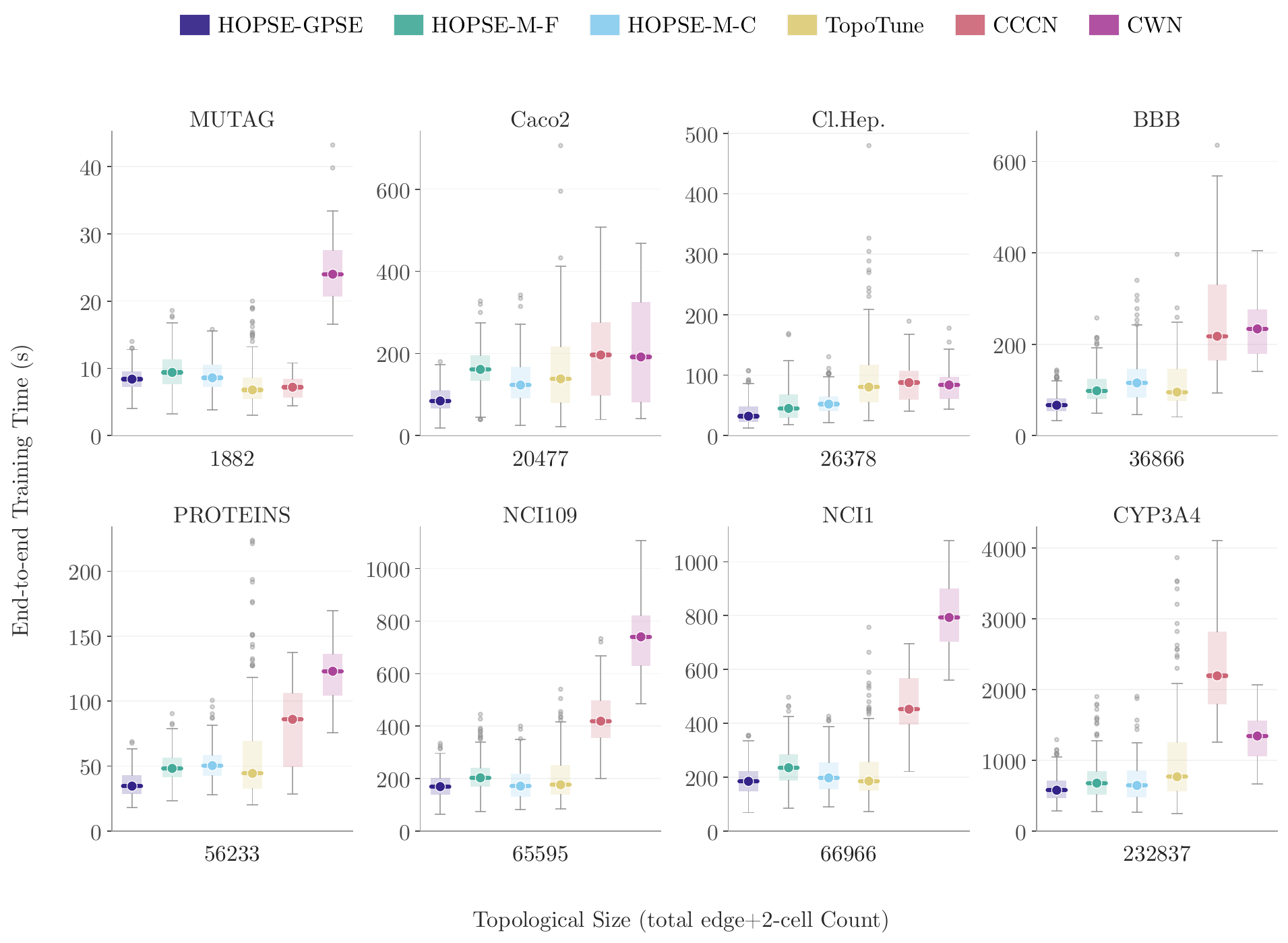}
    \caption{\textbf{Cell domain — Training time vs. topological size.} Topological size = undirected edges + 2-cells (train split). Each box shows the distribution of training times over all hyperparameter runs. HOPSE-GPSE (dark blue) consistently achieves among the fastest times, particularly on larger datasets. The efficiency advantage over HOMP models CWN and CCCN (orange, red) is clear across the complexity spectrum but emphasized on larger more topologically rich datasets ("topological size").}
    \label{fig:train_time_cell}
\end{figure}

\begin{figure}[t]
    \centering
    \includegraphics[width=0.9\textwidth]{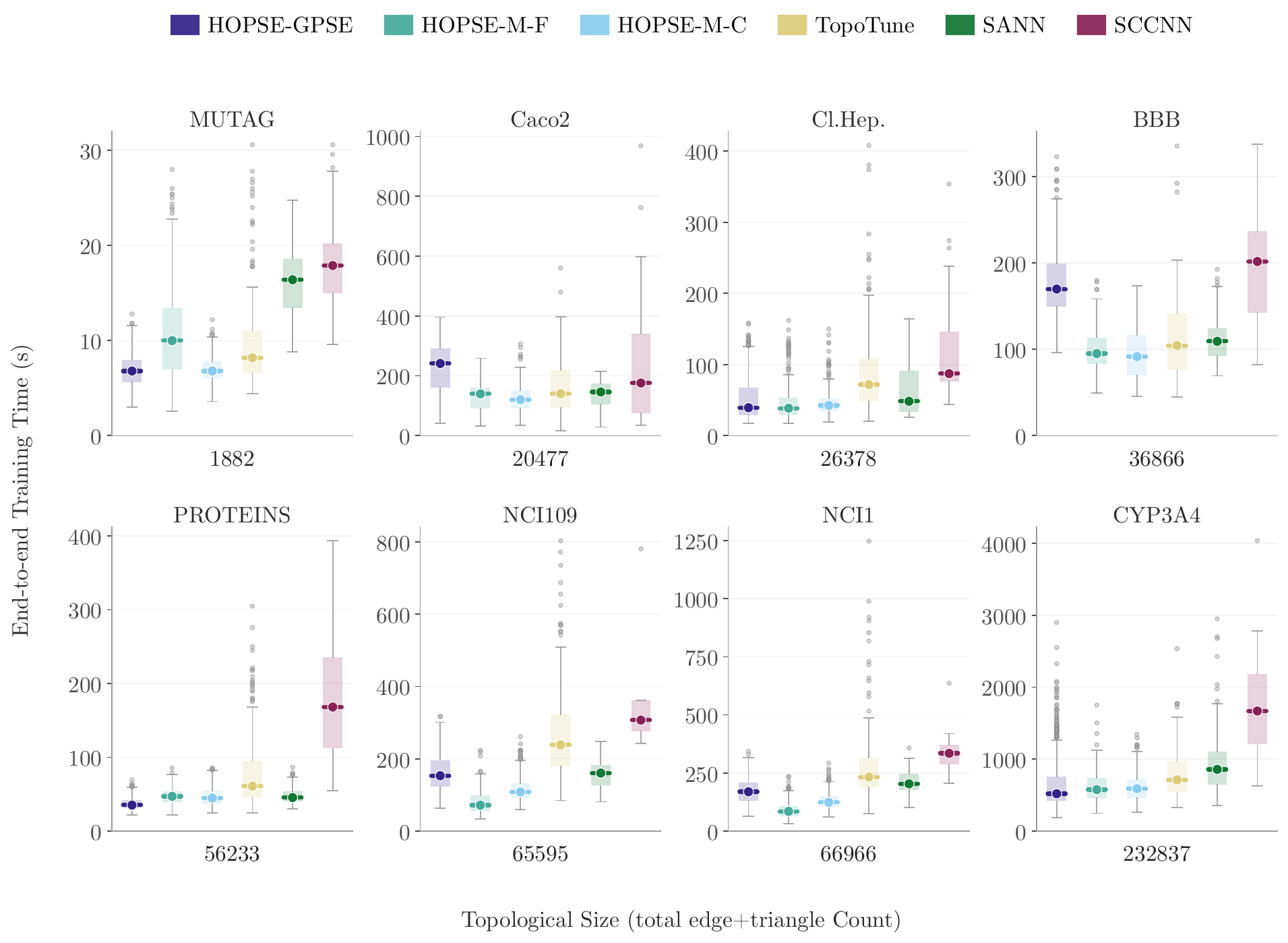}
    \caption{\textbf{Simplicial domain (graph datasets) — Training time vs. topological size.} Topological size = undirected edges + triangles (train split). HOPSE models (three left-most bars) show increasingly favorable scaling compared to the HOMP model SCCNN (red) as topological size grows.}
    \label{fig:train_time_simplicial_graph}
\end{figure}

\begin{figure}[t]
    \centering
    \includegraphics[width=0.8\textwidth]{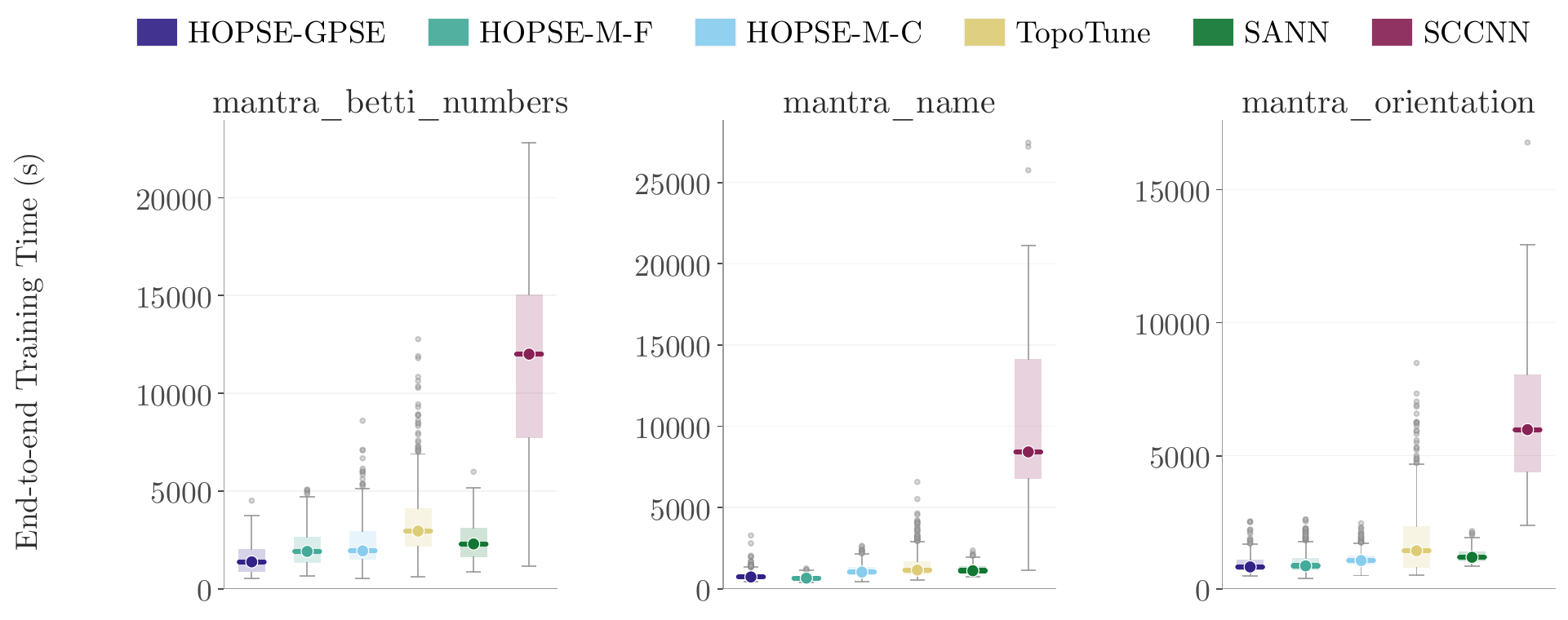}
    \caption{\textbf{Simplicial domain (MANTRA) — Training time vs. topological size.} All three MANTRA tasks have identical topological size (1.24M edges + triangles). HOPSE models demonstrate significant speedup on large-scale topological structures compared to HOMP-based model SCCNN.}
    \label{fig:train_time_simplicial_mantra}
\end{figure}

Figures~\ref{fig:train_time_cell}--\ref{fig:train_time_simplicial_mantra} show end-to-end training time distributions across \emph{all hyperparameter configurations tested}. Each boxplot aggregates timing results over every hyperparameter setting for a given model-dataset pair, showing the full distribution rather than single point estimates. Datasets are ordered left-to-right by increasing topological size.

Among baselines, SANN also precomputes structural information (via multi-hop 
feature aggregations) rather than using message passing, making it the most 
directly comparable to HOPSE. Other comparison targets are the 
message-passing HOMP models: SCCNN for simplicial complexes, and CWN and 
CCCN for cell complexes. TopoTune, while offering a flexible GCCN framework, 
employs message passing in the configurations tested here and thus exhibits 
similar computational characteristics to other HOMP approaches.

Across both cell and simplicial domains on graph datasets (Figures~\ref{fig:train_time_cell}--\ref{fig:train_time_simplicial_graph}), HOPSE variants consistently achieve competitive or superior training efficiency compared to all baselines. HOPSE-GPSE demonstrates particularly strong performance, matching or outperforming the optimized TopoTune model while maintaining substantially lower training times than HOMP-based approaches. Critically, the efficiency advantage over HOMP models (CWN, CCCN, SCCNN) grows with topological size: differences are modest on smaller datasets but become increasingly pronounced on larger, more complex structures, validating scalability claims. SANN, which similarly employs precomputation, shows intermediate performance between HOPSE and full HOMP models, while TopoTune remains competitive through architectural optimization.

The efficiency advantage becomes most pronounced on the MANTRA benchmark (Figure~\ref{fig:train_time_simplicial_mantra}), which contains purely topological tasks on substantially larger structures than molecular datasets. HOPSE variants achieve training times comparable to SANN and significantly faster than SCCNN, with the gap widening considerably compared to molecular benchmarks. Even the optimized TopoTune requires substantially more time than precomputation-based approaches, confirming that HOMP architectures face fundamental scalability challenges on large topological structures.

These results validate our theoretical complexity analysis: HOPSE scales linearly with topological size, while HOMP models face quadratic or higher costs. The preprocessing cost occurs only once and becomes increasingly outweighed by per-epoch savings on larger datasets. The consistent pattern across domains demonstrates that the efficiency gap widens as both dataset size and topological complexity increase—exactly where scalable TDL methods are most needed for real-world applications.
\section{Licenses}
\label{app:licenses}
This work does not distribute existing datasets. However, the dataset it uses will be mentioned here along with their licenses. The graph datasets \textsc{MUTAG}, \textsc{PROTEINS}, \textsc{NCI1}, \textsc{NCI109} do not specify a license and are distributed as part of \textsc{TUDatasets} accessed via \href{http:chrsmrrs.github.io/datasets}{https://pubs.acs.org/doi/full/10.1021/acs.jcim.5b00559}. 
The \textsc{BBB\_Martins}, \textsc{CYP3A4\_Veith}, \textsc{Clearance\_Hepatocyte\_AZ}, 
and \textsc{Caco2\_Wang} datasets are obtained from the Therapeutics Data Commons (TDC) 
and are licensed under the Creative Commons Attribution 4.0 International License 
(\href{https://creativecommons.org/licenses/by/4.0/}{CC BY 4.0}).
The \textsc{MANTRA} benchmarking dataset is licensed under the 3-Clause BSD license. 
Implementation-wise, this work is built upon the \texttt{TopoBench} library~\citep{telyatnikov2025topobench}, which is licensed under the MIT license.

\clearpage


\end{document}